\documentclass[10pt,twocolumn,letterpaper]{article}

\usepackage[accsupp]{axessibility}

\usepackage{tabularx}
\newcolumntype{Y}{>{\centering\arraybackslash}X}
\usepackage{booktabs}
\usepackage{pifont}
\usepackage{amssymb}
\usepackage{multirow}

\newcommand{\cmark}{\ding{51}}
\newcommand{\xmark}{\ding{55}}

\usepackage[pagenumbers]{cvpr}
\usepackage{subcaption}
\usepackage{graphicx}

\newcommand{\TODO}[1]{\textbf{\color{red}[TODO: #1]}}
\renewcommand{\TODO}[1]{}

\usepackage[skins]{tcolorbox}
\usepackage[table]{xcolor}

\definecolor{cvprblue}{rgb}{0.21,0.49,0.74}
\definecolor{lightgray}{gray}{0.8}

\newtcolorbox{myframe}[2][]{%
  left=2pt,right=2pt,enhanced,colback=white,colframe=blue,coltitle=black,
   boxrule=0.4pt,
  fonttitle=\itshape,
  attach boxed title to top left={yshift=-0.5\baselineskip-0.4pt,xshift=2mm},
  boxed title style={tile,size=minimal,left=0.1mm,right=0.1mm,
    colback=white,before upper=\strut},
  title=#2,#1
}

\definecolor{cvprblue}{rgb}{0.21,0.49,0.74}
\usepackage[pagebackref,breaklinks,colorlinks,allcolors=cvprblue]{hyperref}

\usepackage{xspace}

\DeclareSymbolFont{extraup}{U}{zavm}{m}{n}
\DeclareMathSymbol{\varheart}{\mathalpha}{extraup}{86}
\DeclareMathSymbol{\vardiamond}{\mathalpha}{extraup}{87}

\newcommand{\dav}{DepthAnything v2\xspace}
\newcommand{\daac}{DepthAnything-AC\xspace}
\newcommand{\depthcrafter}{DepthCrafter\xspace}
\newcommand{\lotus}{Lotus\xspace}
\newcommand{\midas}{MiDaS v3.1\xspace}
\newcommand{\marigold}{Marigold\xspace}
\newcommand{\moge}{MoGe-2\xspace}
\newcommand{\depthpro}{Depth Pro\xspace}
\newcommand{\metricd}{Metric-3D v2\xspace}
\newcommand{\unidepth}{UniDepth v2\xspace}
\newcommand{\mapanything}{MapAnything\xspace}
\newcommand{\vda}{VDA\xspace}
\newcommand{\metricanything}{Metric Anything\xspace}
\newcommand{\davthree}{DepthAnything 3\xspace}
\newcommand{\davft}{DAv2-FT\xspace}
\newcommand{\degree}{^{\circ}}
\newcommand{\lunarsim}{LunarSim\xspace}
\newcommand{\lusnar}{LuSNAR\xspace}
\newcommand{\etna}{Etna-LRNT\xspace}
\newcommand{\seli}{Etna-S3LI\xspace}
\newcommand{\change}{Chang'e-3\xspace}
\newcommand{\cheri}{CHERI\xspace}
\newcommand{\methodname}{LuMon\xspace}

\def\paperID{14}
\def\confName{CVPR}
\def\confYear{2026}

\title{\methodname: A Comprehensive Benchmark and Development Suite with \\ Novel Datasets for Lunar Monocular Depth Estimation}

\author{
Aytac Sekmen$^{1,3, \varheart}$\;\;\; 
Fatih Gunes$^{1,3, \varheart}$\;\;\; 
Furkan Horoz$^{1,3, \varheart}$\;\;\; 
Umut Isik$^{1,3, \varheart}$\;\;\;
Alp Ozaydin$^{1,3, \varheart}$\;\;\; \\
Altay Topaloglu$^{1,3, \varheart}$\;\;\;
Umutcan Ustundas$^{1,3, \varheart}$\;\;\; 
Alp Yeni$^{1,3, \varheart}$ \;\;\; 
Ersin Soken$^{2,3}$ \;\;\; 
Erol Sahin$^{1,3}$ \;\;\; \\ 
Gokberk Cinbis$^{1,3, \vardiamond}$ \;\;\; 
Sinan Kalkan$^{1,3, \vardiamond}$ \vspace{2mm}\\
$^{1}$Department of Computer Engineering, $^{2}$Department of Aerospace Engineering, $^{3}$ROMER \\
Middle East Technical University, Ankara, Turkey
}

\begin{document}
\maketitle
{\renewcommand{\thefootnote}{}\footnotetext{$^{\varheart}$Equal contribution.}}
{\renewcommand{\thefootnote}{}\footnotetext{$^{\vardiamond}$Equal senior contribution. {\tt\small skalkan@metu.edu.tr}}}

\begin{abstract}
Monocular Depth Estimation (MDE) is crucial for autonomous lunar rover navigation using electro-optical cameras. However, deploying terrestrial MDE networks to the Moon brings a severe domain gap due to harsh shadows, textureless regolith, and zero atmospheric scattering. Existing evaluations rely on analogs that fail to replicate these conditions and lack actual metric ground truth. To address this, we present LuMon, a comprehensive benchmarking framework to evaluate MDE methods for lunar exploration. We introduce novel datasets featuring high-quality stereo ground truth depth from the real \change mission and the \cheri dark analog dataset. Utilizing this framework, we conduct a systematic zero-shot evaluation of state-of-the-art architectures across synthetic, analog, and real datasets. We rigorously assess performance against mission critical challenges like craters, rocks, extreme shading, and varying depth ranges. Furthermore, we establish a sim-to-real domain adaptation baseline by fine tuning a foundation model on synthetic data. While this adaptation yields drastic in-domain performance gains, it exhibits minimal generalization to authentic lunar imagery, highlighting a persistent cross-domain transfer gap. Our extensive analysis reveals the inherent limitations of current networks and sets a standard foundation to guide future advancements in extraterrestrial perception and domain adaptation. The LuMon benchmark, datasets, and code are publicly available at \url{https://metulumon.github.io/}.
\end{abstract}    
\section{Introduction}
\label{sec:intro}

\begin{figure*}[hbt!]
\centering
    \includegraphics[width=0.8\linewidth]{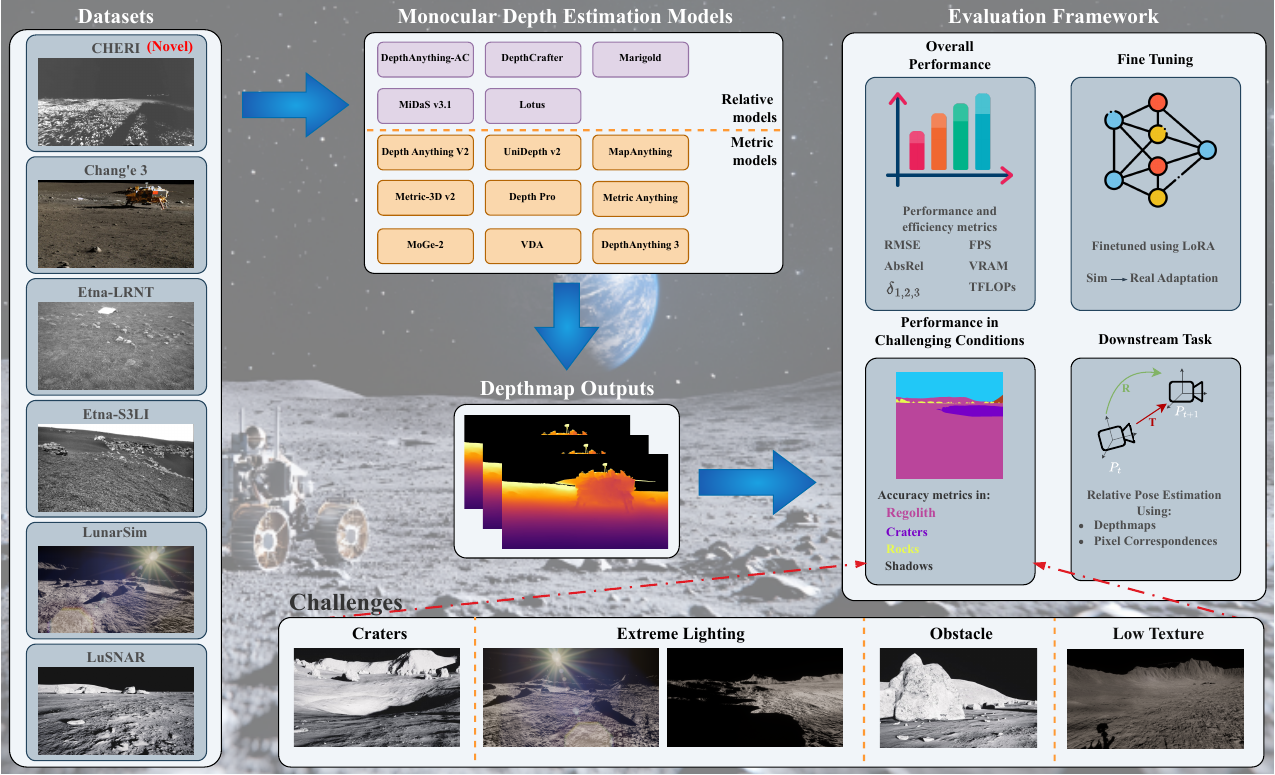}
    \caption[Overview of our evaluation pipeline]{\methodname uses real and synthetic lunar surface datasets. We utilize leading  metric and relative monocular depth estimation models to extract depthmaps, and evaluate their performance in zero-shot lunar settings. We measure the overall performance of the networks in terms of accuracy and efficiency, their accuracy in challenging conditions, and effectiveness in downstream tasks. Finally, we perform fine-tuning to compare it against zero-shot performance.}
    \label{fig:teaser}
\end{figure*}
Electro-optical (EO) cameras offer low mass, high resolution, and low power consumption, making them ideal for mini lunar rovers. Extracting actionable 3D maps from these sensors requires Monocular Depth Estimation (MDE) methods that remain accurate and computationally efficient despite the harsh lighting, low-texture regolith, and unique hazards of the lunar surface.

Although deep-learning MDE networks like \dav{} \cite{yang_depth_2024}, \marigold{} \cite{ke_marigold_2024}, and \depthpro{} \cite{bochkovskii_depth_2024} succeed in terrestrial navigation \cite{xue_2020_toward, kim_2024_instance, zhemg2024monocular}, their lunar feasibility remains unproven. The transition to the Moon introduces a severe visual domain gap. The absence of atmospheric scattering, high-contrast polar shadows, and unique topological textures create environments fundamentally different from Earth-based training datasets.

To address this gap, we present a comprehensive benchmarking and development suite for lunar MDE. As existing terrestrial analogs like Mount Etna \cite{vayugundla2018etnadataset,Giubilato2022S3LI} fail to fully replicate extreme lunar optics, we enable mission-critical validation by constructing high-quality stereo vision based ground truth depth for authentic \change{} \cite{cnsa_ce3_pcam_2020} imagery and introducing the novel \cheri dark-analog dataset.

Our suite \methodname provides five main contributions toward deploying MDE on lunar rovers (see also Fig. \ref{fig:teaser}):
\begin{itemize}
\item \textbf{Framework and Novel Datasets:} We provide a complete MDE evaluation framework (released upon publication), curating six datasets and introducing ground truth depth-maps for \change{} and the \cheri environment.
\item \textbf{Systematic Zero-Shot Analysis:} We evaluate 14 leading models across relative, metric, generative, and video-consistent paradigms, measuring robustness against lunar challenges like craters, rocks, and shading.
\item \textbf{Sim-to-Real Domain Adaptation:} We fine-tune a metric foundation model on synthetic data to expose current generalization gaps in transferring learned features to authentic lunar imagery.
\item \textbf{Downstream Task Integration:} We integrate depth predictions into a relative pose estimation pipeline to quantify how MDE quality directly impacts robotic autonomy.
\item \textbf{Limitations and Open Issues:} We identify inherent network limitations and highlight critical open issues to guide future lunar rover navigation research.
\end{itemize}
\section{Related Work}
\label{sec:relatedwork}

\noindent\textbf{Monocular Depth Estimation (MDE) \& Benchmarks.} While deep learning has significantly advanced MDE for terrestrial autonomy \cite{yang_depth_2024, birkl2023midas, hu_depthcrafter_2024, piccinelli_unidepthv2_2025}, extracting 3D structure from 2D projections remains inherently ill-posed \cite{Rajapaksha_2024}. Existing benchmarks primarily target terrestrial indoor \cite{nyu_depthv2, schoeps2017cvpr} or outdoor driving domains \cite{eigen_depth_2014, make3d}. In lunar environments, this challenge is severely exacerbated by extreme high-contrast lighting, zero atmospheric scattering, textureless regolith, and a lack of large-scale metric ground truth. Our study systematically evaluates SOTA methods under these precise constraints to identify specific success and failure modes.

\vspace{1.8pt} \noindent\textbf{Lunar Depth Estimation \& Environments.} Historical lunar depth estimation relies on orbiter imagery for Digital Terrain Models \cite{chen_elunardtmnet, jia_dispnet, chen_orbiter}. Surface-level evaluations often depend on structurally mismatched bird's-eye data \cite{patil_lunar_2024} or idealized synthetic datasets \cite{Bhaskara_2024, goldperg_planetary, qiu2023dsmnetdeephighprecision3d, xu2023iterativegeometryencodingvolume, liang_disparity_2024} that miss authentic lighting, noise, and terrain complexities. While physical testbeds \cite{ludivig_lunalab, coloma2022enhancing} and private simulators \cite{bingham_dust} exist, they frequently lack synchronized ground truth or public availability. Photorealistic public simulators \cite{lunarsim, liu_lusnarlunar_2024} provide essential dense ground truth but lack physical realism. Real Earth-analog datasets offer physical realism but lack lunar illumination, while some authentic mission datasets suffer from severe sensor misalignment\cite{wong2017polar}. Consequently, a realistic benchmark must integrate cross-domain data, combining controlled simulations \cite{lunarsim, liu_lusnarlunar_2024}, properly ground-truthed Earth analogs \cite{vayugundla_etna_2018, giubilato_s3li_2022}, and carefully calibrated in-situ stereo datasets \cite{cnsa_ce3_pcam_2020}.
\section{\methodname Benchmark Framework}
\label{sec:framework}

\subsection{Objectives}
This study explores the feasibility of existing deep-learning MDE methods for lunar exploration. Our benchmark evaluates state-of-the-art models across four key axes: (1) zero-shot accuracy on synthetic and real lunar imagery; (2) robustness to edge cases like craters, rocks, and low illumination; (3) computational efficiency, including latency, memory, and GFLOPs; and (4) depth reliability across varying distance intervals.

\subsection{Simulation Environments}

We utilize two photorealistic simulation environments to obtain high-precision depth maps under controlled parameters. \textbf{\lunarsim} \cite{lunarsim} is a Unity-based simulator providing stereo imagery and pixel-aligned depth. We captured a custom trajectory sequence, including gt depth-maps and pose data, to evaluate the MDE models. \textbf{\lusnar} \cite{liu_lusnarlunar_2024} features nine Unreal Engine 4 sequences containing stereo images, dense depth, and semantic segmentation masks for five geological classes, enabling our category-specific evaluations.

\subsection{Real \& Analog Environments}

We incorporate diverse real extraterrestrial imagery and planetary analogs to bridge the persistent sim-to-real gap (See \Cref{tab:used_datasets_summary}). Specifically, a primary contribution of our work is the introduction of the novel \textbf{\cheri dataset}, captured in a dark lunar-analog environment, alongside the generation of high-quality ground truth depth for real stereo images from the \textbf{\change} \cite{cnsa_ce3_pcam_2020} mission. Since neither the raw analog images nor the historical mission data include native depth, we construct this ground truth using stereo reconstruction. We refine stereo geometry via sparse ORB \cite{rublee2011orb} feature correspondences to ensure strict epipolar consistency, apply sub-pixel accurate rectification, and generate disparity maps using FoundationStereo \cite{bowen2025foundationstereo} before converting to metric depth. We mask invalid or uncertain pixels and validate scale accuracy against known physical dimensions. For existing analogs, we evaluate the \etna \cite{vayugundla2018etnadataset}, featuring SGM-generated \cite{hirschmuller2008sgm} stereo depth, and the \textbf{\seli dataset} \cite{Giubilato2022S3LI}, which provides LiDAR ground truth.

\begin{table}[t]
    \centering
    \caption{A summary of the datasets used. \textbf{GT}: Ground Truth; \textbf{Seg. Masks}: Segmentation Masks; \textbf{Sim}: Simulation.}
    \label{tab:used_datasets_summary}
    \scriptsize\setlength{\tabcolsep}{2pt}
    \renewcommand{\arraystretch}{1.1}
    \begin{tabularx}{\columnwidth}{l|Y|Y|Y|c|c|c} 
        \hline
        \textbf{Dataset} & \textbf{Domain} & \textbf{Seg. Masks} & \textbf{GT Source} & \textbf{Baseline} & \textbf{Resolutions} & \textbf{Max Depth} \\
        \hline
        \textbf{\lunarsim} & Sim. & \xmark & Sim. & 12 cm & $1280 \times 720$ & - \\
        \textbf{\lusnar} & Sim.& \cmark & Sim. & 31 cm & $1024 \times 1024$ & 50 m \\
        \textbf{\seli} & Analog  & \xmark & LiDAR & 20 cm & $688 \times 512$ & 30 m \\
        \textbf{\etna} & Analog & \xmark & Stereo & 9 cm & $1292 \times 964$ & 15 m \\
        \textbf{\cheri} & Analog & \xmark & Stereo& 11.2 cm & $1280 \times 720$ & 17 m \\
        \textbf{\change} & In-Situ & \xmark & Stereo& 27 cm & $2352 \times 1728$ & 25 m \\
        \hline
    \end{tabularx}
\end{table}

\subsection{Compared Models}

We evaluated 14 models across two paradigms:

\noindent\textbf{Metric Depth Estimators} to test absolute scale retention and include \metricd{} \cite{hu_metric3dv2_2024}, \unidepth{} \cite{piccinelli_unidepthv2_2025}, \depthpro{} \cite{bochkovskii_depth_2024}, \moge{} \cite{wang2025moge2}, \mapanything{} \cite{keetha2025mapanything}, and \metricanything{} \cite{metricanything2026}. We also evaluate the metric configurations of foundation models \davthree{} \cite{depthanything3}, \vda{} \cite{video_depth_anything}, and \dav{} \cite{yang_depth_2024}. 

\noindent\textbf{Relative Depth Estimators} to assess the transferability of scale-invariant pre-training. This includes discriminative models like \midas{} \cite{birkl2023midas} and \daac{} \cite{sun2025depth}, diffusion-based approaches \marigold{} \cite{ke_marigold_2024}, \lotus{} \cite{he_lotus_2025}, and \depthcrafter{} \cite{hu_depthcrafter_2024} to test generative surface detailing.

\subsection{Evaluation Procedure}
\label{sect:evaluation procedure}
Following established literature \cite{obukhov2025fourthmdechallenge}, we evaluate all models under a unified setup. Predictions are resized to ground truth resolution. Inverse depth outputs are mapped to standard depth with numerical stabilization ($\epsilon=1\times10^{-6}$). We mask invalid, non-finite, and out-of-range pixels based on dataset modality. We clip stereo datasets (\cheri, \change, \etna) using KITTI-adapted \cite{Geiger2013kitti} thresholds because disparity quantization errors become highly unreliable at far ranges. We empirically mask extreme distances in \seli and \lusnar to exclude sky and background artifacts. \lunarsim remains unclipped because its relative depth lacks an absolute metric scale. We detail the exact maximum evaluation depths in \Cref{tab:used_datasets_summary}. Finally, we apply a per-image least squares affine alignment uniformly across all architectures, including metric estimators, for a fair comparison. Since metric models learn scale priors primarily from terrestrial features, they exhibit severe global scaling failures when transferred to the lunar domain. This alignment isolates the structural fidelity of the predictions instead of penalizing pure domain-induced scale drift.

\vspace{1.8pt} \noindent\textbf{Performance metrics.} While our evaluation encompasses a comprehensive suite of metrics (such as MAE and SILog), page limits restrict our reported results to the primary metrics: $\delta_1$, AbsRel, and RMSE.
\section{EXPERIMENTS}

\subsection{Experiment Setup and Details} 

Our comprehensive suite encompasses multiple evaluation axes, in addition to a direct overall comparison:

\noindent\textbf{Evaluation Protocol.} To ensure a fair and rigorous comparison, particularly concerning our fine-tuned \dav{} model, all evaluations on the \lusnar{} dataset are strictly conducted on its designated test set. For all other datasets, evaluations are performed across their entirety.

\noindent\textbf{Semantic-aware Analysis.} We use the  segmentation labels from \lusnar to assess performance on distinct geological features, such as regoliths, craters, and rocks.

\noindent\textbf{Shadow Analysis.} We generate masks using dataset-specific pixel brightness thresholds rather than a global value, refining these regions with a 5x5 morphological opening to reduce noise.

\noindent\textbf{Pose Estimation.} We investigate the practical utility of MDE models for downstream applications by estimating pose derived from the predicted depth.  

\noindent\textbf{Domain Adaptation.} We also explore the impact of domain-specific adaptation by finetuning the metric outdoor weights of the \dav model on the \lusnar dataset, quantifying the resulting performance shifts in lunar environments.

\noindent\textbf{Efficiency.} We report speed and memory metrics at a standard $1280 \times 720$ resolution, evaluated on a standardized setup featuring an Intel(R) Xeon(R) Gold 6326 CPU and an NVIDIA RTX A6000 GPU.

\noindent\textbf{MDE Network Details.} We evaluated the highest-performing pre-trained weights. Metric models: \dav (metric-outdoor ViT-L), \vda (metric ViT-L), \davthree (mono-large), \moge (ViT-L normal), \unidepth (v2 ViT-L/14), \metricd (ViT-L), \depthpro (standard), \metricanything (student pointmap), \mapanything (CC). Relative models: \daac (ViT-S), \midas (BEiT$_{512}$-L), \marigold (v1.1), \lotus (v2.0 disparity), \depthcrafter (SVD-xt).

\subsection{Exp. 1: Overall Comparison}
\begin{table*}[t]
    \centering
    \caption{Exp. 1: Evaluation on LunarSim, LuSNAR, Etna, S3LI, Chang'e 3, and Cheri datasets. \textcolor{red}{\textbf{Red}} and \textcolor{blue}{blue} respectively indicate best and second-best results. }
    \label{tab:exp1_overall}
    \setlength{\tabcolsep}{1.2pt}
    \scriptsize
    \renewcommand{\arraystretch}{1.15}
    
    \begin{tabular}{c|l|ccc|ccc|ccc|ccc|ccc|ccc} 
        \hline 
        
        \multirow{3}{*}{\textbf{Type}} & \multirow{3}{*}{{Model}} &
        \multicolumn{3}{c|}{\rule{0pt}{3ex}{\lunarsim}} &
        \multicolumn{3}{c|}{{\lusnar}} &
        \multicolumn{3}{c|}{{\etna}} &
        \multicolumn{3}{c|}{{\seli}} &
        \multicolumn{3}{c|}{{\change}} &
        \multicolumn{3}{c}{{\cheri}} 
        \\
        
        \cline{3-20}
        \cline{3-20}
        
        & & \rule[-1.5ex]{0pt}{4ex} $\delta_{1}\uparrow$ & A.Rel$\downarrow$ & RMSE$\downarrow$
        & $\delta_{1}\uparrow$ & A.Rel$\downarrow$ & RMSE$\downarrow$
        & $\delta_{1}\uparrow$ & A.Rel$\downarrow$ & RMSE$\downarrow$
        & $\delta_{1}\uparrow$ & A.Rel$\downarrow$ & RMSE$\downarrow$ 
        & $\delta_{1}\uparrow$ & A.Rel$\downarrow$ & RMSE$\downarrow$
        & $\delta_{1}\uparrow$ & A.Rel$\downarrow$ & RMSE$\downarrow$ \\
        \hline 
        
        \multirow{10}{*}{\rotatebox[origin=c]{90}{\textbf{Metric}}} 
        & \davft{} (Ours) & \textcolor{blue}{0.84} & \textcolor{blue}{0.13} & 13.96 & \textcolor{red}{\textbf{0.93}} & \textcolor{red}{\textbf{0.07}} & \textcolor{red}{\textbf{1.07}} & \textcolor{red}{\textbf{1.00}} & 0.05 & 0.13 & 0.84 & 0.13 & 1.30 & 0.92 & 0.08 & 0.70 & 0.36 & 0.50 & 3.90 \\
        \cline{2-20}
        & \dav{} \cite{yang_depth_2024} & 0.56 & 0.26 & 15.18 & 0.31 & 0.45 & 2.89 & \textcolor{red}{\textbf{1.00}} & 0.06 & 0.15 & 0.88 & 0.12 & 1.22 & \textcolor{blue}{0.95} & \textcolor{blue}{0.06} & 0.56 & 0.32 & 0.54 & 4.14 \\
        & \davthree{} \cite{depthanything3} & 0.63 & 0.22 & 14.15 & 0.53 & 0.29 & 1.99 & \textcolor{red}{\textbf{1.00}} & \textcolor{blue}{0.04} & 0.11 & \textcolor{red}{\textbf{0.91}} & \textcolor{red}{\textbf{0.10}} & 1.12 & \textcolor{red}{\textbf{0.96}} & \textcolor{red}{\textbf{0.05}} & \textcolor{blue}{0.44} & 0.34 & 0.51 & 3.97 \\
        & \moge{} \cite{wang2025moge2} & 0.77 & 0.15 & 12.17 & 0.53 & 0.29 & 1.80 & \textcolor{blue}{0.99} & 0.05 & 0.15 & \textcolor{red}{\textbf{0.91}} & \textcolor{red}{\textbf{0.10}} & \textcolor{blue}{1.08} & \textcolor{blue}{0.95} & \textcolor{blue}{0.06} & 0.45 & 0.32 & 0.54 & 4.12 \\
        & \depthpro{} \cite{bochkovskii_depth_2024} & 0.50 & 0.29 & 17.57 & 0.33 & 0.45 & 2.70 & 0.73 & 241.71 & 450.05 & \textcolor{blue}{0.89} & \textcolor{blue}{0.11} & 1.11 & 0.89 & 0.10 & 0.89 & 0.32 & 0.55 & 4.18 \\
        & \metricd{} \cite{hu_metric3dv2_2024} & 0.75 & 0.16 & 12.53 & 0.61 & 0.25 & 1.71 & \textcolor{red}{\textbf{1.00}} & \textcolor{red}{\textbf{0.03}} & 0.10 & 0.87 & 0.12 & 1.19 & \textcolor{blue}{0.95} & \textcolor{red}{\textbf{0.05}} & \textcolor{red}{\textbf{0.43}} & 0.33 & 0.53 & 4.09 \\ 
        & \unidepth{} \cite{piccinelli_unidepthv2_2025} & 0.76 & 0.21 & 16.16 & 0.63 & 0.26 & 2.14 & \textcolor{red}{\textbf{1.00}} & \textcolor{red}{\textbf{0.03}} & \textcolor{red}{\textbf{0.07}} & 0.87 & 0.12 & 1.20 & 0.94 & 0.07 & 0.54 & 0.35 & 0.51 & 3.99 \\
        & \mapanything{} \cite{keetha2025mapanything} & \textcolor{red}{\textbf{0.86}} & \textcolor{red}{\textbf{0.11}} & \textcolor{red}{\textbf{10.73}} & 0.74 & 0.17 & \textcolor{blue}{1.53} & \textcolor{red}{\textbf{1.00}} & \textcolor{red}{\textbf{0.03}} & \textcolor{blue}{0.08} & \textcolor{red}{\textbf{0.91}} & \textcolor{red}{\textbf{0.10}} & \textcolor{red}{\textbf{1.05}} & \textcolor{blue}{0.95} & 0.07 & 0.52 & 0.36 & 0.50 & 3.91 \\
        & \vda{} \cite{video_depth_anything} & 0.75 & 0.16 & 13.94 & 0.77 & 0.17 & 2.05 & \textcolor{red}{\textbf{1.00}} & \textcolor{blue}{0.04} & 0.12 & 0.83 & 0.14 & 1.34 & 0.93 & 0.08 & 0.68 & \textcolor{blue}{0.43} & \textcolor{blue}{0.44} & \textcolor{blue}{3.56} \\
        & \metricanything{} \cite{metricanything2026} & 0.79 & 0.15 & \textcolor{blue}{12.08} & 0.51 & 0.30 & 1.83 & 0.98 & 0.07 & 0.22 & \textcolor{red}{\textbf{0.91}} & \textcolor{red}{\textbf{0.10}} & \textcolor{red}{\textbf{1.05}} & \textcolor{blue}{0.95} & \textcolor{blue}{0.06} & \textcolor{blue}{0.44} & 0.32 & 0.55 & 4.17 \\
        \hline 
        
        \multirow{5}{*}{\rotatebox[origin=c]{90}{\textbf{Relative}}}
        & \daac{} \cite{sun2025depth} & 0.70 & 0.28 & 19.12 & 0.70 & 0.26 & 2.63 & \textcolor{red}{\textbf{1.00}} & 0.05 & 0.12 & 0.87 & 0.12 & 1.19 & 0.66 & 0.22 & 1.55 & \textcolor{red}{\textbf{0.55}} & \textcolor{red}{\textbf{0.35}} & \textcolor{red}{\textbf{3.09}} \\
        & \depthcrafter{} \cite{hu_depthcrafter_2024} & 0.40 & 0.37 & 20.28 & 0.41 & 0.47 & 2.96 & 0.92 & 0.10 & 0.28 & 0.80 & 0.15 & 1.44 & 0.87 & 0.13 & 1.05 & 0.33 & 0.53 & 4.08 \\
        & \lotus{} \cite{he_lotus_2025} & 0.59 & 0.27 & 20.43 & 0.36 & 0.44 & 3.08 & 0.65 & 0.19 & 0.49 & 0.80 & 0.15 & 1.45 & 0.70 & 0.19 & 1.64 & 0.35 & 0.49 & 3.90 \\
        & \midas{} \cite{birkl2023midas} & 0.78 & 0.17 & 14.97 & \textcolor{blue}{0.81} & \textcolor{blue}{0.13} & 1.68 & \textcolor{red}{\textbf{1.00}} & \textcolor{blue}{0.04} & 0.12 & \textcolor{blue}{0.89} & \textcolor{blue}{0.11} & 1.15 & 0.81 & 0.15 & 1.12 & 0.40 & 0.46 & 3.71 \\
        & \marigold{} \cite{ke_marigold_2024} & 0.67 & 0.21 & 17.25 & 0.62 & 0.24 & 2.02 & 0.94 & 0.10 & 0.27 & 0.86 & 0.12 & 1.26 & 0.91 & 0.09 & 0.69 & 0.38 & 0.47 & 3.75 \\
        \hline 
    \end{tabular}
\end{table*}
\begin{table}[hbt!]
\centering
\caption{Efficiency metric results. \textcolor{red}{\textbf{Red}} indicates the best result and \textcolor{blue}{blue} indicates the second best.}
\label{tab:efficiency_metrics}
\scriptsize
    \setlength{\tabcolsep}{1.5pt} 
    \renewcommand{\arraystretch}{1.05}
    \resizebox{0.8\columnwidth}{!}
    {
\begin{tabular}{c|l|ccc}
\hline
\textbf{Type} & \textbf{Model} & \textbf{FPS $\uparrow$} & \textbf{VRAM (GB) $\downarrow$} & \textbf{TFLOPs $\downarrow$} \\
\hline

\multirow{10}{*}{\rotatebox[origin=c]{90}{\textbf{Metric}}} 
& \davft{} (Ours) & 4.64 & \textcolor{blue}{1.926} & 1.309 \\
\cline{2-5}
& \dav{} & 4.54 & \textcolor{blue}{1.926} & 1.309 \\
& \davthree{} & \textcolor{blue}{24.65} & 2.007 & \textcolor{blue}{0.658} \\
& \moge{} & 2.63 & 2.234 & 2.150 \\
& \depthpro{} & 0.91 & 7.949 & 9.677 \\
& \metricd{} & 2.47 & 2.990 & 2.870 \\
& \unidepth{} & 12.39 & 3.126 & 1.868 \\
& \mapanything{} & 13.22 & 7.080 & 1.101 \\
& \vda{} & 0.51 & 10.385 & 75.444 \\
& \metricanything{} & 11.04 & 2.832 & 2.150 \\
\hline

\multirow{5}{*}{\rotatebox[origin=c]{90}{\textbf{Relative}}}
& \daac{} & \textcolor{red}{\textbf{34.24}} & \textcolor{red}{\textbf{0.509}} & \textcolor{red}{\textbf{0.130}} \\
& \depthcrafter{} & 0.01 & 20.166 & 18.472 \\
& \lotus{} & 5.83 & 3.510 & 2.851 \\
& \midas{} & 3.46 & 2.135 & 0.920 \\
& \marigold{} & 0.02 & 6.117 & 272.109 \\
\hline
\end{tabular}
}
\end{table}

In this experiment, we evaluate the zero-shot performances of the selected MDE networks alongside our domain-adapted baseline across all six datasets (Table \ref{tab:exp1_overall}). Rather than listing isolated dataset metrics, our analysis reveals two critical generalization trends.

\noindent\textbf{Metric Foundation Model Dominance.} Metric depth estimators consistently outrank relative models across the benchmarking suite. Architectures like \mapanything{}, \vda{}, and \davthree{} exhibit the most robust cross-domain zero-shot performance. While no single network dominates synthetic datasets, metric foundation models maintain superior geometric consistency and structural fidelity on authentic extraterrestrial imagery like the \change{} dataset. Because our affine-aligned protocol controls for absolute scale recovery, this advantage strictly reflects stronger learned geometric priors. Additionally, our fine-tuned \davft{} baseline establishes an upper bound for synthetic environments, achieving state-of-the-art accuracy on its training domain (\lusnar{}) with excellent sim-to-sim transfer on \lunarsim{}.

\noindent\textbf{The Earth-Analog Illusion.} Traditional Earth-based analogs often prove deceptive for benchmarking. Most models achieve near-perfect accuracy on the well-lit \etna{} dataset because its volcanic terrain closely mirrors standard terrestrial training distributions. However, performance universally plummets on \seli{} due to stricter LiDAR ground truths, and on the \cheri{} dataset, where harsh, high-contrast polar lighting confounds all architectures.

\begin{myframe}{Main Result}
Metric foundation models provide the strongest zero-shot baseline for lunar topography, yet all architectures degrade severely under authentic high-contrast lighting. Near-perfect results on well-lit Earth analogs prove deceptive, confirming that current terrestrial proxies are insufficient for evaluating mission-critical extraterrestrial deployment.
\end{myframe}

\subsection{Exp. 2: Challenging Surface Constructs}
\begin{table}[t] 
    \centering
    \caption{Exp. 2: Results on \textbf{semantically} different regions on LuSNAR dataset (Regolith, Rock and Craters). \textcolor{red}{\textbf{Red}} indicates the best result and \textcolor{blue}{blue} indicates the second best.}
    \label{tab:exp2_semantic}
    \setlength{\tabcolsep}{1.5pt} 
    \renewcommand{\arraystretch}{1.05}
    \resizebox{0.8\columnwidth}{!}{%
    \begin{tabular}{c|l| cc cc cc} 
        \hline
        \multirow{2}{*}{\textbf{Type}} &
        \multirow{2}{*}{\textbf{Model}} &
        \multicolumn{2}{c}{\textbf{Regolith}} & 
        \multicolumn{2}{c}{\textbf{Rocks}} & 
        \multicolumn{2}{c}{\textbf{Craters}} \\
        \cline{3-8}
        & & $\delta_{1}\uparrow$ & A.Rel$\downarrow$ 
        & $\delta_{1}\uparrow$ & A.Rel$\downarrow$ 
        & $\delta_{1}\uparrow$ & A.Rel$\downarrow$ \\
        \hline
        
        \multirow{10}{*}{\rotatebox[origin=c]{90}{\textbf{Metric}}} 
        & \davft{} (Ours) & \textcolor{red}{\textbf{0.93}} & \textcolor{red}{\textbf{0.07}} & \textcolor{red}{\textbf{0.92}} & \textcolor{red}{\textbf{0.09}} & \textcolor{red}{\textbf{0.87}} & \textcolor{red}{\textbf{0.10}} \\
        \cline{2-8}
        & \dav{} \cite{yang_depth_2024}                  & 0.30 & 0.46 & 0.44 & 0.28 & 0.23 & 0.37 \\
        & \davthree{} \cite{depthanything3}              & 0.52 & 0.29 & 0.65 & 0.19 & 0.37 & 0.27 \\
        & \moge{} \cite{wang2025moge2}                   & 0.52 & 0.30 & 0.69 & 0.18 & 0.40 & 0.27 \\
        & \depthpro{} \cite{bochkovskii_depth_2024}      & 0.32 & 0.47 & 0.45 & 0.28 & 0.23 & 0.37 \\
        & \metricd{} \cite{hu_metric3dv2_2024}           & 0.60 & 0.26 & 0.80 & 0.15 & 0.41 & 0.27 \\ 
        & \unidepth{} \cite{piccinelli_unidepthv2_2025}  & 0.62 & 0.27 & 0.76 & 0.18 & \textcolor{blue}{0.45} & \textcolor{blue}{0.25} \\
        & \mapanything{} \cite{keetha2025mapanything}    & 0.73 & 0.17 & \textcolor{blue}{0.85} & \textcolor{blue}{0.14} & 0.42 & 0.26 \\
        & \vda{} \cite{video_depth_anything}             & 0.77 & 0.17 & 0.80 & 0.16 & 0.36 & 0.29 \\
        & \metricanything{} \cite{metricanything2026}    & 0.50 & 0.31 & 0.67 & 0.18 & 0.41 & 0.27 \\
        \hline
        
        \multirow{5}{*}{\rotatebox[origin=c]{90}{\textbf{Relative}}}
        & \daac{} \cite{sun2025depth}                    & 0.69 & 0.27 & 0.66 & 0.24 & 0.31 & 0.31 \\
        & \depthcrafter{} \cite{hu_depthcrafter_2024}    & 0.40 & 0.48 & 0.43 & 0.27 & 0.17 & 0.41 \\
        & \lotus{} \cite{he_lotus_2025}                  & 0.35 & 0.46 & 0.51 & 0.28 & 0.21 & 0.39 \\
        & \midas{} \cite{birkl2023midas}                 & \textcolor{blue}{0.81} & \textcolor{blue}{0.13} & 0.76 & 0.19 & 0.34 & 0.28 \\
        & \marigold{} \cite{ke_marigold_2024}            & 0.62 & 0.24 & 0.66 & 0.21 & 0.26 & 0.33 \\
        \hline 
    \end{tabular}%
    }
\end{table}
\begin{table*}[t]
    \centering
    \caption{Exp. 2: Performances on shaded regions. \textcolor{red}{\textbf{Red}} indicates the best result and \textcolor{blue}{blue} indicates the second best.}
    \label{tab:exp2_shaded}
    \setlength{\tabcolsep}{1.8pt}
    \scriptsize
    \renewcommand{\arraystretch}{1.1}
    
    \begin{tabular}{c|l|ccc|ccc|ccc|ccc|ccc|ccc} 
        \hline 
        
        \multirow{3}{*}{\textbf{Type}} & \multirow{3}{*}{{Model}} &
        \multicolumn{3}{c|}{\rule{0pt}{3ex}{\lunarsim}} &
        \multicolumn{3}{c|}{{\lusnar}} &
        \multicolumn{3}{c|}{{\etna}} &
        \multicolumn{3}{c|}{{\seli}} &
        \multicolumn{3}{c|}{{\change}} &
        \multicolumn{3}{c}{{\cheri}} 
        \\
        
        \cline{3-20}
        \cline{3-20}
        
        & & \rule[-1.5ex]{0pt}{4ex} $\delta_{1}\uparrow$ & A.Rel$\downarrow$ & RMSE$\downarrow$
        & $\delta_{1}\uparrow$ & A.Rel$\downarrow$ & RMSE$\downarrow$
        & $\delta_{1}\uparrow$ & A.Rel$\downarrow$ & RMSE$\downarrow$
        & $\delta_{1}\uparrow$ & A.Rel$\downarrow$ & RMSE$\downarrow$ 
        & $\delta_{1}\uparrow$ & A.Rel$\downarrow$ & RMSE$\downarrow$
        & $\delta_{1}\uparrow$ & A.Rel$\downarrow$ & RMSE$\downarrow$ \\
        \hline 
        
        \multirow{10}{*}{\rotatebox[origin=c]{90}{\textbf{Metric}}} 
        & \davft{} (Ours) & \textcolor{blue}{0.79} & \textcolor{blue}{0.15} & 15.10 & \textcolor{red}{\textbf{0.84}} & \textcolor{red}{\textbf{0.11}} & \textcolor{red}{\textbf{0.95}} & \textcolor{red}{\textbf{1.00}} & 0.06 & 0.10 & 0.46 & 0.29 & 2.88 & 0.89 & 0.10 & 0.73 & 0.16 & \textcolor{blue}{0.51} & 4.76 \\
        \cline{2-20}
        & \dav{} \cite{yang_depth_2024} & 0.47 & 0.29 & 16.39 & 0.20 & 0.66 & 2.86 & \textcolor{blue}{0.99} & 0.07 & 0.12 & 0.51 & 0.33 & 2.85 & \textcolor{red}{\textbf{0.94}} & \textcolor{blue}{0.07} & 0.59 & 0.15 & 0.72 & 5.18 \\
        & \davthree{} \cite{depthanything3} & 0.54 & 0.26 & 15.77 & 0.34 & 0.40 & 2.15 & \textcolor{red}{\textbf{1.00}} & \textcolor{blue}{0.05} & {0.07} & 0.53 & 0.46 & 3.58 & \textcolor{red}{\textbf{0.94}} & \textcolor{red}{\textbf{0.06}} & \textcolor{blue}{0.45} & 0.18 & 0.62 & 4.75 \\
        & \moge{} \cite{wang2025moge2} & 0.68 & 0.18 & 14.42 & 0.35 & 0.41 & 1.90 & \textcolor{blue}{0.99} & \textcolor{blue}{0.05} & 0.08 & 0.56 & 0.38 & 3.22 & \textcolor{blue}{0.93} & \textcolor{blue}{0.07} & 0.46 & 0.15 & 0.73 & 5.20 \\
        & \depthpro{} \cite{bochkovskii_depth_2024} & 0.38 & 0.35 & 21.33 & 0.23 & 0.68 & 2.83 & 0.39 & 559.37 & 818.97 & 0.59 & 0.25 & 2.36 & 0.85 & 0.11 & 0.87 & 0.13 & 0.78 & 5.27 \\
        & \metricd{} \cite{hu_metric3dv2_2024} & 0.62 & 0.20 & 14.35 & 0.39 & 0.38 & 1.92 & \textcolor{red}{\textbf{1.00}} & \textcolor{blue}{0.05} & {0.07} & 0.68 & 0.23 & 2.35 & \textcolor{red}{\textbf{0.94}} & \textcolor{red}{\textbf{0.06}} & \textcolor{red}{\textbf{0.44}} & 0.16 & 0.68 & 5.06 \\ 
        & \unidepth{} \cite{piccinelli_unidepthv2_2025} & 0.70 & 0.23 & 24.12 & 0.45 & 0.36 & 2.85 & \textcolor{red}{\textbf{1.00}} & \textcolor{red}{\textbf{0.03}} & \textcolor{blue}{0.05} & 0.75 & \textcolor{blue}{0.17} & 1.90 & 0.92 & 0.08 & 0.54 & 0.17 & 0.64 & 4.94 \\
        & \mapanything{} \cite{keetha2025mapanything} & \textcolor{red}{\textbf{0.80}} & \textcolor{red}{\textbf{0.13}} & \textcolor{red}{\textbf{11.48}} & 0.60 & 0.22 & \textcolor{blue}{1.64} & \textcolor{red}{\textbf{1.00}} & \textcolor{red}{\textbf{0.03}} & \textcolor{red}{\textbf{0.04}} & \textcolor{red}{\textbf{0.80}} & \textcolor{red}{\textbf{0.15}} & \textcolor{red}{\textbf{1.63}} & \textcolor{blue}{0.93} & 0.08 & 0.55 & 0.15 & 0.63 & 4.95 \\
        & \vda{} \cite{video_depth_anything} & 0.66 & 0.19 & 16.31 & \textcolor{blue}{0.65} & 0.23 & 2.19 & \textcolor{red}{\textbf{1.00}} & 0.06 & 0.09 & 0.48 & 0.28 & 2.61 & 0.90 & 0.10 & 0.68 & \textcolor{blue}{0.27} & \textcolor{blue}{0.51} & \textcolor{blue}{4.23} \\
        & \metricanything{} \cite{metricanything2026} & 0.74 & 0.16 & \textcolor{blue}{13.19} & 0.32 & 0.42 & 2.00 & \textcolor{blue}{0.99} & 0.07 & 0.12 & 0.68 & 0.23 & 2.22 & \textcolor{red}{\textbf{0.94}} & \textcolor{red}{\textbf{0.06}} & 0.46 & 0.14 & 0.74 & 5.24 \\
        \hline 
        
        \multirow{5}{*}{\rotatebox[origin=c]{90}{\textbf{Relative}}}
        & \daac{} \cite{sun2025depth} & 0.63 & 0.33 & 27.04 & 0.54 & 0.38 & 3.80 & \textcolor{red}{\textbf{1.00}} & 0.09 & 0.13 & 0.62 & 0.20 & 2.13 & 0.58 & 0.26 & 1.68 & \textcolor{red}{\textbf{0.47}} & \textcolor{red}{\textbf{0.41}} & \textcolor{red}{\textbf{3.96}} \\
        & \depthcrafter{} \cite{hu_depthcrafter_2024} & 0.37 & 0.45 & 22.22 & 0.28 & 0.81 & 3.17 & 0.89 & 0.11 & 0.18 & 0.46 & 0.38 & 3.33 & 0.83 & 0.14 & 1.11 & 0.21 & 0.76 & 5.01 \\
        & \lotus{} \cite{he_lotus_2025} & 0.46 & 0.32 & 29.36 & 0.28 & 0.64 & 3.94 & 0.45 & 0.29 & 0.42 & 0.48 & 0.32 & 3.17 & 0.63 & 0.23 & 1.81 & 0.21 & 0.58 & 4.71 \\
        & \midas{} \cite{birkl2023midas} & 0.73 & 0.19 & 16.70 & 0.64 & \textcolor{blue}{0.20} & 1.99 & \textcolor{blue}{0.99} & 0.07 & 0.10 & \textcolor{blue}{0.79} & \textcolor{red}{\textbf{0.15}} & \textcolor{blue}{1.75} & 0.74 & 0.18 & 1.17 & \textcolor{blue}{0.27} & 0.55 & \textcolor{blue}{4.23} \\
        & \marigold{} \cite{ke_marigold_2024} & 0.51 & 0.29 & 21.92 & 0.43 & 0.37 & 2.40 & 0.90 & 0.11 & 0.17 & 0.49 & 0.35 & 3.21 & 0.89 & 0.10 & 0.72 & 0.18 & \textcolor{blue}{0.51} & 4.66 \\
        \hline 
    \end{tabular}
\end{table*}
\begin{table}[t]
\centering
\caption{Exp. 3: Evaluation of Absolute Relative Error (A.Rel $\downarrow$) on all datasets except \lunarsim across Near and Far distances. \textcolor{red}{\textbf{Red}}: Best, \textcolor{blue}{blue}: Second-best.}
\label{tab:exp3_combined_arel}
\setlength{\tabcolsep}{1.5pt}
\scriptsize
\renewcommand{\arraystretch}{1.1}
\resizebox{\columnwidth}{!}{%
\begin{tabular}{c|l| ccccc | @{\hskip 3pt} ccccc}
\hline
 &  &
\multicolumn{5}{c| @{\hskip 3pt}}{\textbf{Near}} &
\multicolumn{5}{c}{\textbf{Far}} \\
\cline{3-7} \cline{8-12} 
\textbf{Type} & \textbf{Model} & \rotatebox{90}{\lusnar} & \rotatebox{90}{\etna} & \rotatebox{90}{\seli} & \rotatebox{90}{\change} & \rotatebox{90}{\cheri}
& \rotatebox{90}{\lusnar} & \rotatebox{90}{\etna} & \rotatebox{90}{\seli} & \rotatebox{90}{\change} & \rotatebox{90}{\cheri} \\
\hline

\multirow{10}{*}{\rotatebox[origin=c]{90}{\textbf{Metric}}} 
& Dav2-FT (Ours) & \textcolor{red}{\textbf{0.07}} & 0.05 & 0.18 & 0.34 & 1.55 & \textcolor{red}{\textbf{0.09}} & 0.20 & 0.29 & 0.15 & 0.23 \\
\cline{2-12}
& \dav{} \cite{yang_depth_2024} & 0.49 & 0.06 & 0.15 & \textcolor{red}{\textbf{0.18}} & 1.74 & 0.20 & 0.30 & 0.29 & 0.16 & 0.25 \\
& \davthree{} \cite{depthanything3} & 0.31 & \textcolor{blue}{0.04} & \textcolor{blue}{0.14} & \textcolor{blue}{0.20} & 1.58 & 0.16 & 0.29 & 0.27 & \textcolor{blue}{0.14} & 0.24 \\
& \moge{} \cite{wang2025moge2} & 0.32 & 0.05 & \textcolor{red}{\textbf{0.13}} & 0.22 & 1.72 & 0.13 & 0.56 & \textcolor{blue}{0.26} & \textcolor{blue}{0.14} & 0.25 \\
& \depthpro{} \cite{bochkovskii_depth_2024} & 0.50 & 241.77 & \textcolor{red}{\textbf{0.13}} & 0.34 & 1.77 & 0.18 & 0.69 & 0.27 & 0.20 & 0.25 \\
& \metricd{} \cite{hu_metric3dv2_2024} & 0.28 & \textcolor{blue}{0.04} & 0.17 & \textcolor{blue}{0.20} & 1.71 & 0.14 & 0.26 & \textcolor{blue}{0.26} & \textcolor{red}{\textbf{0.13}} & 0.25 \\
& \unidepth{} \cite{piccinelli_unidepthv2_2025} & 0.30 & \textcolor{red}{\textbf{0.03}} & 0.18 & 0.24 & 1.61 & 0.15 & 0.12 & 0.28 & \textcolor{blue}{0.14} & 0.24 \\
& \mapanything{} \cite{keetha2025mapanything} & 0.18 & \textcolor{red}{\textbf{0.03}} & \textcolor{red}{\textbf{0.13}} & 0.25 & 1.59 & \textcolor{blue}{0.12} & 0.25 & \textcolor{red}{\textbf{0.25}} & 0.15 & 0.23 \\
& \vda{} \cite{video_depth_anything} & 0.19 & \textcolor{blue}{0.04} & 0.19 & 0.26 & \textcolor{blue}{1.28} & 0.16 & \textcolor{blue}{0.11} & 0.29 & 0.16 & \textcolor{blue}{0.21} \\
& \metricanything{} \cite{metricanything2026} & 0.33 & 0.07 & \textcolor{red}{\textbf{0.13}} & 0.21 & 1.77 & 0.13 & 0.46 & \textcolor{blue}{0.26} & \textcolor{blue}{0.14} & 0.25 \\
\hline

\multirow{5}{*}{\rotatebox[origin=c]{90}{\textbf{Relative}}}
& \daac{} \cite{sun2025depth} & 0.31 & 0.05 & 0.16 & 0.57 & \textcolor{red}{\textbf{0.93}} & 0.21 & \textcolor{red}{\textbf{0.06}} & 0.27 & 0.26 & \textcolor{red}{\textbf{0.17}} \\
& \depthcrafter{} \cite{hu_depthcrafter_2024} & 0.52 & 0.10 & 0.21 & 0.22 & 1.68 & 0.21 & 0.48 & 0.34 & 0.21 & 0.25 \\
& \lotus{} \cite{he_lotus_2025} & 0.53 & 0.19 & 0.21 & 0.49 & 1.54 & 0.25 & 0.39 & 0.33 & 0.29 & 0.24 \\
& \midas{} \cite{birkl2023midas} & \textcolor{blue}{0.14} & \textcolor{blue}{0.04} & 0.15 & 0.45 & 1.38 & 0.14 & 0.17 & \textcolor{red}{\textbf{0.25}} & 0.19 & 0.22 \\
& \marigold{} \cite{ke_marigold_2024} & 0.25 & 0.10 & 0.16 & 0.29 & 1.44 & 0.16 & 0.51 & 0.30 & 0.17 & 0.23 \\
\hline
\end{tabular}%
}
\end{table}

We evaluate model robustness against specific lighting and topological edge cases. We apply brightness-based shadow masks across all datasets to analyze shading impacts (Table~\ref{tab:exp2_shaded}) and leverage \lusnar segmentation labels to isolate performance on regolith, rocks, and craters (Table~\ref{tab:exp2_semantic}). We refer readers to Section~\ref{sec:supp_semantic} of the supplementary material for our complete analysis.

\noindent\textbf{Shaded Regions.} Shading causes less overall degradation than complex surface geometry. Our fine-tuned baseline, \davft{}, demonstrates exceptional robustness on synthetic benchmarks like \lusnar{}, while \mapanything{} leads on \lunarsim{} and \seli{}. For authentic extraterrestrial imagery (\change{}), metric foundation models like \dav{}, \davthree{}, and \metricd{} deliver the highest accuracy. Notably, \depthpro{} exhibits a catastrophic failure within the \etna{} shaded split.

\noindent\textbf{Topological Features.} Our adapted \davft{} pipeline comprehensively dominates all topological edge cases, showcasing excellent recovery of regolith, rocks, and craters. Among zero-shot architectures, while \midas{} and \vda{} perform reasonably well on flat regolith, metric models like \mapanything{} and \metricd{} show a distinct advantage in resolving rock obstacles. Craters cause severe performance degradation across all zero-shot architectures, although metric models remain comparatively more resilient. Ultimately, craters and rocks pose a significantly greater hurdle for current networks than extreme shading.

\begin{myframe}{Main Result}
Complex surface geometries like craters and rocks degrade zero-shot depth estimation accuracy significantly more than severe lunar shading. Metric models consistently outperform relative architectures across these edge cases, and our domain-adapted \davft{} demonstrates that targeted fine-tuning can successfully overcome these topological and lighting hazards within simulated environments.
\end{myframe}

\subsection{Exp. 3: Performance vs. Distance}
\label{sect:distance_vs_performance}

In this experiment, we analyze how depth estimation performance changes with distance to evaluate global metric consistency (Table \ref{tab:exp3_combined_arel}). We apply a global least-squares affine alignment, which is heavily influenced by the dominant median depth range. We therefore treat this mid-range as an implicit anchor for testing \textit{internal linearity}. Because this alignment inherently stabilizes central distances, distance-dependent scale drift becomes most pronounced at the depth extremes. Consequently, our main evaluation isolates the unanchored Near and Far boundary regions to specifically highlight these non-linear distortions (complete mid-range analysis is provided in Suppl. Sec.~\ref{sec:supp_distance}).

Using this setup, we observe that relative and diffusion-based models (\lotus, \depthcrafter) exhibit strong non-linear distortions, failing to transfer scale consistently into the boundary regions. In contrast, metric models (\mapanything, \metricd) and structurally constrained methods (\vda, \midas) maintain significantly lower and more uniform errors, better preserving the internal linearity vital for stable 3D reconstruction. However, near-range performance degrades sharply on authentic \change and \cheri imagery across most architectures, revealing boundary cases where linear scaling assumptions and domain-specific priors break down.

\begin{myframe}{Main Result}
Metric and structurally constrained models preserve the internal linearity required for stable 3D geometry, overcoming the distance-dependent scale drift prevalent in relative and diffusion methods.
\end{myframe}

\subsection{Exp. 4: Training Regime and Data Analysis}
\label{subsec:exp4}

\begin{figure}[t]
    \centering
    \begin{subfigure}[b]{0.60\linewidth}
        \centering
        \includegraphics[width=\linewidth]{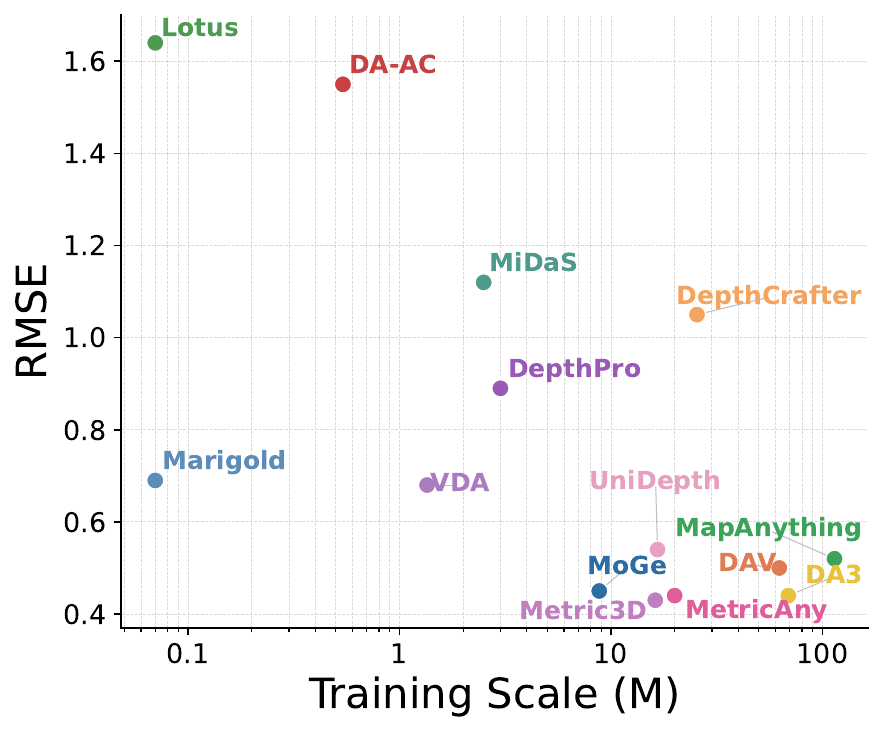}
        \caption{Training Scale vs. RMSE}
        \label{fig:scale_rmse}
    \end{subfigure}
    \hfill
    \begin{subfigure}[b]{0.39\linewidth}
        \centering
        \includegraphics[width=\linewidth]{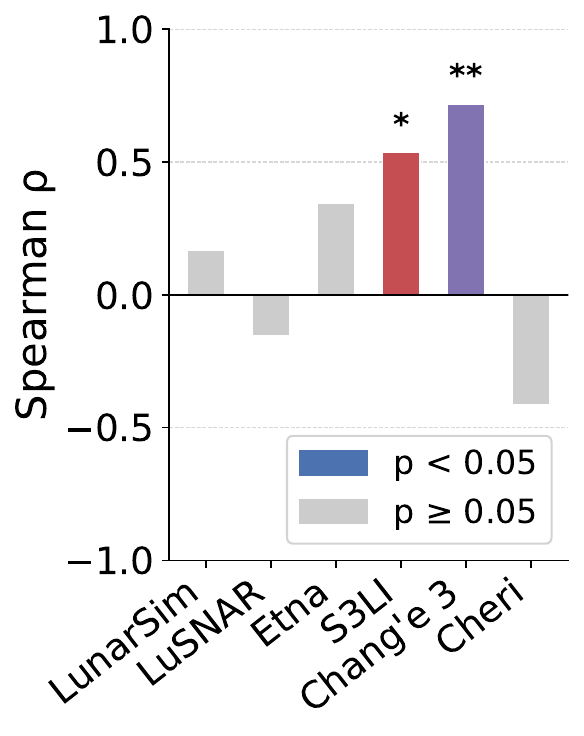}
        \caption{Training Scale vs. $\delta_1$}
        \label{fig:scale_delta}
    \end{subfigure}
    
    \caption{Effect of training dataset scale for one dataset. (a) Scale vs. RMSE. (b) Spearman correlation of scale vs. $\delta_1$ ($*$: $p < 0.05$, $**$: $p < 0.01$).}
    \label{fig:combined_plots}
\end{figure}

We investigate zero-shot generalization factors using a Spearman rank correlation ($\rho$, \cref{fig:combined_plots}) between performance metrics and three quantitative training attributes (\cref{tab:data_comparison_citations}): Training Scale (total image count), Real Data Ratio, and Metric Supervision Ratio. We refer readers to Section~\ref{sec:training_regime} of the supplementary material for our complete analysis.

\begin{table}[hbt!]
  \centering
  \caption{Exp 4: Comparison of Training Data Scales and Composition. \textbf{Scale:} Total data size. \textbf{Real \%:} Percentage of real data. \textbf{MetSup:} Percentage of data with metric supervision.}
  \label{tab:data_comparison_citations}
  
  \renewcommand{\arraystretch}{1.1}
  \setlength{\tabcolsep}{2pt}
  
  \resizebox{\columnwidth}{!}{
  \begin{tabular}{c|l|c|c|c|l}
    \hline 
     
    \rule[-2ex]{0pt}{5.5ex} \textbf{Type} & \textbf{Model} & \textbf{Scale} & \textbf{Real\%} & \textbf{\shortstack{MetSup}} & \textbf{Dominant Scene} \\
     
    \hline 
     
    \multirow{10}{*}{\rotatebox[origin=c]{90}{\textbf{Metric}}} 
    & \davft{} (Ours) & 62.51M & 99\% & 1\%* & Gen. Open-World, \lusnar. \\
    \cline{2-6}
    & \dav{} \cite{yang_depth_2024} & 62.5M & 99\% & 1\%* & General Open-World \\
    & \davthree{} \cite{depthanything3} & 69M & 97\% & 100\%* & General Open-World \\
    & \moge{} \cite{wang2025moge2} & 8.8M & 74\% & 98\% & Indoor \\
    & \depthpro{} \cite{bochkovskii_depth_2024} & 3M & 15\% & 98\% & Synthetic Simulation \\
    & \metricd{} \cite{hu_metric3dv2_2024} & 16.2M & 99\% & 85\% & Indoor, Outdoor Driving \\
    & \unidepth{} \cite{piccinelli_unidepthv2_2025} & 16.6M & 87\% & 100\% & General Open-World \\
    & \mapanything{} \cite{keetha2025mapanything} & 114M$^{\ddagger}$ & 49\% & 54\% & Indoor, Gen. OpenWorld\\
    & \vda{}\cite{video_depth_anything} & 1.35M & 59\% & 41\% & General Open-World \\
    & \metricanything{} \cite{metricanything2026} & 20M & 88\% & 100\% & Indoor, Outdoor Driving \\
    
    \hline 
     
    \multirow{5}{*}{\rotatebox[origin=c]{90}{\textbf{Relative}}}
    & \daac{}\cite{sun2025depth} & 0.54M & 92\% & 23\% & General Open-World \\
    & \depthcrafter{} \cite{hu_depthcrafter_2024} & 25.5M$^{\dagger}$ & 98\% & 2\% & General Open-World \\
    & \lotus{} \cite{he_lotus_2025} & 0.07M & 0\% & 100\% & Indoor \\
    & \midas{} \cite{birkl2023midas} & 2.5M & 80\% & 35\% & General Open-World \\
    & \marigold{} \cite{ke_marigold_2024} & 0.07M & 0\% & 100\% & Indoor \\
     
    \hline 
    \multicolumn{6}{l}{$^{\dagger}$ $\approx$25.45M frames estimated from 203k video sequences.} \\
    \multicolumn{6}{l}{$^{\ddagger}$ Estimated from $\approx$188k scenes.} \\
  \end{tabular}%
  }
  \vspace{-10pt}
  \begin{flushleft}
        \scriptsize \hspace{1pt} $^{*}$ Utilize a Teacher-Student paradigm; DAv2 uses affine-invariant pseudo-labels, while DA3 aligns labels to metric ground truth via RANSAC.

    \end{flushleft}
    
\end{table}

\noindent\textbf{Impact of Training Scale.} Training scale appears to be dataset dependent: On authentic \change imagery, scale correlates significantly with higher depth accuracy ($\delta_1$: $\rho=0.718$, $p=0.004$) and reduced error (RMSE: $\rho=-0.619$, $p=0.018$), even within heavy shadows. However, these correlations weaken or turn negative on the Cheri analog and synthetic benchmarks, proving massive training scale alone cannot universally solve lunar adaptation.

\noindent\textbf{Real Training Data.} The ratio of real terrestrial training data does not correlate with performance improvements ($\rho=-0.677$, $p=0.008$). Thus, terrestrial texture priors primarily assist with nearby high frequency details but fail to provide geometric understanding required for long range lunar topography.

\noindent\textbf{Role of Metric Supervision.} We observed no statistically significant correlation between metric supervision percentage and zero-shot performance across any dataset ($p > 0.05$). This demonstrates that raw metric label volume cannot guarantee cross domain robustness, particularly alongside affine invariant loss functions. Lunar terrain reliability stems from architectural choices or massive scale invariant learning rather than absolute metric data quantities.

\begin{myframe}{Main Result}
While large training scale and terrestrial data marginally improve near-field lunar perception, they fail to resolve the broader extraterrestrial domain gap. Since raw metric supervision also cannot guarantee cross-domain robustness, future advancements require domain-specific architectural innovations rather than simple data scaling.
\end{myframe}

\begin{figure*}
    \centering
    \includegraphics[width=0.90\linewidth]{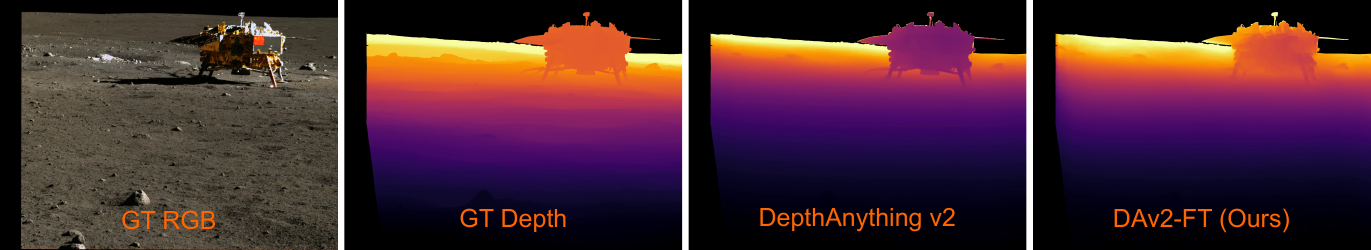}\\[-6pt]
    
    \caption{An example scene from Chang'e 3 (\texttt{RGB GT}), its ground-truth depth map (\texttt{GT Depth}), and the DepthAnything v2 predictions (base and fine-tuned). Note: Differences in visual boundaries between the \texttt{RGB GT} and depth maps result from the evaluation procedure (Section  \ref{sect:evaluation procedure}); where ground-truth maps and predictions are masked by maximum  depth boundaries and invalid pixels. Additionally, the depth maps generated by the other models can be found in the supplementary material. 
    \label{fig:qual_results_merged}}
\end{figure*}

\subsection{Exp. 5: Sim-to-Real Domain Adaptation}As a practical demonstration of our benchmarking framework, we finetuned the metric-outdoor weights of \dav explicitly on the \lusnar dataset. For high computational efficiency, we employed Low-Rank Adaptation (LoRA, $r=8$). By freezing the encoder backbone and updating only the decoder with LoRA, we reduced the trainable parameter count to $\sim$34.1M from 335.3M.

When evaluated on the \lusnar test set, the finetuned \davft model demonstrates a drastic performance improvement over its zero-shot baseline (See \cref{tab:exp1_overall,tab:exp2_shaded,tab:exp3_combined_arel,tab:exp2_semantic,tab:pose_estimation}). This accuracy leap is particularly pronounced across complex crater and rock formations, as well as heavily shaded regions. Furthermore, when applied to our downstream relative pose estimation pipeline, \davft delivers highly consistent geometric priors, achieving the lowest median rotation error and significantly outperforming the zero-shot base model.Crucially, this overall boost extends beyond the training domain: On \lunarsim with a distinct photorealism, \davft exhibits strong sim-to-sim generalization and outperforms the base model. However, on a broader selection of real and analog datasets, such as \etna, \seli, and \change, \davft achieves only marginal improvements. The model successfully aligns with specific simulator characteristics but struggles to transfer those learned features to authentic lunar photography and physical analogs.\begin{myframe}{Main Result}Parameter-efficient finetuning on synthetic data vastly improves downstream geometric consistency and bridges the sim-to-sim domain gap, but it fails to resolve the broader sim-to-real domain shift. Relying solely on simulated environments remains insufficient for lunar deployment, highlighting the necessity of diverse, real-world benchmarks.\end{myframe}

\subsection{Exp. 6: Relative Pose Est. using MDE Output}
\begin{table}[hbt!]
    \centering
    \scriptsize
    \caption{Exp. 6: Relative Pose Estimation using MADPose on \lusnar{} test-set. \textcolor{red}{\textbf{Red}}: Best, \textcolor{blue}{blue}: Second-best. \label{tab:pose_estimation}}
    
    \renewcommand{\arraystretch}{1.07}
    \setlength{\tabcolsep}{2.5pt}
    
    \begin{tabular}{c|l|cc|ccc}
        \hline
        
        \multirow{2}{*}{\textbf{Type}} & \multirow{2}{*}{\textbf{Method}}  & \multicolumn{2}{c|}{\rule{0pt}{3ex}\textit{Med. Err. $\downarrow$}} & \multicolumn{3}{c}{\textit{Pose Err. AUC (\%) $\uparrow$}}\\   
        
        \cline{3-7}
        
        & & \rule[-1.2ex]{0pt}{3.5ex} $\epsilon_R\ (\degree)$ & $\epsilon_t\ (\degree)$ & $@5\degree$ & $@10\degree$ & $@20\degree$\\
        \hline
        
        \multirow{10}{*}{\rotatebox[origin=c]{90}{\textbf{Metric}}} 
        & \davft{} (Ours) & \textcolor{red}{\textbf{0.12}} & \textcolor{blue}{\textbf{1.59}} & 57.70 & 78.05 & \textcolor{blue}{\textbf{89.03}} \\
        \cline{2-7}
        & \dav{} \cite{yang_depth_2024} & 0.15 & 1.89 & 56.22 & 77.27 & 88.64 \\
        & \davthree{} \cite{depthanything3} & \textcolor{blue}{\textbf{0.14}} & 2.17 & 53.28 & 76.04 & 88.02 \\
        & \moge{} \cite{wang2025moge2} & 0.15 & 1.68 & \textcolor{blue}{\textbf{58.40}} & \textcolor{red}{\textbf{78.52}} & \textcolor{red}{\textbf{89.26}} \\
        & \depthpro{} \cite{bochkovskii_depth_2024} & 0.22 & 2.84 & 41.12 & 69.78 & 84.89 \\
        & \metricd{} \cite{hu_metric3dv2_2024} & 0.29 & 3.24 & 37.25 & 67.22 & 83.61 \\
        & \unidepth{} \cite{piccinelli_unidepthv2_2025} & 0.15 & 2.12 & 53.69 & 76.11 & 88.06 \\
        & \mapanything{} \cite{keetha2025mapanything} & 0.21 & 2.36 & 47.26 & 72.19 & 86.09 \\
        & \vda{} \cite{video_depth_anything} & 0.22 & 2.98 & 38.86 & 68.39 & 84.20 \\
        & \metricanything{} \cite{metricanything2026} & 0.19 & 2.63 & 44.40 & 70.49 & 85.24 \\
        \hline
        
        \multirow{5}{*}{\rotatebox[origin=c]{90}{\textbf{Relative}}}
        & \daac{} \cite{sun2025depth} & 0.15 & \textcolor{red}{\textbf{1.48}} & \textcolor{red}{\textbf{58.70}} & \textcolor{blue}{\textbf{78.29}} & 88.81 \\
        & \depthcrafter{} \cite{hu_depthcrafter_2024} & 0.48 & 3.54 & 30.02 & 60.96 & 79.37 \\
        & \lotus{} \cite{he_lotus_2025} & 0.28 & 3.21 & 33.37 & 61.85 & 79.94 \\
        & \midas{} \cite{birkl2023midas}  & 0.26 & 3.72 & 30.18 & 58.25 & 77.59 \\
        & \marigold{} \cite{ke_marigold_2024} & 0.21 & 2.68 & 47.11 & 72.69 & 86.34 \\
        \hline
        
        & MASt3R \cite{mast3r}+GT & 0.11 & 1.32 & 61.78 & 80.25 & 90.12 \\
        \hline
    \end{tabular}
\end{table}

We further analyze the suitability of MDE models for robotic autonomy in lunar environments by integrating them into a relative pose estimation pipeline.

We adopt MADPose \cite{madpose}, a recent relative pose estimation solver that combines dense pixel correspondences with depth priors extracted from image pairs. By substituting the depth prior in MADPose with outputs from different MDE networks, we measure how depth quality affects pose estimation accuracy under lunar-like conditions. For our experiments, we use MASt3R \cite{mast3r} as the correspondence model and the previously evaluated MDE models as depth estimators.Given two successive frames $I_1$ and $I_2$ in a trajectory, MASt3R produces dense 2D pixel correspondences $p_{1j}$ and $p_{2j}$, while the MDE models predict the per-pixel depth maps $D_1$ and $D_2$. Using these predicted depths, we back-project each pixel correspondence into 3D camera coordinates $P_{1j}$ and $P_{2j}$. MADPose solvers then estimate the relative pose $T \in \text{SE}(3)$ such that $P_{2j} = RP_{1j} + t$, where $T = [R \mid t]$, $R \in SO(3)$, and $t \in \mathbb{R}^3$. During the evaluation, we mask out sky regions to eliminate invalid correspondences and improve solver performance. Following \cite{madpose} and \cite{wang2024posediffusionsolvingposeestimation}, we report the median rotation error $\epsilon_R$, median translation error $\epsilon_t$, and AUC@$\tau$ for $\tau \in \{5\degree, 10\degree, 20\degree\}$.

As shown in \Cref{tab:pose_estimation}, MDE models provide accurate depth priors for pose estimation in lunar surface conditions, indicating that MDE-based pose estimation is a viable solution for extraterrestrial navigation. Furthermore, metric MDE models consistently produce more geometrically reliable and consistent priors than relative MDE models.

\begin{myframe}{Main Result}Metric MDE models generate highly consistent geometric priors that enable accurate relative pose estimation on the lunar surface, significantly outperforming relative architectures and proving their direct viability for downstream extraterrestrial navigation tasks. Furthermore, finetuning on lunar data improves pose estimation results.\end{myframe}

\subsection{Supplementary Material}
The supplementary material includes an appendix with additional experimental results and detailed descriptions of the experimental setups. In addition, we provide video demonstrations of the results and the scripts used to generate the results.
\section{Conclusion and Discussion}

In this paper, we addressed the critical domain gap in deploying terrestrial Monocular Depth Estimation (MDE) networks for autonomous lunar rover navigation. We introduced LuMon, a comprehensive benchmarking framework and development suite featuring six diverse datasets. This suite notably includes novel high-quality ground truth for the authentic Chang'e-3 mission and the challenging Cheri dark analog environment. By systematically evaluating 14 state-of-the-art architectures, we established a rigorous foundation for understanding how current models handle the harsh lighting, textureless regolith, and unique topological hazards of the lunar surface.

\subsection{Summary of Findings}
Rather than restating our quantitative results, we highlight the overarching themes revealed by our benchmarking suite:

\noindent\textbf{The Limits of Terrestrial Priors.} Massive training scale and high ratios of real terrestrial data marginally improve near-field details but fail to bridge the broader extraterrestrial domain gap. Furthermore, Earth-based analogs can be deceptive; models easily parse well-lit volcanic terrain due to its similarity to terrestrial training distributions, yet they degrade severely under authentic high-contrast polar lighting and on complex geometries like craters.

\noindent\textbf{Advantage of Metric Foundation Models.} Metric depth estimators consistently outrank relative and diffusion-based models across the benchmark. They maintain superior internal linearity and structural integrity, proving far more robust for downstream tasks like relative pose estimation where consistent metric scaling is paramount.

\noindent\textbf{The Sim-to-Real Bottleneck.} Our parameter-efficient fine-tuning baseline demonstrates that while adapting foundation models on synthetic data yields excellent sim-to-sim transfer, it produces minimal gains on authentic lunar imagery. This stagnation proves that relying solely on simulated environments or raw data scaling is insufficient for mission-critical deployment.

\subsection{The LuMon Development Suite}
We designed LuMon not just as a static benchmark, but as an active development suite for the robotics and vision communities. Researchers can leverage our curated synthetic, analog, and real datasets to rigorously test novel architectures against specific lunar edge cases, such as extreme shadows and obstacle detection. Additionally, our established zero-shot rankings, downstream pose estimation pipeline, and LoRA fine-tuning baseline provide a standardized framework for developing and validating new domain adaptation strategies.

\subsection{Open Challenges and Future Work}
Based on our analysis, we identify three primary challenges that must guide future lunar perception research:

\begin{itemize}
    \item \textbf{Domain-Specific Architectural Innovation.} Since scaling terrestrial data and raw metric supervision does not guarantee cross-domain robustness, future work must focus on architectural innovations. Developing models that explicitly decouple illumination from geometric features or integrate physics-based rendering priors will be necessary to overcome the sim-to-real domain shift.
\item \textbf{Robustness to Sensor Calibration Artifacts.} Our experiments expose a systematic vulnerability in architectures, notably Depth Pro, to black border artifacts caused by stereo rectification. Since authentic planetary missions rely on stereo cameras, these zero-value boundaries are ubiquitous. While cropping restores performance, this extreme sensitivity highlights a critical flaw in terrestrial models trained on idealized rectangular images. Future methodologies must teach networks to natively ignore unstructured boundaries, ensuring robust perception on raw mission data.
    \item \textbf{On-Board Computational Constraints.} Our evaluation utilized the most capable weights for each architecture to establish an accuracy upper bound. However, practical lunar rovers operate under extreme power and memory constraints. Future research must utilize the LuMon suite to investigate the trade-off between estimation accuracy and computational efficiency, ultimately identifying or distilling lightweight architectures suitable for deployment on resource-constrained space hardware.
\end{itemize}

\section*{Acknowledgment}

We acknowledge the computational resources kindly provided by METU-ROMER. G. Cinbis \& S. Kalkan are supported by the “Young Scientist Awards Program (BAGEP)” of Science Academy, T\"urkiye. This study is supported in part by the ADEP project with grant no. ADEP-313-2025-11656. We acknowledge the support of CHERI Micro Lunar Rovers Project Group.
{
    \small
    \bibliography{main}
}
\appendix
\clearpage
\onecolumn
\setcounter{page}{1}

\setcounter{section}{0}
\setcounter{figure}{0}
\setcounter{table}{0}
\setcounter{equation}{0}

\renewcommand{\thesection}{\Alph{section}}
\renewcommand{\thefigure}{S\arabic{figure}}
\renewcommand{\thetable}{S\arabic{table}}
\renewcommand{\theequation}{S\arabic{equation}}

\begin{center}
   \Large \textbf{\methodname: A Comprehensive Benchmark and Development Suite with Novel Datasets for Lunar Monocular Depth Estimation} \\
   \vspace{1em} 
   
   \large \textbf{Supplementary Material} \\
   \vspace{0.5em}
   
\end{center}
\vspace{2em}


\section{Evaluation Metrics}

To quantitatively evaluate the depth prediction performance against the ground truth, we utilize a standard set of depth estimation metrics \cite{eigen_depth_2014}. Let $N$ be the total number of valid pixels in the evaluation set, $d_i$ denote the predicted depth at pixel $i$, and $d_i^*$ denote the corresponding ground-truth depth.

\textbf{Threshold Accuracy ($\delta_n$):} This metric measures the percentage of pixels where the relative error between the predicted and ground-truth depth falls within a specified threshold. Higher values indicate better performance.
\begin{equation}
    \delta_n = \frac{1}{N} \sum_{i=1}^{N} \mathbf{1} \left( \max \left( \frac{d_i}{d_i^*}, \frac{d_i^*}{d_i} \right) < 1.25^n \right)
\end{equation}
where $n \in \{1, 2, 3\}$ and $\mathbf{1}(\cdot)$ is the indicator function.

\textbf{Absolute Relative Error (A.Rel):} This computes the mean of the absolute percentage errors across all valid pixels.
\begin{equation}
    \text{A.Rel} = \frac{1}{N} \sum_{i=1}^{N} \frac{|d_i - d_i^*|}{d_i^*}
\end{equation}

\textbf{Squared Relative Error (Sq.Rel):} Similar to A.Rel, but the relative differences are squared, which more heavily penalizes large errors.
\begin{equation}
    \text{Sq.Rel} = \frac{1}{N} \sum_{i=1}^{N} \frac{(d_i - d_i^*)^2}{d_i^*}
\end{equation}

\textbf{Root Mean Square Error (RMSE):} A standard statistical metric that measures the root of the average squared difference, capturing the magnitude of the error in the original metric space.
\begin{equation}
    \text{RMSE} = \sqrt{\frac{1}{N} \sum_{i=1}^{N} (d_i - d_i^*)^2}
\end{equation}

\textbf{Mean Absolute Error (MAE):} The simple average of the absolute differences between the predicted and ground-truth depth values.
\begin{equation}
    \text{MAE} = \frac{1}{N} \sum_{i=1}^{N} |d_i - d_i^*|
\end{equation}

\textbf{Scale-Invariant Logarithmic Error (SILog):} This metric evaluates the error independently of the global scale, focusing strictly on the quality of the relative depth structures.
\begin{equation}
    \text{SILog} = \frac{1}{N} \sum_{i=1}^{N} (\log d_i - \log d_i^*)^2 - \frac{1}{N^2} \left( \sum_{i=1}^{N} (\log d_i - \log d_i^*) \right)^2
\end{equation}

\textbf{Edge Metrics (E.Acc and E.Comp):} To evaluate the structural fidelity at depth boundaries, we employ the edge accuracy (Acc-Edges) and edge completion (Comp-Edges) metrics from the IBims-1 benchmark \cite{koch2018evaluationcnnbasedsingleimagedepth}. These metrics isolate depth boundary pixels identified via edge detection on the log-depth maps. Let $E_p$ be the set of spatial pixel coordinates representing edges in the predicted depth map, and $E_{gt}$ be the set of spatial pixel coordinates for edges in the ground-truth depth map. 

Edge Accuracy ($E_{acc}$) measures the precision of the predicted depth discontinuities by computing the average distance from each predicted edge pixel to its nearest ground-truth counterpart:
\begin{equation}
    E_{acc} = \frac{1}{|E_p|} \sum_{x \in E_p} \min \left( \min_{y \in E_{gt}} ||x - y||_2, \theta \right)
\end{equation}

Edge Completion ($E_{comp}$) measures the recall, effectively evaluating how well the model maintains the sharpness of edges compared to the ground truth by computing the average distance from each ground-truth edge pixel to the nearest predicted edge:
\begin{equation}
    E_{comp} = \frac{1}{|E_{gt}|} \sum_{y \in E_{gt}} \min \left( \min_{x \in E_p} ||x - y||_2, \theta \right)
\end{equation}

where $|E|$ denotes the total number of pixels in the respective edge set, $||x - y||_2$ represents the Euclidean distance between two pixel coordinates, and $\theta$ is a maximum distance threshold (typically $\theta = 10$ pixels) used to cap the penalty for extreme outliers. For both metrics, a lower value indicates better structural fidelity at depth boundaries. \\
\textbf{Median Translation Error:} Let $R$ and $\hat{R}$ denote the ground-truth and estimated relative rotation matrices between two views. The rotation error is defined as the geodesic distance between the two rotations:

\begin{equation}
\epsilon_R = \arccos\left(\frac{\mathrm{Tr}\left(R^{\top}\hat{R}\right)-1}{2}\right)
\end{equation}

where $\mathrm{Tr}(\cdot)$ denotes the matrix trace. The median rotation error is computed across all evaluated image pairs.

\textbf{Median Translation Error:} Let $t$ and $\hat{t}$ denote the ground-truth and estimated relative translation vectors. We measure the angular error between the translation directions:

\begin{equation}
\epsilon_t = \arccos\left(\frac{t^{\top}\hat{t}}{\|t\| \, \|\hat{t}\|}\right)
\end{equation}

The median translation error is then computed over all image pairs.

\textbf{Area Under Curve:} W evaluate the overall pose accuracy using the Area Under the Curve (AUC) of the pose error distribution. For each image pair, the pose error is defined as

\begin{equation}
\epsilon = \max(\epsilon_R, \epsilon_t).
\end{equation}

The AUC is computed as the area under the cumulative accuracy curve up to a threshold $\tau$:

\begin{equation}
\text{AUC}@\tau = \frac{1}{\tau} \int_{0}^{\tau} A(e) \, de
\end{equation}

where $A(e)$ denotes the fraction of image pairs whose pose error $\epsilon$ is smaller than $e$. We report AUC values for $\tau \in \{5^\circ, 10^\circ, 20^\circ\}$.

\section{Relative Pose Estimation}
Given two RGB images $I_1$ and $I_2$, we first compute dense 2D pixel correspondences using the fast reciprocal matching procedure of MASt3R \cite{mast3r}:
\begin{equation}
    p_{12}, p_{21} = \text{MASt3R}(I_1, I_2)
\end{equation}
where $p_{12}$ and $p_{21}$ denote the sets of matched pixel coordinates $(u,v)$ from $I_1$ to $I_2$ and from $I_2$ to $I_1$, respectively. Then, we extract pixel-wise depth maps $D_1$, $D_2$ using an MDE network:

\begin{equation}
    D_i = \text{MDE\_Model}(I_i), \quad i \in \{1,2\}.
\end{equation}

The resulting depth maps $D_1, D_2$ and pixel correspondences $p_{12}, p_{21}$ are then provided to the MADPose framework \cite{madpose}. Given that both images share the same intrinsic calibration matrix $K$, we employ the calibrated solver of MADPose to estimate the relative camera pose between $I_1$ and $I_2$.

\clearpage

\section{Detailed Fine-Tuning Protocol for Sim-to-Real Adaptation (Exp.~5)}
\label{sec:supp_finetune_lusnar}

We fine-tune the \textit{DepthAnything v2 metric-outdoor (ViT-L)} checkpoint on LuSNAR depth supervision using a decoder-focused LoRA setup. 
Starting from the full model (\(\sim 335.3\)M trainable parameters under full fine-tuning), we freeze the pretrained encoder backbone and inject LoRA adapters into encoder linear projections while keeping the depth decoder trainable. 
With LoRA rank \(r=8\), scaling \(\alpha=16\), and dropout \(0.0\), this reduces trainable parameters to \(34{,}092{,}737\) (\(\approx 34.1\)M), while preserving strong adaptation capacity.

\paragraph{Data split and preprocessing.}
LuSNAR sequence splits are fixed as:
\[
\mathcal{S}_{train}=\{\text{Moon\_2, Moon\_3, Moon\_4, Moon\_5, Moon\_7, Moon\_9}\},
\]
\[
\mathcal{S}_{val}=\{\text{Moon\_6}\}, \quad
\mathcal{S}_{test}=\{\text{Moon\_1, Moon\_8}\}.
\]
RGB frames are normalized with ImageNet statistics, and image/depth pairs are resized with aspect-ratio preservation to input size \(518\) (rounded to a multiple of 14 for backbone compatibility). 
Depth validity masking removes non-finite values, non-positive depths, and invalid sentinel values (\(65504\)); for training/evaluation loss we further restrict to \(d \le 50\) m.

\paragraph{Objective function.}
Let \(M\) denote valid pixels in a sample, \(y_i\) the ground-truth depth, and \(\hat y_i\) the prediction. 
The total training loss is
\[
\mathcal{L} = \lambda_s \mathcal{L}_{\text{SiLog}} + \lambda_g \mathcal{L}_{\nabla},
\]
with \(\lambda_s=\lambda_g=1\).
For scale-invariant logarithmic loss, we define
\[
\delta_i = \log(y_i+\epsilon)-\log(\hat y_i+\epsilon),
\]
\[
\mathcal{L}_{\text{SiLog}}
=
\sqrt{
\frac{1}{|M|}\sum_{i\in M}\delta_i^2
-
\lambda
\left(
\frac{1}{|M|}\sum_{i\in M}\delta_i
\right)^2
+\epsilon
},
\]
where \(\lambda=0.5\) and \(\epsilon=10^{-6}\).
We complement this with a gradient-matching term over valid horizontal/vertical depth differences:
\[
\mathcal{L}_{\nabla}
=
\frac{1}{2}\left(
\operatorname*{mean}_{(i,j)\in M_x}\left|(\partial_x \hat y)_{ij}-(\partial_x y)_{ij}\right|
+
\operatorname*{mean}_{(i,j)\in M_y}\left|(\partial_y \hat y)_{ij}-(\partial_y y)_{ij}\right|
\right).
\]

\paragraph{Optimization details.}
We optimize trainable parameters with AdamW (learning rate \(5\times10^{-5}\), weight decay \(10^{-2}\)); total epochs are set to 80 with linear warm-up for 2 epochs followed by cosine decay to \(10^{-6}\). 
Batch size is selected automatically by OOM-safe probing (search range 128\(\rightarrow\)2), resulting in effective batch size 16 in the reported LoRA-\(r=8\) run. 
Early stopping uses validation \(\text{AbsRel}\) with patience 10; checkpoint selection is based on best validation \(\text{AbsRel}\).

\paragraph{Evaluation protocol during training.}
For monitoring, we report $\delta_1$,$\delta_2$, $\delta_3$, MAE, AbsRel, RMSE, SiLog, SqRel, EdgeAcc and EdgeComp. 
Predictions are least-squares aligned to GT before metric computation, and the same \(50\) m validity mask is applied. 
Best-model selection is then used for the test-time results discussed in the main paper.

\clearpage

\section{Extended Quantitative Results}


\subsection{Overall Evaluation}
\label{sec:supp_overall}

\Cref{tab:exp1_classic_lunarsim,tab:exp1_classic_lusnar,tab:exp1_classic_etna,tab:exp1_classic_s3li,tab:exp1_classic_change,tab:exp1_classic_cheri} present the disaggregated per-dataset results that underpin the main text's overall comparison (Exp.~1). While the main paper reports aggregated trends, here we provide the full metric breakdown for each benchmark individually to enable fine-grained analysis of cross-domain generalization patterns. Our affine-aligned evaluation protocol ensures that all comparisons, including those involving relative methods, reflect geometric and structural accuracy rather than raw scale recovery.\\

\noindent\textbf{\lunarsim} (\cref{tab:exp1_classic_lunarsim}). On this synthetic benchmark, \mapanything{} achieves the strongest zero-shot performance among metric methods ($\delta_1{=}0.87$, A.Rel${=}0.11$, RMSE${=}10.74$), followed by \davft{} ($\delta_1{=}0.84$) and \metricanything{} (RMSE${=}12.08$). Among relative methods, \midas{} attains the highest accuracy ($\delta_1{=}0.78$), yet remains behind the metric leaders. This gap indicates that metric foundation models encode stronger geometric priors that transfer more robustly to synthetic lunar terrain, even without domain-specific finetuning.

\begin{table*}[t]
    \centering
    \caption{Exp. 1: Results on \textbf{\lunarsim} dataset. \textcolor{red}{\textbf{Red}} indicates the best result and \textcolor{blue}{blue} indicates the second best.}
    \label{tab:exp1_classic_lunarsim}
    \setlength{\tabcolsep}{1.5pt}
    \renewcommand{\arraystretch}{1.05}
    \resizebox{\textwidth}{!}{
    \begin{tabular}{c|l| c c c c c c c c c c}
        \hline
        \textbf{Type} & \textbf{Method} & $\delta_{1}$$\uparrow$ & $\delta_{2}$$\uparrow$ & $\delta_{3}$$\uparrow$ & A.Rel$\downarrow$ & Sq.Rel$\downarrow$ & RMSE$\downarrow$ & MAE$\downarrow$ & SILog$\downarrow$ & E.Acc$\downarrow$ & E.Comp$\downarrow$ \\
        \hline
        \multirow{10}{*}{\rotatebox[origin=c]{90}{\textbf{Metric}}}
        & \davft{} & \textcolor{blue}{0.84} & \textcolor{red}{\textbf{0.97}} & \textcolor{red}{\textbf{0.99}} & \textcolor{blue}{0.13} & 2.04 & 13.96 & 7.04 & \textcolor{red}{\textbf{15.95}} & 2.56 & 130.43 \\
        \cline{2-12}
        & \dav{} \cite{yang_depth_2024} & 0.56 & 0.81 & 0.89 & 0.26 & 3.62 & 15.18 & 10.04 & 95.84 & 3.57 & 106.37 \\
        & \davthree{} \cite{depthanything3} & 0.63 & 0.85 & 0.91 & 0.22 & 2.84 & 14.15 & 8.89 & 55.14 & 3.19 & 157.79 \\
        & \moge{} \cite{wang2025moge2} & 0.77 & 0.93 & 0.97 & 0.15 & 1.76 & 12.17 & 6.80 & 24.89 & \textcolor{blue}{2.02} & 140.01 \\
        & \depthpro{} \cite{bochkovskii_depth_2024} & 0.50 & 0.79 & 0.88 & 0.29 & 4.64 & 17.57 & 11.45 & 84.24 & 3.58 & 152.95 \\
        & \metricd{} \cite{hu_metric3dv2_2024} & 0.75 & 0.91 & 0.97 & 0.16 & 1.99 & 12.53 & 7.10 & 29.10 & 2.47 & \textcolor{blue}{72.61} \\
        & \unidepth{} \cite{piccinelli_unidepthv2_2025} & 0.76 & 0.90 & 0.95 & 0.21 & 5.12 & 16.16 & 8.89 & 23.48 & \textcolor{red}{\textbf{1.98}} & 168.42 \\
        & \mapanything{} \cite{keetha2025mapanything} & \textcolor{red}{\textbf{0.87}} & \textcolor{red}{\textbf{0.97}} & \textcolor{red}{\textbf{0.99}} & \textcolor{red}{\textbf{0.11}} & \textcolor{red}{\textbf{1.23}} & \textcolor{red}{\textbf{10.74}} & \textcolor{red}{\textbf{5.53}} & 25.10 & 2.42 & 169.53 \\
        & \vda{} \cite{video_depth_anything} & 0.75 & 0.93 & \textcolor{blue}{0.98} & 0.16 & 2.23 & 13.94 & 7.41 & 25.10 & 2.51 & 96.24 \\
        & \metricanything{} \cite{metricanything2026} & 0.79 & 0.93 & 0.97 & 0.15 & \textcolor{blue}{1.71} & \textcolor{blue}{12.08} & \textcolor{blue}{6.68} & 28.04 & 2.25 & 129.25 \\
        \hline
        \multirow{5}{*}{\rotatebox[origin=c]{90}{\textbf{Relative}}}
        & \daac{} \cite{sun2025depth} & 0.70 & 0.85 & 0.91 & 0.28 & 8.12 & 19.12 & 11.43 & 26.00 & 2.36 & 152.78 \\
        & \depthcrafter{} \cite{hu_depthcrafter_2024} & 0.40 & 0.73 & 0.81 & 0.37 & 6.62 & 20.28 & 14.05 & 316.46 & 4.39 & 208.08 \\
        & \lotus{} \cite{he_lotus_2025} & 0.59 & 0.84 & 0.93 & 0.27 & 5.70 & 20.43 & 11.67 & 36.21 & 2.85 & 97.52 \\
        & \midas{} \cite{birkl2023midas} & 0.78 & \textcolor{blue}{0.95} & \textcolor{red}{\textbf{0.99}} & 0.17 & 2.68 & 14.97 & 7.85 & \textcolor{blue}{20.56} & 2.60 & 147.71 \\
        & \marigold{} \cite{ke_marigold_2024} & 0.67 & 0.89 & 0.96 & 0.21 & 3.73 & 17.25 & 9.88 & 37.98 & 2.84 & \textcolor{red}{\textbf{49.23}} \\
        \hline
    \end{tabular}
    }
\end{table*}

\clearpage
\noindent\textbf{\lusnar} (\cref{tab:exp1_classic_lusnar}). The finetuned \davft{} model dominates across every metric ($\delta_1{=}0.93$, A.Rel${=}0.07$, RMSE${=}1.07$), establishing the upper bound on this dataset with a substantial margin over the second-best method, \midas{} ($\delta_1{=}0.81$, RMSE${=}1.68$). This result quantifies the direct benefit of domain-specific adaptation on synthetic lunar terrain. However, as discussed in Exp.~5 of the main text, this accuracy advantage does not transfer reliably to real-world imagery, underscoring the sim-to-real bottleneck that persists even after targeted finetuning.

\begin{table*}[t]
    \centering
    \caption{Exp. 1: Results on \textbf{\lusnar} dataset. \textcolor{red}{\textbf{Red}} indicates the best result and \textcolor{blue}{blue} indicates the second best.}
    \label{tab:exp1_classic_lusnar}
    \setlength{\tabcolsep}{1.5pt}
    \renewcommand{\arraystretch}{1.05}
    \resizebox{\textwidth}{!}{
    \begin{tabular}{c|l| c c c c c c c c c c}
        \hline
        \textbf{Type} & \textbf{Method} & $\delta_{1}$$\uparrow$ & $\delta_{2}$$\uparrow$ & $\delta_{3}$$\uparrow$ & A.Rel$\downarrow$ & Sq.Rel$\downarrow$ & RMSE$\downarrow$ & MAE$\downarrow$ & SILog$\downarrow$ & E.Acc$\downarrow$ & E.Comp$\downarrow$ \\
        \hline
        \multirow{10}{*}{\rotatebox[origin=c]{90}{\textbf{Metric}}}
        & \davft{} & \textcolor{red}{\textbf{0.93}} & \textcolor{red}{\textbf{0.99}} & \textcolor{red}{\textbf{1.00}} & \textcolor{red}{\textbf{0.07}} & \textcolor{red}{\textbf{0.07}} & \textcolor{red}{\textbf{1.07}} & \textcolor{red}{\textbf{0.41}} & \textcolor{red}{\textbf{9.13}} & \textcolor{red}{\textbf{1.09}} & \textcolor{red}{\textbf{2.53}} \\
        \cline{2-12}
        & \dav{} \cite{yang_depth_2024} & 0.31 & 0.63 & 0.73 & 0.45 & 0.92 & 2.89 & 1.77 & 342.41 & 7.10 & 86.42 \\
        & \davthree{} \cite{depthanything3} & 0.53 & 0.76 & 0.85 & 0.29 & 0.44 & 1.99 & 1.13 & 121.12 & 2.26 & 38.61 \\
        & \moge{} \cite{wang2025moge2} & 0.53 & 0.75 & 0.84 & 0.29 & 0.40 & 1.80 & 1.07 & 136.76 & 2.11 & 36.22 \\
        & \depthpro{} \cite{bochkovskii_depth_2024} & 0.33 & 0.64 & 0.72 & 0.45 & 0.91 & 2.70 & 1.69 & 374.12 & 6.65 & 99.65 \\
        & \metricd{} \cite{hu_metric3dv2_2024} & 0.61 & 0.79 & 0.86 & 0.25 & 0.33 & 1.71 & 0.97 & 108.62 & 1.92 & 45.65 \\
        & \unidepth{} \cite{piccinelli_unidepthv2_2025} & 0.63 & 0.81 & 0.88 & 0.26 & 0.50 & 2.14 & 1.04 & 75.96 & 1.27 & 24.04 \\
        & \mapanything{} \cite{keetha2025mapanything} & 0.74 & 0.89 & 0.94 & 0.17 & 0.23 & \textcolor{blue}{1.53} & 0.73 & 66.28 & 2.28 & 50.17 \\
        & \vda{} \cite{video_depth_anything} & 0.77 & 0.93 & 0.97 & 0.17 & 0.29 & 2.05 & 0.80 & 28.94 & \textcolor{blue}{1.21} & 10.25 \\
        & \metricanything{} \cite{metricanything2026} & 0.51 & 0.74 & 0.83 & 0.30 & 0.41 & 1.83 & 1.11 & 136.98 & 1.82 & 48.57 \\
        \hline
        \multirow{5}{*}{\rotatebox[origin=c]{90}{\textbf{Relative}}}
        & \daac{} \cite{sun2025depth} & 0.70 & 0.86 & 0.92 & 0.26 & 0.77 & 2.63 & 1.14 & 24.30 & 1.29 & 8.31 \\
        & \depthcrafter{} \cite{hu_depthcrafter_2024} & 0.41 & 0.69 & 0.78 & 0.47 & 1.13 & 2.96 & 1.77 & 448.40 & 6.53 & 146.50 \\
        & \lotus{} \cite{he_lotus_2025} & 0.36 & 0.70 & 0.85 & 0.44 & 1.05 & 3.08 & 1.71 & 150.41 & 2.82 & 35.69 \\
        & \midas{} \cite{birkl2023midas} & \textcolor{blue}{0.81} & \textcolor{blue}{0.95} & \textcolor{blue}{0.98} & \textcolor{blue}{0.13} & \textcolor{blue}{0.18} & 1.68 & \textcolor{blue}{0.69} & \textcolor{blue}{17.61} & 1.32 & \textcolor{blue}{4.63} \\
        & \marigold{} \cite{ke_marigold_2024} & 0.62 & 0.81 & 0.87 & 0.24 & 0.36 & 2.02 & 1.03 & 72.64 & 1.79 & 12.26 \\
        \hline
    \end{tabular}
    }
\end{table*}

\clearpage
\noindent\textbf{\etna} (\cref{tab:exp1_classic_etna}). This Earth-based analog proves to be the least challenging benchmark, with nine methods reaching $\delta_1{=}1.00$. Because the well-lit volcanic terrain closely overlaps with standard terrestrial training distributions, the dataset cannot meaningfully differentiate model capabilities at the $\delta_1$ level. Discrimination requires finer error metrics: \unidepth{} and \mapanything{} both achieve A.Rel${=}0.03$, with \unidepth{} obtaining the lowest RMSE of $0.07$ and SILog of $3.12$. These results reinforce our main text observation that near-perfect performance on well-lit Earth analogs can be deceptive for evaluating mission-critical extraterrestrial deployment.

\begin{table*}[t]
    \centering
    \caption{Exp. 1: Results on \textbf{\etna} dataset. \textcolor{red}{\textbf{Red}} indicates the best result and \textcolor{blue}{blue} indicates the second best.}
    \label{tab:exp1_classic_etna}
    \setlength{\tabcolsep}{1.5pt}
    \renewcommand{\arraystretch}{1.05}
    \resizebox{\textwidth}{!}{
    \begin{tabular}{c|l| c c c c c c c c c c}
        \hline
        \textbf{Type} & \textbf{Method} & $\delta_{1}$$\uparrow$ & $\delta_{2}$$\uparrow$ & $\delta_{3}$$\uparrow$ & A.Rel$\downarrow$ & Sq.Rel$\downarrow$ & RMSE$\downarrow$ & MAE$\downarrow$ & SILog$\downarrow$ & E.Acc$\downarrow$ & E.Comp$\downarrow$ \\
        \hline
        \multirow{10}{*}{\rotatebox[origin=c]{90}{\textbf{Metric}}}
        & \davft{} & \textcolor{red}{\textbf{1.00}} & \textcolor{red}{\textbf{1.00}} & \textcolor{red}{\textbf{1.00}} & 0.05 & \textcolor{blue}{0.01} & 0.13 & 0.10 & 5.56 & 1.90 & 41.06 \\
        \cline{2-12}
        & \dav{} \cite{yang_depth_2024} & \textcolor{red}{\textbf{1.00}} & \textcolor{red}{\textbf{1.00}} & \textcolor{red}{\textbf{1.00}} & 0.06 & \textcolor{blue}{0.01} & 0.15 & 0.12 & 7.01 & 3.81 & \textcolor{blue}{4.84} \\
        & \davthree{} \cite{depthanything3} & \textcolor{red}{\textbf{1.00}} & \textcolor{red}{\textbf{1.00}} & \textcolor{red}{\textbf{1.00}} & \textcolor{blue}{0.04} & \textcolor{blue}{0.01} & 0.11 & 0.08 & 4.91 & 2.39 & 36.57 \\
        & \moge{} \cite{wang2025moge2} & \textcolor{blue}{0.99} & \textcolor{red}{\textbf{1.00}} & \textcolor{red}{\textbf{1.00}} & 0.05 & \textcolor{blue}{0.01} & 0.16 & 0.11 & 6.37 & 3.33 & 13.87 \\
        & \depthpro{} \cite{bochkovskii_depth_2024} & 0.73 & 0.93 & 0.95 & 241.71 & 2415143.42 & 450.05 & 449.98 & 18.12 & 3.01 & 70.35 \\
        & \metricd{} \cite{hu_metric3dv2_2024} & \textcolor{red}{\textbf{1.00}} & \textcolor{red}{\textbf{1.00}} & \textcolor{red}{\textbf{1.00}} & \textcolor{blue}{0.04} & \textcolor{red}{\textbf{0.00}} & 0.10 & 0.07 & 4.40 & 2.44 & 6.42 \\
        & \unidepth{} \cite{piccinelli_unidepthv2_2025} & \textcolor{red}{\textbf{1.00}} & \textcolor{red}{\textbf{1.00}} & \textcolor{red}{\textbf{1.00}} & \textcolor{red}{\textbf{0.03}} & \textcolor{red}{\textbf{0.00}} & \textcolor{red}{\textbf{0.07}} & \textcolor{red}{\textbf{0.05}} & \textcolor{red}{\textbf{3.12}} & 1.73 & 15.91 \\
        & \mapanything{} \cite{keetha2025mapanything} & \textcolor{red}{\textbf{1.00}} & \textcolor{red}{\textbf{1.00}} & \textcolor{red}{\textbf{1.00}} & \textcolor{red}{\textbf{0.03}} & \textcolor{red}{\textbf{0.00}} & \textcolor{blue}{0.08} & \textcolor{blue}{0.06} & \textcolor{blue}{3.26} & 1.73 & 19.73 \\
        & \vda{} \cite{video_depth_anything} & \textcolor{red}{\textbf{1.00}} & \textcolor{red}{\textbf{1.00}} & \textcolor{red}{\textbf{1.00}} & \textcolor{blue}{0.04} & \textcolor{blue}{0.01} & 0.12 & 0.08 & 4.90 & 1.91 & 10.24 \\
        & \metricanything{} \cite{metricanything2026} & 0.98 & \textcolor{red}{\textbf{1.00}} & \textcolor{red}{\textbf{1.00}} & 0.07 & 0.02 & 0.22 & 0.15 & 8.69 & 3.45 & 6.31 \\
        \hline
        \multirow{5}{*}{\rotatebox[origin=c]{90}{\textbf{Relative}}}
        & \daac{} \cite{sun2025depth} & \textcolor{red}{\textbf{1.00}} & \textcolor{red}{\textbf{1.00}} & \textcolor{red}{\textbf{1.00}} & 0.05 & \textcolor{blue}{0.01} & 0.12 & 0.10 & 5.76 & \textcolor{red}{\textbf{1.66}} & 62.56 \\
        & \depthcrafter{} \cite{hu_depthcrafter_2024} & 0.92 & \textcolor{blue}{0.99} & \textcolor{red}{\textbf{1.00}} & 0.10 & 0.03 & 0.27 & 0.20 & 12.71 & 4.21 & 21.02 \\
        & \lotus{} \cite{he_lotus_2025} & 0.65 & 0.96 & \textcolor{blue}{0.99} & 0.19 & 0.11 & 0.49 & 0.38 & 22.54 & 4.16 & 9.65 \\
        & \midas{} \cite{birkl2023midas} & \textcolor{red}{\textbf{1.00}} & \textcolor{red}{\textbf{1.00}} & \textcolor{red}{\textbf{1.00}} & \textcolor{blue}{0.04} & \textcolor{blue}{0.01} & 0.12 & 0.09 & 5.37 & \textcolor{blue}{1.68} & 77.97 \\
        & \marigold{} \cite{ke_marigold_2024} & 0.94 & \textcolor{red}{\textbf{1.00}} & \textcolor{red}{\textbf{1.00}} & 0.10 & 0.03 & 0.27 & 0.20 & 11.91 & 3.75 & \textcolor{red}{\textbf{4.42}} \\
        \hline
    \end{tabular}
    }
\end{table*}

\clearpage
\noindent\textbf{\seli} (\cref{tab:exp1_classic_s3li}). The stricter LiDAR-derived ground truth of this dataset reveals tighter performance clustering among the top metric methods. Four architectures, namely \davthree{}, \moge{}, \mapanything{}, and \metricanything{}, tie at $\delta_1{=}0.91$ and A.Rel${=}0.10$, with \mapanything{} and \metricanything{} sharing the best RMSE of $1.05$. \midas{} is again the strongest relative method ($\delta_1{=}0.89$), narrowing the metric--relative gap relative to other benchmarks. This convergence indicates that when ground truth is calibrated via high-fidelity LiDAR rather than stereo reconstruction, the perceived advantage of metric models diminishes, suggesting that evaluation methodology itself significantly influences benchmark conclusions.

\begin{table*}[t]
    \centering
    \caption{Exp. 1: Results on \textbf{\seli} dataset. \textcolor{red}{\textbf{Red}} indicates the best result and \textcolor{blue}{blue} indicates the second best.}
    \label{tab:exp1_classic_s3li}
    \setlength{\tabcolsep}{1.5pt}
    \renewcommand{\arraystretch}{1.05}
    \resizebox{\textwidth}{!}{%
    \begin{tabular}{c|l| c c c c c c c c}
        \hline
        \textbf{Type} & \textbf{Method} & $\delta_{1}$$\uparrow$ & $\delta_{2}$$\uparrow$ & $\delta_{3}$$\uparrow$ & A.Rel$\downarrow$ & Sq.Rel$\downarrow$ & RMSE$\downarrow$ & MAE$\downarrow$ & SILog$\downarrow$ \\
        \hline
        \multirow{10}{*}{\rotatebox[origin=c]{90}{\textbf{Metric}}}
        & \davft{} & 0.84 & 0.97 & \textcolor{blue}{0.99} & 0.13 & 0.26 & 1.30 & 0.94 & 15.97 \\
        \cline{2-10}
        & \dav{} \cite{yang_depth_2024} & 0.88 & \textcolor{blue}{0.98} & \textcolor{blue}{0.99} & 0.12 & 0.21 & 1.22 & 0.85 & 34.64 \\
        & \davthree{} \cite{depthanything3} & \textcolor{red}{\textbf{0.91}} & \textcolor{blue}{0.98} & \textcolor{blue}{0.99} & \textcolor{red}{\textbf{0.10}} & 0.18 & 1.12 & 0.74 & 38.88 \\
        & \moge{} \cite{wang2025moge2} & \textcolor{red}{\textbf{0.91}} & \textcolor{red}{\textbf{0.99}} & \textcolor{red}{\textbf{1.00}} & \textcolor{red}{\textbf{0.10}} & \textcolor{blue}{0.17} & \textcolor{blue}{1.08} & \textcolor{blue}{0.73} & 28.97 \\
        & \depthpro{} \cite{bochkovskii_depth_2024} & \textcolor{blue}{0.89} & \textcolor{blue}{0.98} & \textcolor{red}{\textbf{1.00}} & \textcolor{blue}{0.11} & 0.18 & 1.11 & 0.78 & 16.58 \\
        & \metricd{} \cite{hu_metric3dv2_2024} & 0.87 & \textcolor{blue}{0.98} & \textcolor{blue}{0.99} & 0.12 & 0.22 & 1.19 & 0.83 & 27.04 \\
        & \unidepth{} \cite{piccinelli_unidepthv2_2025} & 0.87 & 0.97 & \textcolor{blue}{0.99} & 0.12 & 0.24 & 1.20 & 0.86 & 14.18 \\
        & \mapanything{} \cite{keetha2025mapanything} & \textcolor{red}{\textbf{0.91}} & \textcolor{red}{\textbf{0.99}} & \textcolor{red}{\textbf{1.00}} & \textcolor{red}{\textbf{0.10}} & \textcolor{red}{\textbf{0.16}} & \textcolor{red}{\textbf{1.05}} & \textcolor{red}{\textbf{0.72}} & 15.54 \\
        & \vda{} \cite{video_depth_anything} & 0.83 & 0.97 & \textcolor{blue}{0.99} & 0.14 & 0.27 & 1.34 & 0.97 & 17.44 \\
        & \metricanything{} \cite{metricanything2026} & \textcolor{red}{\textbf{0.91}} & \textcolor{red}{\textbf{0.99}} & \textcolor{red}{\textbf{1.00}} & \textcolor{red}{\textbf{0.10}} & \textcolor{red}{\textbf{0.16}} & \textcolor{red}{\textbf{1.05}} & \textcolor{red}{\textbf{0.72}} & 17.23 \\
        \hline
        \multirow{5}{*}{\rotatebox[origin=c]{90}{\textbf{Relative}}}
        & \daac{} \cite{sun2025depth} & 0.87 & \textcolor{blue}{0.98} & \textcolor{red}{\textbf{1.00}} & \textcolor{blue}{0.11} & 0.22 & 1.19 & 0.84 & \textcolor{blue}{14.01} \\
        & \depthcrafter{} \cite{hu_depthcrafter_2024} & 0.79 & 0.96 & \textcolor{blue}{0.99} & 0.15 & 0.30 & 1.44 & 1.06 & 26.78 \\
        & \lotus{} \cite{he_lotus_2025} & 0.80 & 0.96 & \textcolor{blue}{0.99} & 0.15 & 0.31 & 1.45 & 1.05 & 20.62 \\
        & \midas{} \cite{birkl2023midas} & \textcolor{blue}{0.89} & \textcolor{blue}{0.98} & \textcolor{red}{\textbf{1.00}} & \textcolor{blue}{0.11} & 0.20 & 1.15 & 0.81 & \textcolor{red}{\textbf{13.55}} \\
        & \marigold{} \cite{ke_marigold_2024} & 0.86 & 0.97 & \textcolor{blue}{0.99} & 0.12 & 0.24 & 1.26 & 0.89 & 22.09 \\
        \hline
    \end{tabular}%
    }
\end{table*}

\clearpage
\noindent\textbf{\change} (\cref{tab:exp1_classic_change}). On authentic lunar surface imagery, metric foundation models demonstrate the strongest zero-shot transfer. \davthree{} leads with $\delta_1{=}0.96$ and shares the best A.Rel of $0.05$ with \metricd{}. \metricd{} also achieves the lowest RMSE ($0.43$) and SILog ($7.12$), indicating superior absolute scale recovery on real extraterrestrial data. The consistently strong performance of metric architectures on this dataset validates their geometric priors for authentic lunar topography. Relative methods lag substantially; the best among them, \marigold{}, achieves only $\delta_1{=}0.91$ and RMSE${=}0.69$, showing that affine-invariant representations sacrifice structural fidelity that is critical for mission-level accuracy.

\begin{table*}[t]
    \centering
    \caption{Exp. 1: Results on \textbf{\change} dataset. \textcolor{red}{\textbf{Red}} indicates the best result and \textcolor{blue}{blue} indicates the second best.}
    \label{tab:exp1_classic_change}
    \setlength{\tabcolsep}{1.5pt}
    \renewcommand{\arraystretch}{1.05}
    \resizebox{\textwidth}{!}{%
    \begin{tabular}{c|l| c c c c c c c c c c}
        \hline
        \textbf{Type} & \textbf{Method} & $\delta_{1}$$\uparrow$ & $\delta_{2}$$\uparrow$ & $\delta_{3}$$\uparrow$ & A.Rel$\downarrow$ & Sq.Rel$\downarrow$ & RMSE$\downarrow$ & MAE$\downarrow$ & SILog$\downarrow$ & E.Acc$\downarrow$ & E.Comp$\downarrow$ \\
        \hline
        \multirow{10}{*}{\rotatebox[origin=c]{90}{\textbf{Metric}}}
        & \davft{} & 0.92 & \textcolor{red}{\textbf{1.00}} & \textcolor{red}{\textbf{1.00}} & 0.08 & 0.08 & 0.70 & 0.51 & 10.45 & 3.34 & 401.86 \\
        \cline{2-12}
        & \dav{} \cite{yang_depth_2024} & \textcolor{blue}{0.95} & \textcolor{red}{\textbf{1.00}} & \textcolor{red}{\textbf{1.00}} & \textcolor{blue}{0.06} & 0.05 & 0.56 & 0.41 & 8.30 & 3.42 & 71.96 \\
        & \davthree{} \cite{depthanything3} & \textcolor{red}{\textbf{0.96}} & \textcolor{red}{\textbf{1.00}} & \textcolor{red}{\textbf{1.00}} & \textcolor{red}{\textbf{0.05}} & \textcolor{red}{\textbf{0.03}} & \textcolor{blue}{0.44} & \textcolor{red}{\textbf{0.30}} & 7.77 & 3.00 & 242.66 \\
        & \moge{} \cite{wang2025moge2} & \textcolor{blue}{0.95} & \textcolor{red}{\textbf{1.00}} & \textcolor{red}{\textbf{1.00}} & \textcolor{blue}{0.06} & \textcolor{blue}{0.04} & 0.45 & 0.32 & 7.40 & \textcolor{red}{\textbf{2.76}} & 126.03 \\
        & \depthpro{} \cite{bochkovskii_depth_2024} & 0.89 & 0.98 & \textcolor{red}{\textbf{1.00}} & 0.10 & 0.17 & 0.89 & 0.67 & 13.51 & \textcolor{blue}{2.85} & 204.48 \\
        & \metricd{} \cite{hu_metric3dv2_2024} & \textcolor{blue}{0.95} & \textcolor{red}{\textbf{1.00}} & \textcolor{red}{\textbf{1.00}} & \textcolor{red}{\textbf{0.05}} & \textcolor{blue}{0.04} & \textcolor{red}{\textbf{0.43}} & \textcolor{red}{\textbf{0.30}} & \textcolor{red}{\textbf{7.12}} & 3.27 & \textcolor{red}{\textbf{30.09}} \\
        & \unidepth{} \cite{piccinelli_unidepthv2_2025} & 0.94 & \textcolor{red}{\textbf{1.00}} & \textcolor{red}{\textbf{1.00}} & 0.07 & 0.05 & 0.54 & 0.39 & 8.53 & 3.50 & 93.81 \\
        & \mapanything{} \cite{keetha2025mapanything} & \textcolor{blue}{0.95} & \textcolor{red}{\textbf{1.00}} & \textcolor{red}{\textbf{1.00}} & 0.07 & 0.05 & 0.52 & 0.37 & 9.32 & 3.85 & 130.95 \\
        & \vda{} \cite{video_depth_anything} & 0.93 & \textcolor{red}{\textbf{1.00}} & \textcolor{red}{\textbf{1.00}} & 0.08 & 0.07 & 0.68 & 0.48 & 10.22 & 3.72 & 185.00 \\
        & \metricanything{} \cite{metricanything2026} & \textcolor{blue}{0.95} & \textcolor{red}{\textbf{1.00}} & \textcolor{red}{\textbf{1.00}} & \textcolor{blue}{0.06} & \textcolor{red}{\textbf{0.03}} & \textcolor{blue}{0.44} & \textcolor{blue}{0.31} & \textcolor{blue}{7.22} & 2.90 & 117.60 \\
        \hline
        \multirow{5}{*}{\rotatebox[origin=c]{90}{\textbf{Relative}}}
        & \daac{} \cite{sun2025depth} & 0.66 & 0.90 & \textcolor{blue}{0.99} & 0.22 & 0.43 & 1.55 & 1.25 & 22.98 & 3.79 & 529.25 \\
        & \depthcrafter{} \cite{hu_depthcrafter_2024} & 0.86 & \textcolor{blue}{0.99} & \textcolor{red}{\textbf{1.00}} & 0.13 & 0.16 & 1.05 & 0.84 & 16.20 & 3.32 & 323.26 \\
        & \lotus{} \cite{he_lotus_2025} & 0.70 & 0.93 & \textcolor{blue}{0.99} & 0.19 & 0.42 & 1.64 & 1.28 & 21.77 & 3.84 & 89.76 \\
        & \midas{} \cite{birkl2023midas} & 0.81 & 0.96 & \textcolor{red}{\textbf{1.00}} & 0.15 & 0.22 & 1.12 & 0.88 & 16.89 & 4.15 & 392.09 \\
        & \marigold{} \cite{ke_marigold_2024} & 0.91 & \textcolor{blue}{0.99} & \textcolor{red}{\textbf{1.00}} & 0.09 & 0.09 & 0.69 & 0.52 & 11.40 & 3.40 & \textcolor{blue}{44.27} \\
        \hline
    \end{tabular}%
    }
\end{table*}

\clearpage
\noindent\textbf{\cheri} (\cref{tab:exp1_classic_cheri}). This dark analog benchmark represents the most challenging condition in our entire suite, where all methods suffer a severe performance collapse. The relative method \daac{} achieves the best results ($\delta_1{=}0.54$, A.Rel${=}0.35$, RMSE${=}3.09$), outperforming every metric-based approach. This constitutes a reversal of the dominant trend observed across all other benchmarks. The strongest metric method, \vda{}, reaches only $\delta_1{=}0.43$ and RMSE${=}3.56$. This inversion suggests that the extreme high-contrast polar lighting conditions of \cheri severely disrupt the metric scaling learned from terrestrial data, whereas affine-invariant representations are inherently more resilient as they do not commit to an absolute depth scale. These findings show that illumination robustness remains a critical open challenge for deploying MDE models in harsh extraterrestrial lighting environments.

\begin{table*}[t]
    \centering
    \caption{Exp. 1: Results on \textbf{\cheri} dataset. \textcolor{red}{\textbf{Red}} indicates the best result and \textcolor{blue}{blue} indicates the second best.}
    \label{tab:exp1_classic_cheri}
    \setlength{\tabcolsep}{1.5pt}
    \renewcommand{\arraystretch}{1.05}
    \resizebox{\textwidth}{!}{
    \begin{tabular}{c|l| c c c c c c c c c c}
        \hline
        \textbf{Type} & \textbf{Method} & $\delta_{1}$$\uparrow$ & $\delta_{2}$$\uparrow$ & $\delta_{3}$$\uparrow$ & A.Rel$\downarrow$ & Sq.Rel$\downarrow$ & RMSE$\downarrow$ & MAE$\downarrow$ & SILog$\downarrow$ & E.Acc$\downarrow$ & E.Comp$\downarrow$ \\
        \hline
        \multirow{10}{*}{\rotatebox[origin=c]{90}{\textbf{Metric}}}
        & \davft{} & 0.36 & 0.70 & 0.84 & 0.50 & 2.55 & 3.90 & 3.31 & 50.74 & 2.46 & 84.84 \\
        \cline{2-12}
        & \dav{} \cite{yang_depth_2024} & 0.32 & 0.65 & 0.83 & 0.54 & 2.88 & 4.14 & 3.56 & 50.62 & 3.23 & 94.51 \\
        & \davthree{} \cite{depthanything3} & 0.34 & 0.69 & 0.85 & 0.51 & 2.63 & 3.97 & 3.39 & \textcolor{blue}{48.38} & 2.65 & 69.40 \\
        & \moge{} \cite{wang2025moge2} & 0.32 & 0.65 & 0.83 & 0.54 & 2.84 & 4.12 & 3.54 & 50.92 & 3.24 & 174.46 \\
        & \depthpro{} \cite{bochkovskii_depth_2024} & 0.32 & 0.64 & 0.82 & 0.55 & 2.88 & 4.18 & 3.60 & 51.01 & 3.52 & \textcolor{blue}{41.94} \\
        & \metricd{} \cite{hu_metric3dv2_2024} & 0.33 & 0.67 & 0.84 & 0.53 & 2.83 & 4.09 & 3.51 & 51.84 & 2.99 & 130.94 \\
        & \unidepth{} \cite{piccinelli_unidepthv2_2025} & 0.35 & 0.69 & 0.85 & 0.51 & 2.69 & 3.99 & 3.40 & 50.05 & 2.78 & 136.55 \\
        & \mapanything{} \cite{keetha2025mapanything} & 0.36 & 0.71 & 0.85 & 0.50 & 2.62 & 3.91 & 3.31 & \textcolor{red}{\textbf{48.24}} & 3.69 & 221.03 \\
        & \vda{} \cite{video_depth_anything} & \textcolor{blue}{0.43} & \textcolor{blue}{0.76} & \textcolor{blue}{0.87} & \textcolor{blue}{0.44} & \textcolor{blue}{2.18} & \textcolor{blue}{3.56} & \textcolor{blue}{2.94} & 55.49 & 2.40 & 144.99 \\
        & \metricanything{} \cite{metricanything2026} & 0.32 & 0.65 & 0.83 & 0.55 & 2.91 & 4.17 & 3.59 & 50.94 & 3.54 & 164.35 \\
        \hline
        \multirow{5}{*}{\rotatebox[origin=c]{90}{\textbf{Relative}}}
        & \daac{} \cite{sun2025depth} & \textcolor{red}{\textbf{0.54}} & \textcolor{red}{\textbf{0.83}} & \textcolor{red}{\textbf{0.91}} & \textcolor{red}{\textbf{0.35}} & \textcolor{red}{\textbf{1.60}} & \textcolor{red}{\textbf{3.09}} & \textcolor{red}{\textbf{2.42}} & 98.96 & \textcolor{red}{\textbf{1.67}} & 75.36 \\
        & \depthcrafter{} \cite{hu_depthcrafter_2024} & 0.33 & 0.64 & 0.83 & 0.53 & 2.75 & 4.08 & 3.52 & 50.40 & 5.21 & 74.63 \\
        & \lotus{} \cite{he_lotus_2025} & 0.35 & 0.68 & 0.85 & 0.49 & 2.47 & 3.90 & 3.32 & 54.78 & 3.26 & 46.54 \\
        & \midas{} \cite{birkl2023midas} & 0.40 & 0.74 & 0.86 & 0.46 & 2.35 & 3.71 & 3.10 & 52.27 & \textcolor{blue}{2.39} & 97.14 \\
        & \marigold{} \cite{ke_marigold_2024} & 0.38 & 0.71 & 0.86 & 0.47 & 2.30 & 3.75 & 3.16 & 49.41 & 2.69 & \textcolor{red}{\textbf{9.52}} \\
        \hline
    \end{tabular}
    }
\end{table*}


\clearpage
\subsection{Shaded Region Evaluation}
\label{sec:supp_shaded}

\Cref{tab:exp2_shaded_lunarsim,tab:exp2_shaded_lusnar,tab:exp2_shaded_etna,tab:exp2_shaded_s3li,tab:exp2_shaded_change,tab:exp2_shaded_cheri} report performance exclusively within shaded regions, extending the main text's shadow analysis (Exp.~2) with full per-dataset breakdowns. In the absence of atmospheric scattering on the lunar surface, shadows exhibit near-zero illumination with razor-sharp boundaries, a condition fundamentally distinct from terrestrial shadow patterns. These tables quantify how each architecture handles this extreme loss of photometric information, which is essential for assessing safe autonomous navigation through permanently shadowed craters and polar terrain.\\

\noindent\textbf{\lunarsim} (\cref{tab:exp2_shaded_lunarsim}). \mapanything{} and \davft{} share the lead with $\delta_1{=}0.80$, though \mapanything{} achieves a lower A.Rel ($0.13$ vs.\ $0.15$) and considerably better RMSE ($11.48$ vs.\ $15.10$). Compared to the overall evaluation ($\delta_1{=}0.87$ for \mapanything{}), the degradation is moderate ($\Delta\delta_1{\approx}0.07$), suggesting that the synthetic shading model preserves sufficient texture cues for the stronger architectures. Among relative methods, \midas{} drops from $\delta_1{=}0.78$ to $0.73$, indicating comparable resilience. This relatively mild degradation further suggests that synthetic shadow rendering, while useful for controlled evaluation, may underestimate the severity of real vacuum shadows.

\begin{table*}[t]
    \centering
    \caption{Exp. 2: Results on \textbf{\lunarsim} dataset (shaded regions). \textcolor{red}{\textbf{Red}} indicates the best result and \textcolor{blue}{blue} indicates the second best.}
    \label{tab:exp2_shaded_lunarsim}
    \setlength{\tabcolsep}{1.5pt}
    \renewcommand{\arraystretch}{1.05}
    \resizebox{\textwidth}{!}{%
    \begin{tabular}{c|l| c c c c c c c c c c}
        \hline
        \textbf{Type} & \textbf{Method} & $\delta_{1}$$\uparrow$ & $\delta_{2}$$\uparrow$ & $\delta_{3}$$\uparrow$ & A.Rel$\downarrow$ & Sq.Rel$\downarrow$ & RMSE$\downarrow$ & MAE$\downarrow$ & SILog$\downarrow$ & E.Acc$\downarrow$ & E.Comp$\downarrow$ \\
        \hline
        \multirow{10}{*}{\rotatebox[origin=c]{90}{\textbf{Metric}}}
        & \davft{} & \textcolor{blue}{0.79} & \textcolor{red}{\textbf{0.97}} & \textcolor{red}{\textbf{0.99}} & \textcolor{blue}{0.15} & 3.23 & 15.10 & 10.58 & \textcolor{red}{\textbf{12.90}} & \textcolor{red}{\textbf{2.70}} & 71.79 \\
        \cline{2-12}
        & \dav{} \cite{yang_depth_2024} & 0.47 & 0.72 & 0.84 & 0.29 & 4.33 & 16.39 & 12.44 & 75.94 & 4.03 & 96.96 \\
        & \davthree{} \cite{depthanything3} & 0.54 & 0.75 & 0.86 & 0.26 & 3.67 & 15.77 & 11.29 & 50.73 & 3.52 & 125.68 \\
        & \moge{} \cite{wang2025moge2} & 0.68 & 0.90 & 0.96 & 0.17 & 2.79 & 14.42 & 10.28 & 20.76 & \textcolor{blue}{2.71} & 76.19 \\
        & \depthpro{} \cite{bochkovskii_depth_2024} & 0.38 & 0.62 & 0.75 & 0.35 & 6.84 & 21.33 & 15.72 & 91.85 & 4.09 & 116.75 \\
        & \metricd{} \cite{hu_metric3dv2_2024} & 0.62 & 0.85 & 0.94 & 0.20 & 3.11 & 14.35 & 9.59 & 25.92 & 3.32 & \textcolor{red}{\textbf{41.03}} \\
        & \unidepth{} \cite{piccinelli_unidepthv2_2025} & 0.70 & 0.86 & 0.92 & 0.23 & 12.40 & 24.12 & 14.45 & 18.86 & 2.80 & 118.18 \\
        & \mapanything{} \cite{keetha2025mapanything} & \textcolor{red}{\textbf{0.80}} & \textcolor{blue}{0.95} & \textcolor{red}{\textbf{0.99}} & \textcolor{red}{\textbf{0.13}} & \textcolor{red}{\textbf{1.77}} & \textcolor{red}{\textbf{11.48}} & \textcolor{red}{\textbf{7.25}} & 30.21 & 2.92 & 115.59 \\
        & \vda{} \cite{video_depth_anything} & 0.66 & 0.90 & 0.96 & 0.19 & 3.96 & 16.31 & 9.72 & 20.26 & 3.25 & 65.24 \\
        & \metricanything{} \cite{metricanything2026} & 0.74 & 0.92 & 0.96 & 0.16 & \textcolor{blue}{2.39} & \textcolor{blue}{13.19} & \textcolor{blue}{9.40} & 24.83 & 2.99 & 65.14 \\
        \hline
        \multirow{5}{*}{\rotatebox[origin=c]{90}{\textbf{Relative}}}
        & \daac{} \cite{sun2025depth} & 0.63 & 0.81 & 0.87 & 0.32 & 14.68 & 27.04 & 18.49 & 21.41 & 2.98 & 102.68 \\
        & \depthcrafter{} \cite{hu_depthcrafter_2024} & 0.36 & 0.58 & 0.67 & 0.45 & 8.63 & 22.22 & 17.53 & 335.15 & 4.24 & 155.96 \\
        & \lotus{} \cite{he_lotus_2025} & 0.46 & 0.77 & 0.91 & 0.32 & 12.83 & 29.36 & 19.24 & 27.68 & 3.23 & 64.85 \\
        & \midas{} \cite{birkl2023midas} & 0.73 & 0.93 & \textcolor{blue}{0.98} & 0.19 & 4.19 & 16.70 & 11.39 & \textcolor{blue}{16.71} & 2.75 & 113.15 \\
        & \marigold{} \cite{ke_marigold_2024} & 0.51 & 0.79 & 0.90 & 0.29 & 7.52 & 21.92 & 16.57 & 36.53 & 3.05 & \textcolor{blue}{44.39} \\
        \hline
    \end{tabular}%
    }
\end{table*}

\clearpage
\noindent\textbf{\lusnar} (\cref{tab:exp2_shaded_lusnar}). \davft{} maintains clear dominance ($\delta_1{=}0.84$, A.Rel${=}0.11$, RMSE${=}0.95$), benefiting from its explicit finetuning on \lusnar geometry. The gap to the second-best method widens compared to the overall setting: \vda{} reaches only $\delta_1{=}0.65$, and \mapanything{} follows with RMSE${=}1.64$. Among relative methods, \midas{} ($\delta_1{=}0.64$, A.Rel${=}0.20$) outperforms \daac{}, which drops sharply to $\delta_1{=}0.54$. The amplified performance gap under shading suggests that domain-adapted models internalize geometric structure more deeply than photometric appearance, whereas zero-shot models increasingly rely on surface texture that is absent in shadowed regions.

\begin{table*}[t]
    \centering
    \caption{Exp. 2: Results on \textbf{\lusnar} dataset (shaded regions). \textcolor{red}{\textbf{Red}} indicates the best result and \textcolor{blue}{blue} indicates the second best.}
    \label{tab:exp2_shaded_lusnar}
    \setlength{\tabcolsep}{1.5pt}
    \renewcommand{\arraystretch}{1.05}
    \resizebox{\textwidth}{!}{%
    \begin{tabular}{c|l| c c c c c c c c c c}
        \hline
        \textbf{Type} & \textbf{Method} & $\delta_{1}$$\uparrow$ & $\delta_{2}$$\uparrow$ & $\delta_{3}$$\uparrow$ & A.Rel$\downarrow$ & Sq.Rel$\downarrow$ & RMSE$\downarrow$ & MAE$\downarrow$ & SILog$\downarrow$ & E.Acc$\downarrow$ & E.Comp$\downarrow$ \\
        \hline
        \multirow{10}{*}{\rotatebox[origin=c]{90}{\textbf{Metric}}}
        & \davft{} & \textcolor{red}{\textbf{0.84}} & \textcolor{red}{\textbf{0.96}} & \textcolor{red}{\textbf{1.00}} & \textcolor{red}{\textbf{0.11}} & \textcolor{red}{\textbf{0.10}} & \textcolor{red}{\textbf{0.95}} & \textcolor{red}{\textbf{0.49}} & \textcolor{red}{\textbf{7.94}} & \textcolor{red}{\textbf{1.57}} & \textcolor{red}{\textbf{27.25}} \\
        \cline{2-12}
        & \dav{} \cite{yang_depth_2024} & 0.20 & 0.37 & 0.45 & 0.66 & 1.39 & 2.86 & 2.11 & 407.91 & 7.59 & 62.95 \\
        & \davthree{} \cite{depthanything3} & 0.34 & 0.54 & 0.68 & 0.40 & 0.64 & 2.15 & 1.39 & 131.02 & 3.50 & 42.53 \\
        & \moge{} \cite{wang2025moge2} & 0.35 & 0.53 & 0.65 & 0.41 & 0.59 & 1.90 & 1.31 & 147.67 & 3.34 & 44.50 \\
        & \depthpro{} \cite{bochkovskii_depth_2024} & 0.23 & 0.39 & 0.46 & 0.68 & 1.40 & 2.83 & 2.04 & 430.27 & 7.58 & 67.47 \\
        & \metricd{} \cite{hu_metric3dv2_2024} & 0.39 & 0.57 & 0.69 & 0.38 & 0.57 & 1.92 & 1.30 & 120.21 & 3.56 & 52.68 \\
        & \unidepth{} \cite{piccinelli_unidepthv2_2025} & 0.45 & 0.67 & 0.79 & 0.36 & 1.12 & 2.85 & 1.43 & 78.12 & 2.65 & 51.20 \\
        & \mapanything{} \cite{keetha2025mapanything} & 0.60 & 0.81 & 0.90 & 0.22 & \textcolor{blue}{0.32} & \textcolor{blue}{1.64} & \textcolor{blue}{0.95} & 65.72 & 2.67 & 51.88 \\
        & \vda{} \cite{video_depth_anything} & \textcolor{blue}{0.65} & 0.88 & \textcolor{blue}{0.96} & 0.23 & 0.50 & 2.19 & 1.17 & 22.18 & \textcolor{blue}{2.04} & \textcolor{blue}{41.74} \\
        & \metricanything{} \cite{metricanything2026} & 0.32 & 0.50 & 0.63 & 0.42 & 0.62 & 2.00 & 1.38 & 147.60 & 3.66 & 50.11 \\
        \hline
        \multirow{5}{*}{\rotatebox[origin=c]{90}{\textbf{Relative}}}
        & \daac{} \cite{sun2025depth} & 0.54 & 0.76 & 0.87 & 0.38 & 1.95 & 3.80 & 1.73 & 22.53 & 2.24 & 47.40 \\
        & \depthcrafter{} \cite{hu_depthcrafter_2024} & 0.28 & 0.44 & 0.52 & 0.81 & 2.16 & 3.17 & 2.28 & 523.05 & 7.91 & 70.24 \\
        & \lotus{} \cite{he_lotus_2025} & 0.28 & 0.55 & 0.68 & 0.64 & 2.11 & 3.94 & 2.32 & 154.32 & 3.86 & 64.37 \\
        & \midas{} \cite{birkl2023midas} & 0.64 & \textcolor{blue}{0.89} & 0.95 & \textcolor{blue}{0.20} & 0.45 & 1.99 & 1.19 & \textcolor{blue}{16.60} & 2.12 & 45.19 \\
        & \marigold{} \cite{ke_marigold_2024} & 0.43 & 0.60 & 0.69 & 0.37 & 0.75 & 2.40 & 1.49 & 82.79 & 3.49 & 50.92 \\
        \hline
    \end{tabular}%
    }
\end{table*}

\clearpage
\noindent\textbf{\etna} (\cref{tab:exp2_shaded_etna}). Performance remains near-perfect for the leading methods, with six models attaining $\delta_1{=}1.00$. \mapanything{} and \unidepth{} share the best A.Rel of $0.03$, while \mapanything{} achieves the lowest RMSE ($0.04$). The controlled analog environment retains sufficient illumination variation for robust depth recovery even within shadows. As with the overall evaluation, the well-lit terrestrial nature of \etna renders its shaded split too permissive to expose meaningful robustness differences, further confirming that this analog provides limited discriminative power for benchmarking extraterrestrial shadow handling.

\begin{table*}[t]
    \centering
    \caption{Exp. 2: Results on \textbf{\etna} dataset (shaded regions). \textcolor{red}{\textbf{Red}} indicates the best result and \textcolor{blue}{blue} indicates the second best.}
    \label{tab:exp2_shaded_etna}
    \setlength{\tabcolsep}{1.5pt}
    \renewcommand{\arraystretch}{1.05}
    \resizebox{\textwidth}{!}{%
    \begin{tabular}{c|l| c c c c c c c c c c}
        \hline
        \textbf{Type} & \textbf{Method} & $\delta_{1}$$\uparrow$ & $\delta_{2}$$\uparrow$ & $\delta_{3}$$\uparrow$ & A.Rel$\downarrow$ & Sq.Rel$\downarrow$ & RMSE$\downarrow$ & MAE$\downarrow$ & SILog$\downarrow$ & E.Acc$\downarrow$ & E.Comp$\downarrow$ \\
        \hline
        \multirow{10}{*}{\rotatebox[origin=c]{90}{\textbf{Metric}}}
        & \davft{} & \textcolor{red}{\textbf{1.00}} & \textcolor{red}{\textbf{1.00}} & \textcolor{red}{\textbf{1.00}} & 0.06 & \textcolor{blue}{0.01} & 0.10 & 0.09 & 1.25 & 2.85 & 37.38 \\
        \cline{2-12}
        & \dav{} \cite{yang_depth_2024} & \textcolor{blue}{0.99} & \textcolor{red}{\textbf{1.00}} & \textcolor{red}{\textbf{1.00}} & 0.07 & \textcolor{blue}{0.01} & 0.12 & 0.12 & 2.24 & 2.60 & 12.35 \\
        & \davthree{} \cite{depthanything3} & \textcolor{red}{\textbf{1.00}} & \textcolor{red}{\textbf{1.00}} & \textcolor{red}{\textbf{1.00}} & \textcolor{blue}{0.05} & \textcolor{red}{\textbf{0.00}} & 0.07 & \textcolor{blue}{0.07} & 1.50 & \textcolor{blue}{2.55} & 27.67 \\
        & \moge{} \cite{wang2025moge2} & \textcolor{blue}{0.99} & \textcolor{red}{\textbf{1.00}} & \textcolor{red}{\textbf{1.00}} & \textcolor{blue}{0.05} & \textcolor{blue}{0.01} & 0.08 & 0.08 & 1.61 & 2.81 & 21.85 \\
        & \depthpro{} \cite{bochkovskii_depth_2024} & 0.39 & 0.83 & \textcolor{blue}{0.92} & 559.37 & 5590194.98 & 818.97 & 818.96 & 4.29 & 4.19 & 34.97 \\
        & \metricd{} \cite{hu_metric3dv2_2024} & \textcolor{red}{\textbf{1.00}} & \textcolor{red}{\textbf{1.00}} & \textcolor{red}{\textbf{1.00}} & \textcolor{blue}{0.05} & \textcolor{red}{\textbf{0.00}} & 0.07 & \textcolor{blue}{0.07} & 1.51 & 2.69 & 23.61 \\
        & \unidepth{} \cite{piccinelli_unidepthv2_2025} & \textcolor{red}{\textbf{1.00}} & \textcolor{red}{\textbf{1.00}} & \textcolor{red}{\textbf{1.00}} & \textcolor{red}{\textbf{0.03}} & \textcolor{red}{\textbf{0.00}} & \textcolor{blue}{0.05} & \textcolor{red}{\textbf{0.04}} & \textcolor{blue}{1.06} & 2.62 & 9.91 \\
        & \mapanything{} \cite{keetha2025mapanything} & \textcolor{red}{\textbf{1.00}} & \textcolor{red}{\textbf{1.00}} & \textcolor{red}{\textbf{1.00}} & \textcolor{red}{\textbf{0.03}} & \textcolor{red}{\textbf{0.00}} & \textcolor{red}{\textbf{0.04}} & \textcolor{red}{\textbf{0.04}} & \textcolor{red}{\textbf{0.95}} & 2.68 & 21.65 \\
        & \vda{} \cite{video_depth_anything} & \textcolor{red}{\textbf{1.00}} & \textcolor{red}{\textbf{1.00}} & \textcolor{red}{\textbf{1.00}} & 0.06 & \textcolor{blue}{0.01} & 0.09 & 0.09 & 1.34 & \textcolor{red}{\textbf{2.42}} & 16.69 \\
        & \metricanything{} \cite{metricanything2026} & \textcolor{blue}{0.99} & \textcolor{red}{\textbf{1.00}} & \textcolor{red}{\textbf{1.00}} & 0.07 & \textcolor{blue}{0.01} & 0.12 & 0.11 & 2.12 & 2.80 & 15.29 \\
        \hline
        \multirow{5}{*}{\rotatebox[origin=c]{90}{\textbf{Relative}}}
        & \daac{} \cite{sun2025depth} & \textcolor{red}{\textbf{1.00}} & \textcolor{red}{\textbf{1.00}} & \textcolor{red}{\textbf{1.00}} & 0.09 & \textcolor{blue}{0.01} & 0.13 & 0.12 & 1.48 & 2.81 & 40.81 \\
        & \depthcrafter{} \cite{hu_depthcrafter_2024} & 0.89 & \textcolor{red}{\textbf{1.00}} & \textcolor{red}{\textbf{1.00}} & 0.11 & 0.03 & 0.18 & 0.17 & 4.06 & 3.27 & 26.18 \\
        & \lotus{} \cite{he_lotus_2025} & 0.45 & \textcolor{blue}{0.87} & \textcolor{red}{\textbf{1.00}} & 0.29 & 0.17 & 0.42 & 0.41 & 5.76 & 2.89 & \textcolor{blue}{9.48} \\
        & \midas{} \cite{birkl2023midas} & \textcolor{blue}{0.99} & \textcolor{red}{\textbf{1.00}} & \textcolor{red}{\textbf{1.00}} & 0.07 & \textcolor{blue}{0.01} & 0.10 & 0.10 & 1.34 & 2.83 & 47.48 \\
        & \marigold{} \cite{ke_marigold_2024} & 0.90 & \textcolor{red}{\textbf{1.00}} & \textcolor{red}{\textbf{1.00}} & 0.11 & 0.03 & 0.17 & 0.16 & 3.15 & 2.62 & \textcolor{red}{\textbf{8.27}} \\
        \hline
    \end{tabular}%
    }
\end{table*}

\clearpage
\noindent\textbf{\seli} (\cref{tab:exp2_shaded_s3li}). This dataset reveals a notable narrowing of the metric--relative gap under shading. \mapanything{} leads with $\delta_1{=}0.80$ and RMSE${=}1.63$, but \midas{} closely follows with $\delta_1{=}0.79$ and a matching A.Rel of $0.15$. This convergence is significant: it suggests that when shadow-induced contrast loss eliminates the fine photometric details on which metric models rely for scale inference, relative methods, which are inherently scale-agnostic, can achieve competitive geometric accuracy. This observation has practical implications for mission planning in persistently shadowed regions, where relative depth estimation may offer a more robust fallback than metric architectures.

\begin{table*}[t]
    \centering
    \caption{Exp. 2: Results on \textbf{\seli} dataset (shaded regions). \textcolor{red}{\textbf{Red}} indicates the best result and \textcolor{blue}{blue} indicates the second best.}
    \label{tab:exp2_shaded_s3li}
    \setlength{\tabcolsep}{1.5pt}
    \renewcommand{\arraystretch}{1.05}
    \resizebox{\textwidth}{!}{%
    \begin{tabular}{c|l| c c c c c c c c}
        \hline
        \textbf{Type} & \textbf{Method} & $\delta_{1}$$\uparrow$ & $\delta_{2}$$\uparrow$ & $\delta_{3}$$\uparrow$ & A.Rel$\downarrow$ & Sq.Rel$\downarrow$ & RMSE$\downarrow$ & MAE$\downarrow$ & SILog$\downarrow$ \\
        \hline
        \multirow{10}{*}{\rotatebox[origin=c]{90}{\textbf{Metric}}}
        & \davft{} & 0.46 & 0.70 & 0.84 & 0.29 & 1.14 & 2.88 & 2.59 & 15.79 \\
        \cline{2-10}
        & \dav{} \cite{yang_depth_2024} & 0.47 & 0.63 & 0.68 & 0.43 & 2.36 & 3.54 & 3.27 & 109.25 \\
        & \davthree{} \cite{depthanything3} & 0.53 & 0.65 & 0.68 & 0.46 & 2.90 & 3.58 & 3.30 & 80.69 \\
        & \moge{} \cite{wang2025moge2} & 0.56 & 0.69 & 0.73 & 0.38 & 2.30 & 3.22 & 2.92 & 95.08 \\
        & \depthpro{} \cite{bochkovskii_depth_2024} & 0.59 & 0.79 & 0.87 & 0.25 & 0.99 & 2.36 & 2.11 & 25.29 \\
        & \metricd{} \cite{hu_metric3dv2_2024} & 0.68 & 0.87 & 0.93 & 0.23 & 1.01 & 2.35 & 1.95 & 91.02 \\
        & \unidepth{} \cite{piccinelli_unidepthv2_2025} & 0.75 & \textcolor{blue}{0.93} & \textcolor{blue}{0.98} & \textcolor{blue}{0.17} & 0.65 & 1.90 & 1.66 & 10.68 \\
        & \mapanything{} \cite{keetha2025mapanything} & \textcolor{red}{\textbf{0.80}} & \textcolor{red}{\textbf{0.95}} & \textcolor{blue}{0.98} & \textcolor{red}{\textbf{0.15}} & \textcolor{red}{\textbf{0.47}} & \textcolor{red}{\textbf{1.63}} & \textcolor{red}{\textbf{1.40}} & \textcolor{red}{\textbf{9.75}} \\
        & \vda{} \cite{video_depth_anything} & 0.48 & 0.71 & 0.85 & 0.28 & 0.99 & 2.61 & 2.34 & 14.30 \\
        & \metricanything{} \cite{metricanything2026} & 0.68 & 0.85 & 0.90 & 0.23 & 1.04 & 2.22 & 1.96 & 33.42 \\
        \hline
        \multirow{5}{*}{\rotatebox[origin=c]{90}{\textbf{Relative}}}
        & \daac{} \cite{sun2025depth} & 0.62 & 0.90 & 0.97 & 0.20 & 0.63 & 2.13 & 1.87 & 11.47 \\
        & \depthcrafter{} \cite{hu_depthcrafter_2024} & 0.46 & 0.67 & 0.75 & 0.38 & 1.93 & 3.33 & 3.02 & 62.79 \\
        & \lotus{} \cite{he_lotus_2025} & 0.48 & 0.70 & 0.79 & 0.32 & 1.49 & 3.17 & 2.83 & 25.22 \\
        & \midas{} \cite{birkl2023midas} & \textcolor{blue}{0.79} & \textcolor{red}{\textbf{0.95}} & \textcolor{red}{\textbf{0.99}} & \textcolor{red}{\textbf{0.15}} & \textcolor{blue}{0.49} & \textcolor{blue}{1.75} & \textcolor{blue}{1.52} & \textcolor{blue}{9.79} \\
        & \marigold{} \cite{ke_marigold_2024} & 0.49 & 0.67 & 0.75 & 0.35 & 1.65 & 3.21 & 2.87 & 62.92 \\
        \hline
    \end{tabular}%
    }
\end{table*}

\clearpage
\noindent\textbf{\change} (\cref{tab:exp2_shaded_change}). Four metric methods, \dav{}, \davthree{}, \metricd{}, and \metricanything{}, tie at $\delta_1{=}0.94$. \metricd{} achieves the lowest RMSE ($0.44$) and \davthree{} the lowest A.Rel ($0.06$). The shaded performance is remarkably close to the overall results, indicating that the authentic Chang'e-3 imagery, captured under the harsh, unfiltered solar illumination of the lunar surface, already inherently contains substantial high-contrast shadow content. Consequently, the shadow-masked split introduces minimal additional difficulty, and the per-method rankings remain largely consistent with the overall evaluation.

\begin{table*}[t]
    \centering
    \caption{Exp. 2: Results on \textbf{\change} dataset (shaded regions). \textcolor{red}{\textbf{Red}} indicates the best result and \textcolor{blue}{blue} indicates the second best.}
    \label{tab:exp2_shaded_change}
    \setlength{\tabcolsep}{1.5pt}
    \renewcommand{\arraystretch}{1.05}
    \resizebox{\textwidth}{!}{%
    \begin{tabular}{c|l| c c c c c c c c c c}
        \hline
        \textbf{Type} & \textbf{Method} & $\delta_{1}$$\uparrow$ & $\delta_{2}$$\uparrow$ & $\delta_{3}$$\uparrow$ & A.Rel$\downarrow$ & Sq.Rel$\downarrow$ & RMSE$\downarrow$ & MAE$\downarrow$ & SILog$\downarrow$ & E.Acc$\downarrow$ & E.Comp$\downarrow$ \\
        \hline
        \multirow{10}{*}{\rotatebox[origin=c]{90}{\textbf{Metric}}}
        & \davft{} & 0.89 & \textcolor{blue}{0.99} & \textcolor{red}{\textbf{1.00}} & 0.10 & 0.10 & 0.73 & 0.55 & 10.59 & 3.74 & 241.28 \\
        \cline{2-12}
        & \dav{} \cite{yang_depth_2024} & \textcolor{red}{\textbf{0.94}} & \textcolor{red}{\textbf{1.00}} & \textcolor{red}{\textbf{1.00}} & \textcolor{blue}{0.07} & 0.06 & 0.59 & 0.44 & 8.36 & 3.43 & 50.51 \\
        & \davthree{} \cite{depthanything3} & \textcolor{red}{\textbf{0.94}} & \textcolor{red}{\textbf{1.00}} & \textcolor{red}{\textbf{1.00}} & \textcolor{red}{\textbf{0.06}} & \textcolor{red}{\textbf{0.04}} & \textcolor{blue}{0.45} & \textcolor{red}{\textbf{0.32}} & 9.67 & 3.09 & 119.34 \\
        & \moge{} \cite{wang2025moge2} & \textcolor{blue}{0.93} & \textcolor{red}{\textbf{1.00}} & \textcolor{red}{\textbf{1.00}} & \textcolor{blue}{0.07} & \textcolor{blue}{0.05} & 0.46 & 0.34 & 7.81 & \textcolor{red}{\textbf{2.71}} & 95.83 \\
        & \depthpro{} \cite{bochkovskii_depth_2024} & 0.85 & 0.98 & \textcolor{red}{\textbf{1.00}} & 0.11 & 0.18 & 0.87 & 0.69 & 15.04 & 3.08 & 139.39 \\
        & \metricd{} \cite{hu_metric3dv2_2024} & \textcolor{red}{\textbf{0.94}} & \textcolor{red}{\textbf{1.00}} & \textcolor{red}{\textbf{1.00}} & \textcolor{red}{\textbf{0.06}} & \textcolor{red}{\textbf{0.04}} & \textcolor{red}{\textbf{0.44}} & \textcolor{red}{\textbf{0.32}} & \textcolor{red}{\textbf{7.41}} & \textcolor{red}{\textbf{2.71}} & \textcolor{red}{\textbf{44.76}} \\
        & \unidepth{} \cite{piccinelli_unidepthv2_2025} & 0.92 & \textcolor{red}{\textbf{1.00}} & \textcolor{red}{\textbf{1.00}} & 0.08 & 0.06 & 0.54 & 0.41 & 8.75 & 3.12 & 115.59 \\
        & \mapanything{} \cite{keetha2025mapanything} & \textcolor{blue}{0.93} & \textcolor{red}{\textbf{1.00}} & \textcolor{red}{\textbf{1.00}} & 0.08 & 0.06 & 0.55 & 0.40 & 13.22 & 3.58 & 187.74 \\
        & \vda{} \cite{video_depth_anything} & 0.90 & \textcolor{red}{\textbf{1.00}} & \textcolor{red}{\textbf{1.00}} & 0.10 & 0.09 & 0.68 & 0.51 & 10.33 & 3.67 & 176.46 \\
        & \metricanything{} \cite{metricanything2026} & \textcolor{red}{\textbf{0.94}} & \textcolor{red}{\textbf{1.00}} & \textcolor{red}{\textbf{1.00}} & \textcolor{red}{\textbf{0.06}} & \textcolor{red}{\textbf{0.04}} & 0.46 & \textcolor{blue}{0.33} & \textcolor{blue}{7.48} & \textcolor{blue}{2.80} & 93.81 \\
        \hline
        \multirow{5}{*}{\rotatebox[origin=c]{90}{\textbf{Relative}}}
        & \daac{} \cite{sun2025depth} & 0.58 & 0.85 & 0.98 & 0.26 & 0.58 & 1.68 & 1.34 & 21.47 & 4.64 & 417.57 \\
        & \depthcrafter{} \cite{hu_depthcrafter_2024} & 0.83 & 0.97 & \textcolor{blue}{0.99} & 0.14 & 0.20 & 1.11 & 0.87 & 15.25 & 4.03 & 256.90 \\
        & \lotus{} \cite{he_lotus_2025} & 0.63 & 0.90 & 0.98 & 0.23 & 0.64 & 1.81 & 1.41 & 21.26 & 3.67 & 97.65 \\
        & \midas{} \cite{birkl2023midas} & 0.74 & 0.94 & \textcolor{red}{\textbf{1.00}} & 0.18 & 0.27 & 1.17 & 0.93 & 16.48 & 4.41 & 404.92 \\
        & \marigold{} \cite{ke_marigold_2024} & 0.89 & \textcolor{blue}{0.99} & \textcolor{red}{\textbf{1.00}} & 0.10 & 0.12 & 0.72 & 0.56 & 10.23 & 3.34 & \textcolor{blue}{45.43} \\
        \hline
    \end{tabular}%
    }
\end{table*}

\clearpage
\noindent\textbf{\cheri} (\cref{tab:exp2_shaded_cheri}). This condition represents the hardest evaluation in our entire benchmark suite. \daac{} again leads ($\delta_1{=}0.47$, A.Rel${=}0.41$, RMSE${=}3.96$), though with degraded performance compared to its overall result ($\delta_1{=}0.54$). Metric methods deteriorate severely, with the best (\vda{}) reaching only $\delta_1{=}0.27$. The compounded difficulty of the dark analog terrain and deep shadows reduces many metric architectures to near-random performance, exposing a fundamental limitation: terrestrial metric priors, which implicitly encode assumptions about Earth-like illumination distributions, become unreliable under the extreme photometric conditions of polar and permanently shadowed lunar environments. Addressing this failure mode is essential for enabling safe autonomous navigation during future missions targeting the lunar south pole.

\begin{table*}[t]
    \centering
    \caption{Exp. 2: Results on \textbf{\cheri} dataset (shaded regions). \textcolor{red}{\textbf{Red}} indicates the best result and \textcolor{blue}{blue} indicates the second best.}
    \label{tab:exp2_shaded_cheri}
    \setlength{\tabcolsep}{1.5pt}
    \renewcommand{\arraystretch}{1.05}
    \resizebox{\textwidth}{!}{%
    \begin{tabular}{c|l| c c c c c c c c c c}
        \hline
        \textbf{Type} & \textbf{Method} & $\delta_{1}$$\uparrow$ & $\delta_{2}$$\uparrow$ & $\delta_{3}$$\uparrow$ & A.Rel$\downarrow$ & Sq.Rel$\downarrow$ & RMSE$\downarrow$ & MAE$\downarrow$ & SILog$\downarrow$ & E.Acc$\downarrow$ & E.Comp$\downarrow$ \\
        \hline
        \multirow{10}{*}{\rotatebox[origin=c]{90}{\textbf{Metric}}}
        & \davft{} & 0.16 & 0.50 & \textcolor{blue}{0.82} & \textcolor{blue}{0.51} & \textcolor{blue}{2.51} & 4.76 & 4.27 & 58.39 & 3.28 & 10.11 \\
        \cline{2-12}
        & \dav{} \cite{yang_depth_2024} & 0.15 & 0.39 & 0.76 & 0.72 & 4.12 & 5.18 & 4.81 & 44.18 & 3.69 & 11.32 \\
        & \davthree{} \cite{depthanything3} & 0.18 & 0.49 & 0.79 & 0.62 & 3.17 & 4.75 & 4.37 & \textcolor{red}{\textbf{39.39}} & 3.06 & 9.37 \\
        & \moge{} \cite{wang2025moge2} & 0.15 & 0.40 & 0.76 & 0.73 & 4.26 & 5.20 & 4.83 & 44.93 & 3.01 & 9.48 \\
        & \depthpro{} \cite{bochkovskii_depth_2024} & 0.13 & 0.45 & 0.73 & 0.78 & 4.73 & 5.27 & 4.93 & 47.67 & 3.46 & 9.83 \\
        & \metricd{} \cite{hu_metric3dv2_2024} & 0.16 & 0.43 & 0.78 & 0.68 & 3.83 & 5.06 & 4.67 & 46.47 & 2.88 & 11.23 \\
        & \unidepth{} \cite{piccinelli_unidepthv2_2025} & 0.17 & 0.49 & 0.79 & 0.64 & 3.56 & 4.94 & 4.51 & 45.78 & 2.88 & 8.97 \\
        & \mapanything{} \cite{keetha2025mapanything} & 0.15 & 0.49 & 0.79 & 0.63 & 3.47 & 4.95 & 4.53 & \textcolor{blue}{40.04} & 3.08 & 10.13 \\
        & \vda{} \cite{video_depth_anything} & \textcolor{blue}{0.27} & \textcolor{blue}{0.68} & \textcolor{blue}{0.82} & \textcolor{blue}{0.51} & 2.62 & \textcolor{blue}{4.23} & \textcolor{blue}{3.76} & 68.25 & \textcolor{red}{\textbf{2.70}} & 10.10 \\
        & \metricanything{} \cite{metricanything2026} & 0.14 & 0.40 & 0.75 & 0.74 & 4.33 & 5.24 & 4.88 & 45.18 & 2.98 & 11.05 \\
        \hline
        \multirow{5}{*}{\rotatebox[origin=c]{90}{\textbf{Relative}}}
        & \daac{} \cite{sun2025depth} & \textcolor{red}{\textbf{0.47}} & \textcolor{red}{\textbf{0.76}} & \textcolor{red}{\textbf{0.83}} & \textcolor{red}{\textbf{0.41}} & 2.52 & \textcolor{red}{\textbf{3.96}} & \textcolor{red}{\textbf{3.05}} & 212.22 & \textcolor{blue}{2.72} & 10.43 \\
        & \depthcrafter{} \cite{hu_depthcrafter_2024} & 0.21 & 0.47 & 0.73 & 0.76 & 4.59 & 5.01 & 4.65 & 44.23 & 3.21 & 10.14 \\
        & \lotus{} \cite{he_lotus_2025} & 0.21 & 0.53 & 0.79 & 0.58 & 3.18 & 4.71 & 4.22 & 92.98 & 2.91 & \textcolor{red}{\textbf{5.01}} \\
        & \midas{} \cite{birkl2023midas} & \textcolor{blue}{0.27} & 0.62 & 0.81 & 0.55 & 2.79 & \textcolor{blue}{4.23} & 3.78 & 60.53 & 3.26 & 11.66 \\
        & \marigold{} \cite{ke_marigold_2024} & 0.18 & 0.47 & 0.81 & \textcolor{blue}{0.51} & \textcolor{red}{\textbf{2.40}} & 4.65 & 4.20 & 53.26 & 2.83 & \textcolor{blue}{6.37} \\
        \hline
    \end{tabular}%
    }
\end{table*}


\clearpage
\subsection{Semantic Region Evaluation}
\label{sec:supp_semantic}

\Cref{tab:exp4_semantic_regolith,tab:exp4_semantic_rock,tab:exp4_semantic_crater} extend the main text's topological analysis by providing the full per-method metric breakdown for each semantic surface category on \lusnar: regolith, rock, and crater. For autonomous lunar rover navigation, accurate depth estimation over each terrain type is critical. Regolith constitutes the majority of traversable surface, rocks represent immediate collision hazards requiring reliable obstacle detection, and craters pose the most dangerous traversal risk due to their steep interior slopes and abrupt rim discontinuities. By leveraging the pixel-level semantic labels of \lusnar, we isolate how well each architecture captures the distinct geometric characteristics of these geological constructs.

\noindent\textbf{Regolith} (\cref{tab:exp4_semantic_regolith}). \davft{} achieves the best performance across all metrics ($\delta_1{=}0.93$, A.Rel${=}0.07$, RMSE${=}0.75$). \midas{} is the strongest relative method ($\delta_1{=}0.81$, A.Rel${=}0.13$). The relatively low error magnitudes across methods reflect the geometric simplicity of flat regolith surfaces, which lack the depth discontinuities and self-occlusion patterns that challenge monocular inference. From a deployment perspective, the consistently high accuracy on regolith is encouraging, as it confirms that current architectures can reliably estimate depth over the primary traversable terrain type.

\begin{table*}[t]
    \centering
    \caption{Exp. 2: Results on \textbf{Regolith} regions on \lusnar dataset. \textcolor{red}{\textbf{Red}} indicates the best result and \textcolor{blue}{blue} indicates the second best.}
    \label{tab:exp4_semantic_regolith}
    \setlength{\tabcolsep}{1.5pt}
    \renewcommand{\arraystretch}{1.05}
    \resizebox{\textwidth}{!}{%
    \begin{tabular}{c|l| c c c c c c c c c c}
        \hline
        \textbf{Type} & \textbf{Method} & $\delta_{1}$$\uparrow$ & $\delta_{2}$$\uparrow$ & $\delta_{3}$$\uparrow$ & A.Rel$\downarrow$ & Sq.Rel$\downarrow$ & RMSE$\downarrow$ & MAE$\downarrow$ & SILog$\downarrow$ & E.Acc$\downarrow$ & E.Comp$\downarrow$ \\
        \hline
        \multirow{10}{*}{\rotatebox[origin=c]{90}{\textbf{Metric}}}
        & \davft{} & \textcolor{red}{\textbf{0.93}} & \textcolor{red}{\textbf{0.99}} & \textcolor{red}{\textbf{1.00}} & \textcolor{red}{\textbf{0.07}} & \textcolor{red}{\textbf{0.04}} & \textcolor{red}{\textbf{0.75}} & \textcolor{red}{\textbf{0.30}} & \textcolor{red}{\textbf{8.59}} & \textcolor{red}{\textbf{0.34}} & \textcolor{red}{\textbf{2.92}} \\
        \cline{2-12}
        & \dav{} \cite{yang_depth_2024} & 0.30 & 0.61 & 0.71 & 0.46 & 0.84 & 2.24 & 1.52 & 350.76 & 7.34 & 104.45 \\
        & \davthree{} \cite{depthanything3} & 0.52 & 0.75 & 0.84 & 0.29 & 0.40 & 1.63 & 0.99 & 123.58 & 1.98 & 44.95 \\
        & \moge{} \cite{wang2025moge2} & 0.52 & 0.74 & 0.83 & 0.30 & 0.38 & 1.53 & 0.97 & 139.21 & 2.51 & 39.42 \\
        & \depthpro{} \cite{bochkovskii_depth_2024} & 0.32 & 0.63 & 0.71 & 0.47 & 0.85 & 2.25 & 1.50 & 382.05 & 7.78 & 99.93 \\
        & \metricd{} \cite{hu_metric3dv2_2024} & 0.60 & 0.78 & 0.85 & 0.26 & 0.30 & 1.34 & 0.84 & 111.11 & 1.87 & 56.49 \\
        & \unidepth{} \cite{piccinelli_unidepthv2_2025} & 0.62 & 0.81 & 0.88 & 0.27 & 0.40 & 1.40 & 0.86 & 76.29 & 1.23 & 26.64 \\
        & \mapanything{} \cite{keetha2025mapanything} & 0.73 & 0.89 & 0.94 & 0.17 & 0.19 & 1.14 & 0.61 & 62.38 & 1.37 & 15.17 \\
        & \vda{} \cite{video_depth_anything} & 0.77 & 0.93 & 0.97 & 0.17 & 0.17 & \textcolor{blue}{1.11} & 0.59 & 28.88 & 0.73 & 5.82 \\
        & \metricanything{} \cite{metricanything2026} & 0.50 & 0.73 & 0.82 & 0.31 & 0.40 & 1.54 & 0.99 & 139.69 & 2.28 & 53.12 \\
        \hline
        \multirow{5}{*}{\rotatebox[origin=c]{90}{\textbf{Relative}}}
        & \daac{} \cite{sun2025depth} & 0.69 & 0.86 & 0.92 & 0.27 & 0.61 & 1.54 & 0.87 & 22.25 & \textcolor{blue}{0.39} & 4.22 \\
        & \depthcrafter{} \cite{hu_depthcrafter_2024} & 0.40 & 0.68 & 0.77 & 0.48 & 1.08 & 2.41 & 1.55 & 458.01 & 7.77 & 145.34 \\
        & \lotus{} \cite{he_lotus_2025} & 0.35 & 0.69 & 0.84 & 0.46 & 0.92 & 2.25 & 1.43 & 149.78 & 2.11 & 45.46 \\
        & \midas{} \cite{birkl2023midas} & \textcolor{blue}{0.81} & \textcolor{blue}{0.96} & \textcolor{blue}{0.98} & \textcolor{blue}{0.13} & \textcolor{blue}{0.12} & 1.19 & \textcolor{blue}{0.53} & \textcolor{blue}{17.02} & 0.51 & \textcolor{blue}{4.15} \\
        & \marigold{} \cite{ke_marigold_2024} & 0.62 & 0.81 & 0.87 & 0.24 & 0.29 & 1.48 & 0.84 & 73.58 & 1.31 & 19.35 \\
        \hline
    \end{tabular}%
    }
\end{table*}

\clearpage
\noindent\textbf{Rock Regions} (\cref{tab:exp4_semantic_rock}). Rocks introduce significantly greater difficulty. \davft{} remains the clear leader ($\delta_1{=}0.92$, A.Rel${=}0.09$, RMSE${=}3.56$), but the gap to the second-best method (\mapanything{}: $\delta_1{=}0.85$, RMSE${=}4.72$) narrows compared to regolith. The elevated RMSE values across all methods, roughly $4{\times}$ larger than on regolith, indicate that the irregular geometry, sharp edges, and self-occlusion patterns of lunar boulders pose a tangible challenge even for finetuned models. Several zero-shot metric models (\metricd{}, \moge{}) cluster around $\delta_1{\approx}0.74$--$0.78$, suggesting that generic terrestrial depth priors partially capture boulder-like geometry from their Earth-based training sets but fall short of the precision required for reliable obstacle avoidance during rover traverse planning.

\begin{table*}[t]
    \centering
    \caption{Exp. 2: Results on \textbf{Rocks} regions on \lusnar dataset. \textcolor{red}{\textbf{Red}} indicates the best result and \textcolor{blue}{blue} indicates the second best.}
    \label{tab:exp4_semantic_rock}
    \setlength{\tabcolsep}{1.5pt}
    \renewcommand{\arraystretch}{1.05}
    \resizebox{\textwidth}{!}{%
    \begin{tabular}{c|l| c c c c c c c c c c}
        \hline
        \textbf{Type} & \textbf{Method} & $\delta_{1}$$\uparrow$ & $\delta_{2}$$\uparrow$ & $\delta_{3}$$\uparrow$ & A.Rel$\downarrow$ & Sq.Rel$\downarrow$ & RMSE$\downarrow$ & MAE$\downarrow$ & SILog$\downarrow$ & E.Acc$\downarrow$ & E.Comp$\downarrow$ \\
        \hline
        \multirow{10}{*}{\rotatebox[origin=c]{90}{\textbf{Metric}}}
        & \davft{} & \textcolor{red}{\textbf{0.92}} & \textcolor{red}{\textbf{0.99}} & \textcolor{red}{\textbf{1.00}} & \textcolor{red}{\textbf{0.09}} & \textcolor{red}{\textbf{0.50}} & \textcolor{red}{\textbf{3.56}} & \textcolor{red}{\textbf{2.56}} & \textcolor{red}{\textbf{10.36}} & 2.87 & 12.83 \\
        \cline{2-12}
        & \dav{} \cite{yang_depth_2024} & 0.44 & 0.83 & 0.97 & 0.28 & 2.21 & 7.15 & 6.13 & 18.04 & 3.38 & 11.37 \\
        & \davthree{} \cite{depthanything3} & 0.65 & 0.95 & \textcolor{blue}{0.99} & 0.19 & 1.28 & 5.76 & 4.60 & 15.00 & 3.33 & 17.25 \\
        & \moge{} \cite{wang2025moge2} & 0.69 & 0.96 & \textcolor{red}{\textbf{1.00}} & 0.18 & 1.14 & 5.36 & 4.30 & 13.06 & 2.79 & \textcolor{blue}{7.57} \\
        & \depthpro{} \cite{bochkovskii_depth_2024} & 0.45 & 0.83 & 0.96 & 0.28 & 2.46 & 7.49 & 6.41 & 18.03 & \textcolor{blue}{2.66} & \textcolor{red}{\textbf{6.00}} \\
        & \metricd{} \cite{hu_metric3dv2_2024} & 0.80 & 0.97 & \textcolor{blue}{0.99} & 0.15 & 1.00 & 4.76 & 3.70 & \textcolor{blue}{12.87} & 3.16 & 9.12 \\
        & \unidepth{} \cite{piccinelli_unidepthv2_2025} & 0.76 & 0.93 & 0.97 & 0.18 & 2.39 & 6.87 & 4.49 & 13.88 & 3.11 & 10.39 \\
        & \mapanything{} \cite{keetha2025mapanything} & \textcolor{blue}{0.85} & \textcolor{blue}{0.98} & \textcolor{blue}{0.99} & \textcolor{blue}{0.14} & \textcolor{blue}{0.93} & \textcolor{blue}{4.72} & \textcolor{blue}{3.36} & 25.66 & 3.86 & 24.31 \\
        & \vda{} \cite{video_depth_anything} & 0.80 & 0.97 & \textcolor{blue}{0.99} & 0.16 & 1.89 & 6.58 & 4.30 & 15.07 & 3.78 & 18.22 \\
        & \metricanything{} \cite{metricanything2026} & 0.67 & 0.96 & \textcolor{red}{\textbf{1.00}} & 0.18 & 1.14 & 5.33 & 4.33 & 13.30 & 2.82 & 8.12 \\
        \hline
        \multirow{5}{*}{\rotatebox[origin=c]{90}{\textbf{Relative}}}
        & \daac{} \cite{sun2025depth} & 0.66 & 0.84 & 0.92 & 0.24 & 4.72 & 9.51 & 6.56 & 16.05 & 3.63 & 19.38 \\
        & \depthcrafter{} \cite{hu_depthcrafter_2024} & 0.43 & 0.76 & 0.92 & 0.27 & 3.10 & 8.89 & 7.63 & 18.51 & \textcolor{red}{\textbf{2.63}} & 9.96 \\
        & \lotus{} \cite{he_lotus_2025} & 0.51 & 0.79 & 0.91 & 0.28 & 5.25 & 10.30 & 7.55 & 21.30 & 3.45 & 14.99 \\
        & \midas{} \cite{birkl2023midas} & 0.76 & 0.94 & 0.98 & 0.19 & 2.19 & 6.23 & 4.78 & 14.25 & 3.58 & 18.13 \\
        & \marigold{} \cite{ke_marigold_2024} & 0.66 & 0.94 & \textcolor{blue}{0.99} & 0.21 & 1.76 & 6.06 & 4.83 & 15.19 & 3.21 & 10.19 \\
        \hline
    \end{tabular}%
    }
\end{table*}

\clearpage
\noindent\textbf{Craters} (\cref{tab:exp4_semantic_crater}). Craters constitute the hardest semantic category and expose the most severe performance cliff in our benchmark. While \davft{} achieves $\delta_1{=}0.87$, A.Rel${=}0.10$, and RMSE${=}4.52$, showing that targeted finetuning can meaningfully capture crater morphology, the second-best method, \unidepth{}, reaches only $\delta_1{=}0.45$ with an RMSE of $10.85$, representing a ${\sim}2.4{\times}$ error increase. This dramatic gap shows that crater geometry, with its sharp depth discontinuities at the rim, gradually varying concave interior slopes, and cast shadows, is extremely difficult for models without explicit lunar supervision. Craters represent the most hazardous terrain feature for rover navigation, and the near-total failure of zero-shot architectures to resolve their interior geometry underscores the necessity of domain-specific adaptation or physics-informed depth priors for safe autonomous traverse planning near crater boundaries.

\begin{table*}[t]
    \centering
    \caption{Exp. 2: Results on \textbf{Craters} regions on \lusnar dataset. \textcolor{red}{\textbf{Red}} indicates the best result and \textcolor{blue}{blue} indicates the second best.}
    \label{tab:exp4_semantic_crater}
    \setlength{\tabcolsep}{1.5pt}
    \renewcommand{\arraystretch}{1.05}
    \resizebox{\textwidth}{!}{%
    \begin{tabular}{c|l| c c c c c c c c c c}
        \hline
        \textbf{Type} & \textbf{Method} & $\delta_{1}$$\uparrow$ & $\delta_{2}$$\uparrow$ & $\delta_{3}$$\uparrow$ & A.Rel$\downarrow$ & Sq.Rel$\downarrow$ & RMSE$\downarrow$ & MAE$\downarrow$ & SILog$\downarrow$ & E.Acc$\downarrow$ & E.Comp$\downarrow$ \\
        \hline
        \multirow{10}{*}{\rotatebox[origin=c]{90}{\textbf{Metric}}}
        & \davft{} & \textcolor{red}{\textbf{0.87}} & \textcolor{red}{\textbf{0.98}} & \textcolor{red}{\textbf{1.00}} & \textcolor{red}{\textbf{0.10}} & \textcolor{red}{\textbf{0.77}} & \textcolor{red}{\textbf{4.52}} & \textcolor{red}{\textbf{3.83}} & \textcolor{red}{\textbf{8.48}} & \textcolor{red}{\textbf{2.05}} & 10.06 \\
        \cline{2-12}
        & \dav{} \cite{yang_depth_2024} & 0.23 & 0.45 & 0.73 & 0.37 & 6.69 & 15.30 & 14.44 & 15.69 & 2.67 & 6.68 \\
        & \davthree{} \cite{depthanything3} & 0.37 & 0.75 & 0.90 & 0.27 & 4.16 & 11.53 & 10.82 & 12.38 & 2.25 & \textcolor{blue}{6.50} \\
        & \moge{} \cite{wang2025moge2} & 0.40 & 0.71 & 0.88 & 0.27 & 4.45 & 11.79 & 11.05 & 12.21 & \textcolor{blue}{2.14} & 11.12 \\
        & \depthpro{} \cite{bochkovskii_depth_2024} & 0.23 & 0.44 & 0.72 & 0.37 & 7.14 & 15.72 & 14.82 & 15.87 & 2.29 & \textcolor{red}{\textbf{5.56}} \\
        & \metricd{} \cite{hu_metric3dv2_2024} & 0.41 & 0.75 & 0.90 & 0.27 & 3.87 & 11.11 & 10.31 & 19.84 & 2.46 & 9.85 \\
        & \unidepth{} \cite{piccinelli_unidepthv2_2025} & \textcolor{blue}{0.45} & 0.73 & \textcolor{blue}{0.91} & \textcolor{blue}{0.25} & \textcolor{blue}{3.73} & \textcolor{blue}{10.85} & \textcolor{blue}{10.00} & 12.37 & 2.28 & 13.04 \\
        & \mapanything{} \cite{keetha2025mapanything} & 0.42 & \textcolor{blue}{0.76} & 0.90 & 0.26 & 3.76 & 11.07 & 10.24 & 13.35 & 2.78 & 11.57 \\
        & \vda{} \cite{video_depth_anything} & 0.36 & 0.66 & 0.87 & 0.29 & 4.66 & 12.46 & 11.47 & 14.32 & 2.81 & 8.78 \\
        & \metricanything{} \cite{metricanything2026} & 0.41 & 0.73 & 0.88 & 0.27 & 4.42 & 11.60 & 10.93 & \textcolor{blue}{11.99} & 2.19 & 10.07 \\
        \hline
        \multirow{5}{*}{\rotatebox[origin=c]{90}{\textbf{Relative}}}
        & \daac{} \cite{sun2025depth} & 0.31 & 0.60 & 0.86 & 0.31 & 5.00 & 13.29 & 12.11 & 16.88 & 2.97 & 10.98 \\
        & \depthcrafter{} \cite{hu_depthcrafter_2024} & 0.17 & 0.38 & 0.65 & 0.41 & 8.25 & 17.29 & 16.23 & 19.17 & 3.37 & 16.15 \\
        & \lotus{} \cite{he_lotus_2025} & 0.21 & 0.47 & 0.68 & 0.39 & 7.22 & 15.50 & 14.38 & 19.93 & 3.66 & 13.84 \\
        & \midas{} \cite{birkl2023midas} & 0.34 & 0.70 & \textcolor{blue}{0.91} & 0.28 & 4.15 & 12.10 & 10.95 & 15.84 & 3.05 & 17.69 \\
        & \marigold{} \cite{ke_marigold_2024} & 0.26 & 0.58 & 0.80 & 0.33 & 5.70 & 14.11 & 12.98 & 17.45 & 3.24 & 7.93 \\
        \hline
    \end{tabular}%
    }
\end{table*}


\clearpage

\subsection{Distance-Stratified Evaluation}
\label{sec:supp_distance}

\Cref{tab:exp3_distance_near,tab:exp3_distance_mid,tab:exp3_distance_far} extend the main text's distance analysis (Exp.~3) by providing the full metric breakdown for each depth interval on \lusnar. The main paper established that metric and structurally constrained models preserve internal linearity; here, we present the detailed per-method results across three ranges, near (0--5\,m), medium (5--15\,m), and far (15--50\,m), to expose how depth estimation quality degrades as a function of range. As discussed in Exp.~3, we apply a global least-squares affine alignment anchored by the dominant mid-range terrain, which allows us to interpret deviations in the near and far ranges as direct indicators of distance-dependent scale drift.\\

\noindent\textbf{Near Range (0--5\,m)} (\cref{tab:exp3_distance_near}). \davft{} achieves outstanding accuracy ($\delta_1{=}0.92$, A.Rel${=}0.07$, RMSE${=}0.16$). \midas{} is the best relative method ($\delta_1{=}0.79$, RMSE${=}0.33$). While the low absolute RMSE values across all methods reflect the inherently small error magnitudes at short distances, relative errors (A.Rel) reveal meaningful performance differences. Several metric methods (\dav{}, \depthpro{}) exhibit A.Rel values exceeding $0.49$, significantly worse than their mid-range performance. This near-field degradation, also observed in the main text for \change and \cheri, suggests that the global affine alignment, fitted primarily to the mid-range, cannot correct for non-linear distortions at close proximity, where scale drift is most pronounced for models that lack internal linearity.

\begin{table*}[t]
    \centering
    \caption{Exp. 3: Results on \textbf{Near (0--5m)} distances on \lusnar dataset. \textcolor{red}{\textbf{Red}} indicates the best result and \textcolor{blue}{blue} indicates the second best.}
    \label{tab:exp3_distance_near}
    \setlength{\tabcolsep}{1.5pt}
    \renewcommand{\arraystretch}{1.05}
    \resizebox{\textwidth}{!}{%
    \begin{tabular}{c|l| c c c c c c c c c c}
        \hline
        \textbf{Type} & \textbf{Method} & $\delta_{1}$$\uparrow$ & $\delta_{2}$$\uparrow$ & $\delta_{3}$$\uparrow$ & A.Rel$\downarrow$ & Sq.Rel$\downarrow$ & RMSE$\downarrow$ & MAE$\downarrow$ & SILog$\downarrow$ & E.Acc$\downarrow$ & E.Comp$\downarrow$ \\
        \hline
        \multirow{10}{*}{\rotatebox[origin=c]{90}{\textbf{Metric}}}
        & \davft{} & \textcolor{red}{\textbf{0.92}} & \textcolor{red}{\textbf{0.98}} & \textcolor{red}{\textbf{1.00}} & \textcolor{red}{\textbf{0.07}} & \textcolor{red}{\textbf{0.02}} & \textcolor{red}{\textbf{0.16}} & \textcolor{red}{\textbf{0.14}} & \textcolor{red}{\textbf{6.96}} & \textcolor{red}{\textbf{0.41}} & \textcolor{red}{\textbf{7.46}} \\
        \cline{2-12}
        & \dav{} \cite{yang_depth_2024} & 0.30 & 0.55 & 0.64 & 0.49 & 0.68 & 1.13 & 0.99 & 379.83 & 7.18 & 158.41 \\
        & \davthree{} \cite{depthanything3} & 0.48 & 0.69 & 0.80 & 0.31 & 0.31 & 0.70 & 0.61 & 130.89 & 2.85 & 84.58 \\
        & \moge{} \cite{wang2025moge2} & 0.48 & 0.68 & 0.78 & 0.32 & 0.31 & 0.70 & 0.62 & 148.40 & 3.24 & 102.99 \\
        & \depthpro{} \cite{bochkovskii_depth_2024} & 0.33 & 0.55 & 0.63 & 0.50 & 0.73 & 1.15 & 0.99 & 413.67 & 7.55 & 165.65 \\
        & \metricd{} \cite{hu_metric3dv2_2024} & 0.55 & 0.73 & 0.81 & 0.28 & 0.26 & 0.64 & 0.56 & 119.17 & 3.30 & 91.01 \\
        & \unidepth{} \cite{piccinelli_unidepthv2_2025} & 0.57 & 0.76 & 0.85 & 0.30 & 0.38 & 0.66 & 0.59 & 79.06 & 2.07 & 57.54 \\
        & \mapanything{} \cite{keetha2025mapanything} & 0.71 & 0.87 & 0.92 & 0.18 & 0.14 & 0.41 & 0.35 & 64.98 & 1.62 & 33.76 \\
        & \vda{} \cite{video_depth_anything} & 0.74 & 0.91 & 0.96 & 0.19 & 0.14 & 0.43 & 0.37 & 28.28 & 0.95 & 15.13 \\
        & \metricanything{} \cite{metricanything2026} & 0.47 & 0.66 & 0.77 & 0.33 & 0.32 & 0.72 & 0.63 & 148.68 & 3.69 & 110.19 \\
        \hline
        \multirow{5}{*}{\rotatebox[origin=c]{90}{\textbf{Relative}}}
        & \daac{} \cite{sun2025depth} & 0.66 & 0.83 & 0.91 & 0.31 & 0.61 & 0.66 & 0.61 & 17.25 & \textcolor{blue}{0.61} & \textcolor{blue}{14.86} \\
        & \depthcrafter{} \cite{hu_depthcrafter_2024} & 0.42 & 0.64 & 0.71 & 0.52 & 0.96 & 1.25 & 1.00 & 500.38 & 7.46 & 163.07 \\
        & \lotus{} \cite{he_lotus_2025} & 0.27 & 0.64 & 0.81 & 0.53 & 0.94 & 1.28 & 1.14 & 154.55 & 3.93 & 73.23 \\
        & \midas{} \cite{birkl2023midas} & \textcolor{blue}{0.79} & \textcolor{blue}{0.95} & \textcolor{blue}{0.98} & \textcolor{blue}{0.14} & \textcolor{blue}{0.07} & \textcolor{blue}{0.33} & \textcolor{blue}{0.28} & \textcolor{blue}{15.01} & 0.69 & 21.74 \\
        & \marigold{} \cite{ke_marigold_2024} & 0.59 & 0.76 & 0.83 & 0.25 & 0.21 & 0.60 & 0.50 & 77.24 & 2.45 & 46.42 \\
        \hline
    \end{tabular}%
    }
\end{table*}

\clearpage
\noindent\textbf{Medium Range (5--15\,m)} (\cref{tab:exp3_distance_mid}). As the implicit anchor range for our global affine alignment, the medium interval yields the highest overall accuracy. \davft{} reaches $\delta_1{=}0.99$ and A.Rel${=}0.07$ with RMSE${=}0.88$, approaching near-perfect performance. \vda{} and \midas{} share the second-best $\delta_1$ of $0.89$, with \midas{} achieving a slightly lower RMSE ($1.27$ vs.\ $1.33$). The strong mid-range performance across a broad set of methods is consistent with the main text's observation that models with adequate internal linearity recover well within the range that dominates the alignment fit. This serves as a useful baseline: deviations from this accuracy in the near and far ranges directly quantify the severity of each model's scale drift.

\begin{table*}[t]
    \centering
    \caption{Exp. 3: Results on \textbf{Medium (5--15m)} distances on \lusnar dataset. \textcolor{red}{\textbf{Red}} indicates the best result and \textcolor{blue}{blue} indicates the second best.}
    \label{tab:exp3_distance_mid}
    \setlength{\tabcolsep}{1.5pt}
    \renewcommand{\arraystretch}{1.05}
    \resizebox{\textwidth}{!}{%
    \begin{tabular}{c|l| c c c c c c c c c c}
        \hline
        \textbf{Type} & \textbf{Method} & $\delta_{1}$$\uparrow$ & $\delta_{2}$$\uparrow$ & $\delta_{3}$$\uparrow$ & A.Rel$\downarrow$ & Sq.Rel$\downarrow$ & RMSE$\downarrow$ & MAE$\downarrow$ & SILog$\downarrow$ & E.Acc$\downarrow$ & E.Comp$\downarrow$ \\
        \hline
        \multirow{10}{*}{\rotatebox[origin=c]{90}{\textbf{Metric}}}
        & \davft{} & \textcolor{red}{\textbf{0.99}} & \textcolor{red}{\textbf{1.00}} & \textcolor{red}{\textbf{1.00}} & \textcolor{red}{\textbf{0.07}} & \textcolor{red}{\textbf{0.10}} & \textcolor{red}{\textbf{0.88}} & \textcolor{red}{\textbf{0.60}} & \textcolor{red}{\textbf{4.47}} & 1.17 & 36.08 \\
        \cline{2-12}
        & \dav{} \cite{yang_depth_2024} & 0.20 & 0.83 & \textcolor{blue}{0.99} & 0.40 & 1.49 & 3.29 & 3.13 & 7.61 & 2.03 & \textcolor{blue}{9.27} \\
        & \davthree{} \cite{depthanything3} & 0.58 & 0.96 & \textcolor{red}{\textbf{1.00}} & 0.25 & 0.68 & 2.23 & 2.00 & 6.96 & 1.21 & 33.29 \\
        & \moge{} \cite{wang2025moge2} & 0.59 & 0.96 & \textcolor{red}{\textbf{1.00}} & 0.25 & 0.64 & 2.12 & 1.97 & 5.86 & \textcolor{red}{\textbf{0.92}} & 18.43 \\
        & \depthpro{} \cite{bochkovskii_depth_2024} & 0.19 & 0.87 & \textcolor{blue}{0.99} & 0.39 & 1.44 & 3.21 & 3.05 & 7.35 & 1.31 & 26.11 \\
        & \metricd{} \cite{hu_metric3dv2_2024} & 0.74 & 0.98 & \textcolor{red}{\textbf{1.00}} & 0.19 & 0.45 & 1.77 & 1.53 & 8.70 & 1.34 & 13.91 \\
        & \unidepth{} \cite{piccinelli_unidepthv2_2025} & 0.78 & 0.98 & \textcolor{red}{\textbf{1.00}} & 0.17 & 0.39 & 1.61 & 1.39 & 6.10 & 1.16 & 28.51 \\
        & \mapanything{} \cite{keetha2025mapanything} & 0.81 & 0.97 & \textcolor{red}{\textbf{1.00}} & 0.16 & 0.41 & 1.55 & 1.28 & 7.19 & 1.40 & 50.80 \\
        & \vda{} \cite{video_depth_anything} & \textcolor{blue}{0.89} & \textcolor{blue}{0.99} & \textcolor{red}{\textbf{1.00}} & \textcolor{blue}{0.11} & 0.28 & 1.33 & \textcolor{blue}{0.94} & 7.32 & 1.53 & 29.17 \\
        & \metricanything{} \cite{metricanything2026} & 0.56 & 0.96 & \textcolor{red}{\textbf{1.00}} & 0.26 & 0.68 & 2.22 & 2.06 & \textcolor{blue}{5.79} & \textcolor{blue}{0.99} & 21.98 \\
        \hline
        \multirow{5}{*}{\rotatebox[origin=c]{90}{\textbf{Relative}}}
        & \daac{} \cite{sun2025depth} & 0.83 & 0.97 & \textcolor{blue}{0.99} & 0.13 & 0.37 & 1.53 & 1.13 & 7.87 & 1.53 & 30.53 \\
        & \depthcrafter{} \cite{hu_depthcrafter_2024} & 0.28 & 0.81 & \textcolor{blue}{0.99} & 0.38 & 1.45 & 3.28 & 3.02 & 10.55 & 1.79 & 32.34 \\
        & \lotus{} \cite{he_lotus_2025} & 0.64 & 0.92 & \textcolor{blue}{0.99} & 0.19 & 0.49 & 1.99 & 1.58 & 14.25 & 2.22 & 10.18 \\
        & \midas{} \cite{birkl2023midas} & \textcolor{blue}{0.89} & \textcolor{red}{\textbf{1.00}} & \textcolor{red}{\textbf{1.00}} & 0.12 & \textcolor{blue}{0.22} & \textcolor{blue}{1.27} & 0.96 & 6.56 & 1.60 & 34.96 \\
        & \marigold{} \cite{ke_marigold_2024} & 0.66 & 0.97 & \textcolor{red}{\textbf{1.00}} & 0.21 & 0.53 & 2.01 & 1.71 & 8.49 & 2.11 & \textcolor{red}{\textbf{8.47}} \\
        \hline
    \end{tabular}%
    }
\end{table*}

\clearpage
\noindent\textbf{Far Range (15--50\,m)} (\cref{tab:exp3_distance_far}). \davft{} maintains its lead ($\delta_1{=}0.93$, A.Rel${=}0.09$, RMSE${=}3.70$), while \mapanything{} emerges as the second-best metric method ($\delta_1{=}0.85$, RMSE${=}5.02$). The absolute RMSE values increase substantially across all methods compared to nearer ranges, as expected due to the larger physical depth magnitudes. Among relative methods, \midas{} performs competitively ($\delta_1{=}0.79$, RMSE${=}5.77$), yet the gap to the top metric methods widens compared to the medium range. This widening gap confirms that metric supervision provides a stronger inductive bias for recovering absolute scale at longer distances, where monocular depth cues grow increasingly ambiguous and the textureless lunar regolith offers minimal visual guidance. For autonomous rover navigation, reliable far-field depth estimation is critical for long-horizon path planning and hazard anticipation, and the present results show that it remains achievable primarily through domain-adapted or metrically supervised architectures.

\begin{table*}[t]
    \centering
    \caption{Exp. 3: Results on \textbf{Far (15--50m)} distances on \lusnar dataset. \textcolor{red}{\textbf{Red}} indicates the best result and \textcolor{blue}{blue} indicates the second best.}
    \label{tab:exp3_distance_far}
    \setlength{\tabcolsep}{1.5pt}
    \renewcommand{\arraystretch}{1.05}
    \resizebox{\textwidth}{!}{%
    \begin{tabular}{c|l| c c c c c c c c c c}
        \hline
        \textbf{Type} & \textbf{Method} & $\delta_{1}$$\uparrow$ & $\delta_{2}$$\uparrow$ & $\delta_{3}$$\uparrow$ & A.Rel$\downarrow$ & Sq.Rel$\downarrow$ & RMSE$\downarrow$ & MAE$\downarrow$ & SILog$\downarrow$ & E.Acc$\downarrow$ & E.Comp$\downarrow$ \\
        \hline
        \multirow{10}{*}{\rotatebox[origin=c]{90}{\textbf{Metric}}}
        & \davft{} & \textcolor{red}{\textbf{0.93}} & \textcolor{red}{\textbf{0.99}} & \textcolor{red}{\textbf{0.99}} & \textcolor{red}{\textbf{0.09}} & \textcolor{red}{\textbf{0.49}} & \textcolor{red}{\textbf{3.70}} & \textcolor{red}{\textbf{2.60}} & \textcolor{red}{\textbf{11.91}} & 1.52 & 11.10 \\
        \cline{2-12}
        & \dav{} \cite{yang_depth_2024} & 0.60 & 0.85 & 0.95 & 0.20 & 2.28 & 8.78 & 6.50 & 25.11 & 2.77 & 14.19 \\
        & \davthree{} \cite{depthanything3} & 0.75 & 0.93 & \textcolor{blue}{0.98} & 0.16 & 1.31 & 6.27 & 4.59 & 19.20 & 1.96 & 41.20 \\
        & \moge{} \cite{wang2025moge2} & 0.82 & 0.96 & \textcolor{red}{\textbf{0.99}} & 0.13 & 0.97 & 5.57 & 3.91 & 16.50 & \textcolor{red}{\textbf{1.13}} & 20.87 \\
        & \depthpro{} \cite{bochkovskii_depth_2024} & 0.65 & 0.88 & 0.97 & 0.18 & 1.91 & 8.13 & 5.86 & 22.27 & 1.42 & 28.95 \\
        & \metricd{} \cite{hu_metric3dv2_2024} & 0.81 & 0.96 & \textcolor{red}{\textbf{0.99}} & 0.14 & 1.02 & 5.46 & 4.06 & 18.08 & 1.77 & 11.11 \\
        & \unidepth{} \cite{piccinelli_unidepthv2_2025} & 0.76 & 0.92 & 0.96 & 0.15 & 1.82 & 6.85 & 4.53 & \textcolor{blue}{16.10} & 1.63 & 36.02 \\
        & \mapanything{} \cite{keetha2025mapanything} & \textcolor{blue}{0.85} & \textcolor{blue}{0.97} & \textcolor{red}{\textbf{0.99}} & \textcolor{blue}{0.12} & \textcolor{blue}{0.86} & \textcolor{blue}{5.02} & \textcolor{blue}{3.51} & 32.38 & 3.02 & 74.35 \\
        & \vda{} \cite{video_depth_anything} & 0.76 & 0.94 & \textcolor{red}{\textbf{0.99}} & 0.16 & 1.62 & 6.83 & 4.58 & 19.16 & 2.73 & 58.16 \\
        & \metricanything{} \cite{metricanything2026} & 0.80 & 0.96 & \textcolor{red}{\textbf{0.99}} & 0.13 & 1.02 & 5.63 & 4.08 & 16.89 & \textcolor{blue}{1.15} & 20.52 \\
        \hline
        \multirow{5}{*}{\rotatebox[origin=c]{90}{\textbf{Relative}}}
        & \daac{} \cite{sun2025depth} & 0.66 & 0.83 & 0.90 & 0.21 & 3.12 & 8.75 & 6.06 & 19.86 & 2.29 & 27.67 \\
        & \depthcrafter{} \cite{hu_depthcrafter_2024} & 0.57 & 0.84 & 0.95 & 0.21 & 2.31 & 8.90 & 6.62 & 24.53 & 1.80 & 40.18 \\
        & \lotus{} \cite{he_lotus_2025} & 0.49 & 0.77 & 0.89 & 0.25 & 3.25 & 9.71 & 7.23 & 27.52 & 2.52 & \textcolor{red}{\textbf{8.26}} \\
        & \midas{} \cite{birkl2023midas} & 0.79 & 0.94 & \textcolor{blue}{0.98} & 0.14 & 1.19 & 5.77 & 4.10 & 18.24 & 2.15 & 33.28 \\
        & \marigold{} \cite{ke_marigold_2024} & 0.75 & 0.93 & 0.97 & 0.16 & 1.52 & 6.58 & 4.81 & 20.67 & 2.32 & \textcolor{blue}{9.50} \\
        \hline
    \end{tabular}%
    }
\end{table*}

\subsection{Training Regime and Data Analysis}
\label{sec:training_regime}

\subsubsection{Training Scale}

\begin{figure}
    \centering
    \includegraphics[width=0.95\linewidth]{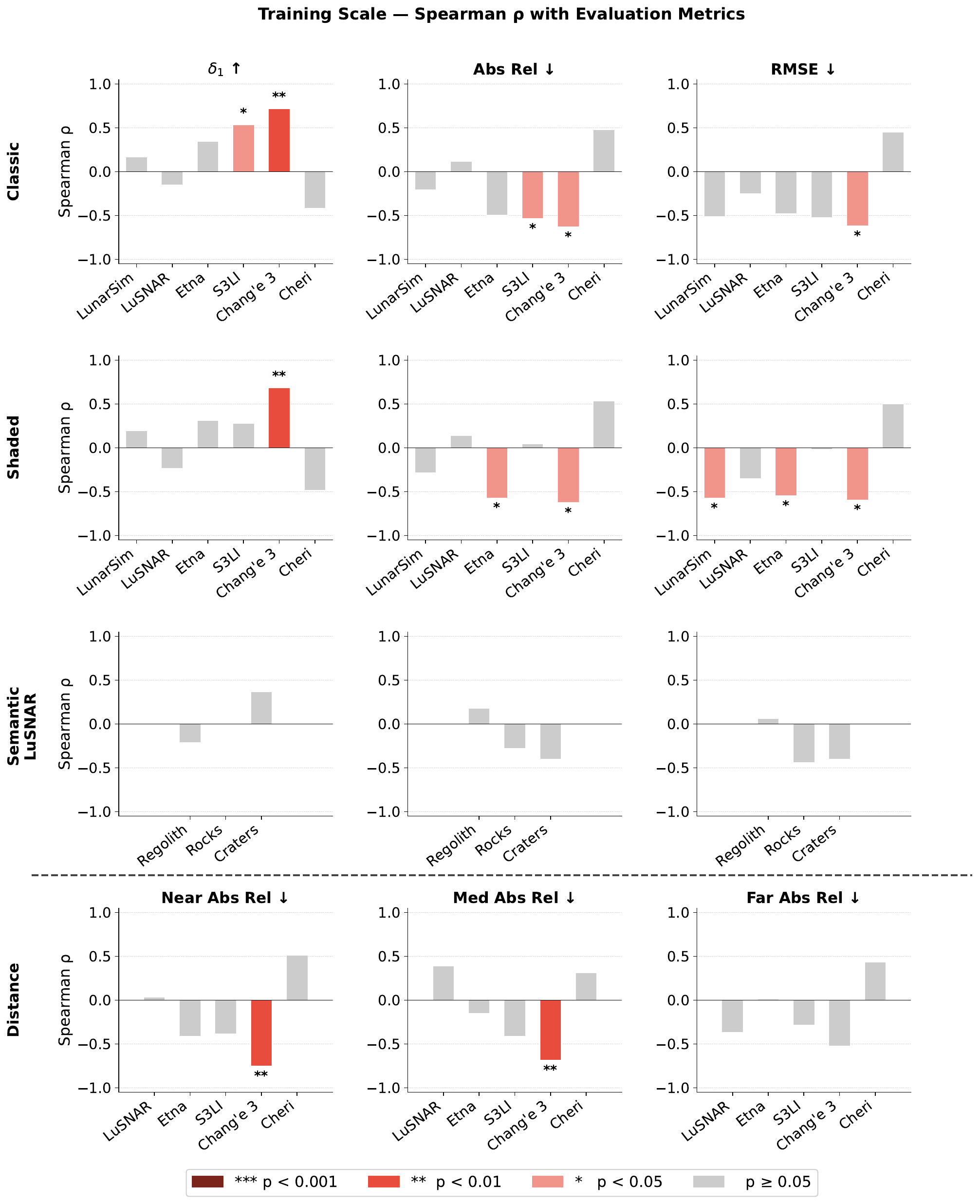}
    \caption{Spearman rank correlation ($\rho$) between model Training Scale and evaluation metrics. The figure illustrates the correlation of the models' total training scale with depth estimation performance ($\delta_1$, Abs Rel, RMSE) across standard dataset evaluations (Classic), extreme lighting conditions (Shaded), specific geological features (Semantic LuSNAR: Regolith, Rocks, Craters), and varying depth intervals (Distance: Near, Medium, Far). Statistical significance is indicated by p-values.}
    \label{fig:scale_superplot}
\end{figure}

As illustrated in \cref{fig:scale_superplot} , we expand upon the main text's observation that massive training scale cannot universally solve the lunar domain gap. By decomposing performance across specific conditions, we isolate exactly where data scaling succeeds and fails.

\textbf{Overall and Shaded Performance:} On the authentic Chang'e-3 dataset, increasing a model's training scale strongly correlates with higher depth accuracy and reduced error. Crucially, this advantage holds steady even within heavily shaded regions. However, this trend does not generalize. On the challenging Cheri dark analog, scaling exhibits a reversed, non-significant trend, and it offers little to no consistent benefit across the synthetic benchmarks.

\textbf{Distance-Dependent Scaling:} When analyzing performance across depth ranges, we find that the benefits of scaling on Chang'e-3 are heavily localized to the near and medium fields. As the depth extends into the far range, the correlation weakens and loses strict statistical significance. This indicates that while massive terrestrial datasets help models extract nearby, high-frequency details, they provide diminishing returns for understanding distant, large-scale lunar topography.

\textbf{Semantic Features:} Finally, our evaluation of specific semantic regions within the LuSNAR dataset confirms that scale offers no statistically significant improvement for distinct geological constructs. The correlations for regolith, rocks, and craters all fall well short of significance. This reinforces a core takeaway: simply feeding a network more data does not inherently teach it to parse complex extraterrestrial geometries like craters and boulders.

\subsubsection{Real Training Data}

\begin{figure}
    \centering
    \includegraphics[width=0.95\linewidth]{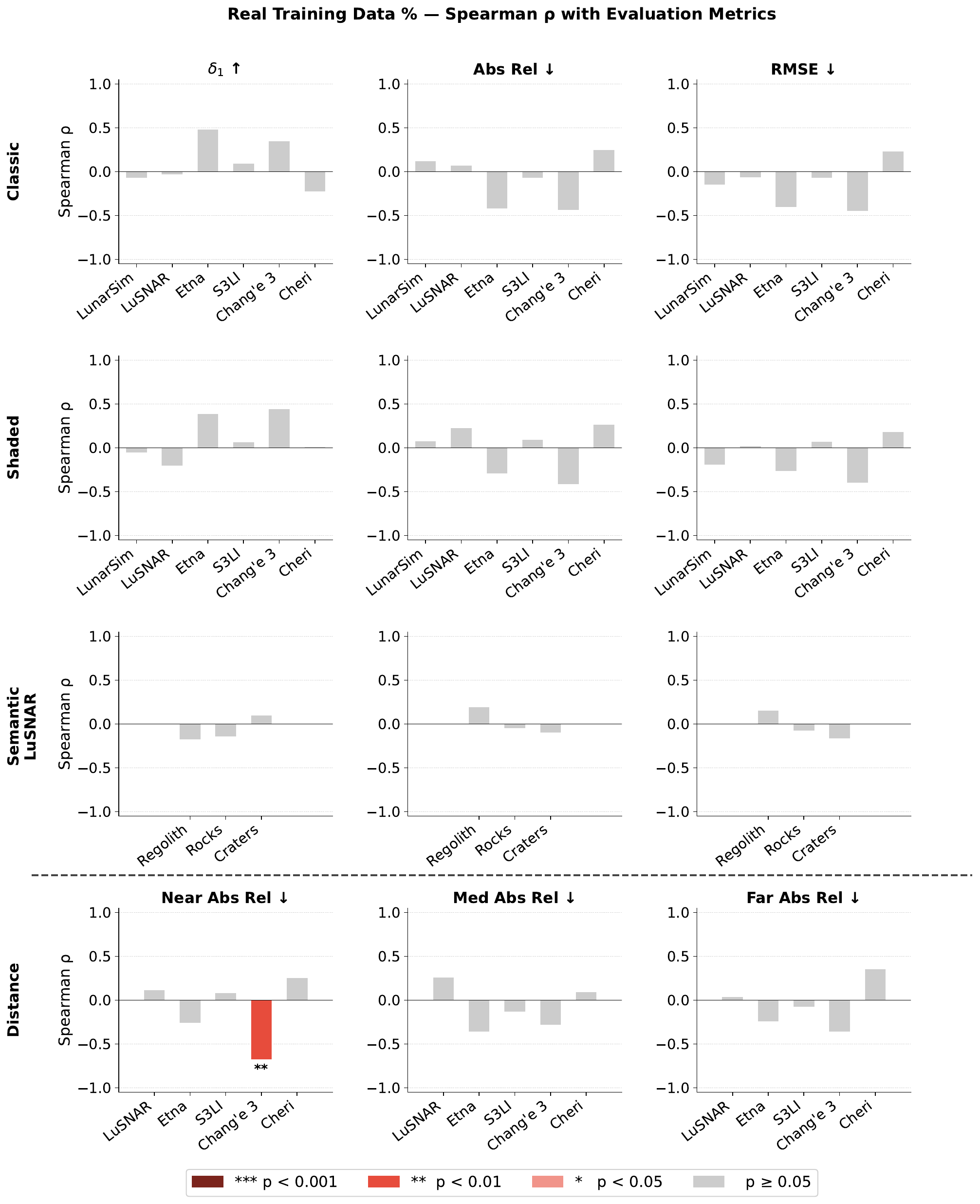}
    \caption{Spearman rank correlation (p) between the percentage of Real Training Data Percentage and evaluation metrics. This figure illustrates the correlation of the proportion of real-world training data with depth estimation performance ($\delta_1$, Abs Rel, RMSE) across standard dataset evaluations (Classic), extreme lighting conditions (Shaded), specific geological features (Semantic LuSNAR: Regolith, Rocks, Craters), and varying depth intervals (Distance: Near, Medium, Far). Statistical significance is indicated by p-values}
    \label{fig:real_superplot}
\end{figure}

As illustrated in \cref{fig:real_superplot}, we evaluate whether increasing the proportion of real-world terrestrial images in the training set improves lunar adaptation. Given the severe visual domain gap, relying on Earth-based priors yields largely negligible improvements.

\textbf{Overall and Semantic Performance:} Increasing the ratio of real training data shows no statistically significant correlation with depth accuracy or error reduction across the classic evaluations, extreme lighting conditions, or specific geological features such as regolith, rocks, and craters. Both the synthetic benchmarks and the authentic lunar analogs yield $p$-values well above the significance threshold. This indicates that simply injecting more terrestrial photography into the training pipeline does not resolve the broader lunar domain gap, nor does it improve the structural understanding of extraterrestrial obstacles.

\textbf{Distance-Dependent Scaling:} The distance-stratified evaluation reveals a highly isolated benefit: increasing the real data ratio correlates with reduced error strictly within the near-field range of the authentic Chang'e-3 imagery ($\rho=-0.677$, $p=0.008$). However, this advantage entirely vanishes in the medium and far ranges, and critically, it does not generalize to any other dataset, including the Cheri analog. This lack of generalizability raises significant concerns regarding the use of terrestrial priors for space applications; while Earth-based textures might occasionally assist in resolving nearby, high-frequency surface details, they completely fail to supply the robust, transferrable geometric understanding required for safe and reliable lunar navigation.

\subsubsection{Metric Supervision}

\begin{figure}
    \centering
    \includegraphics[width=0.9\linewidth]{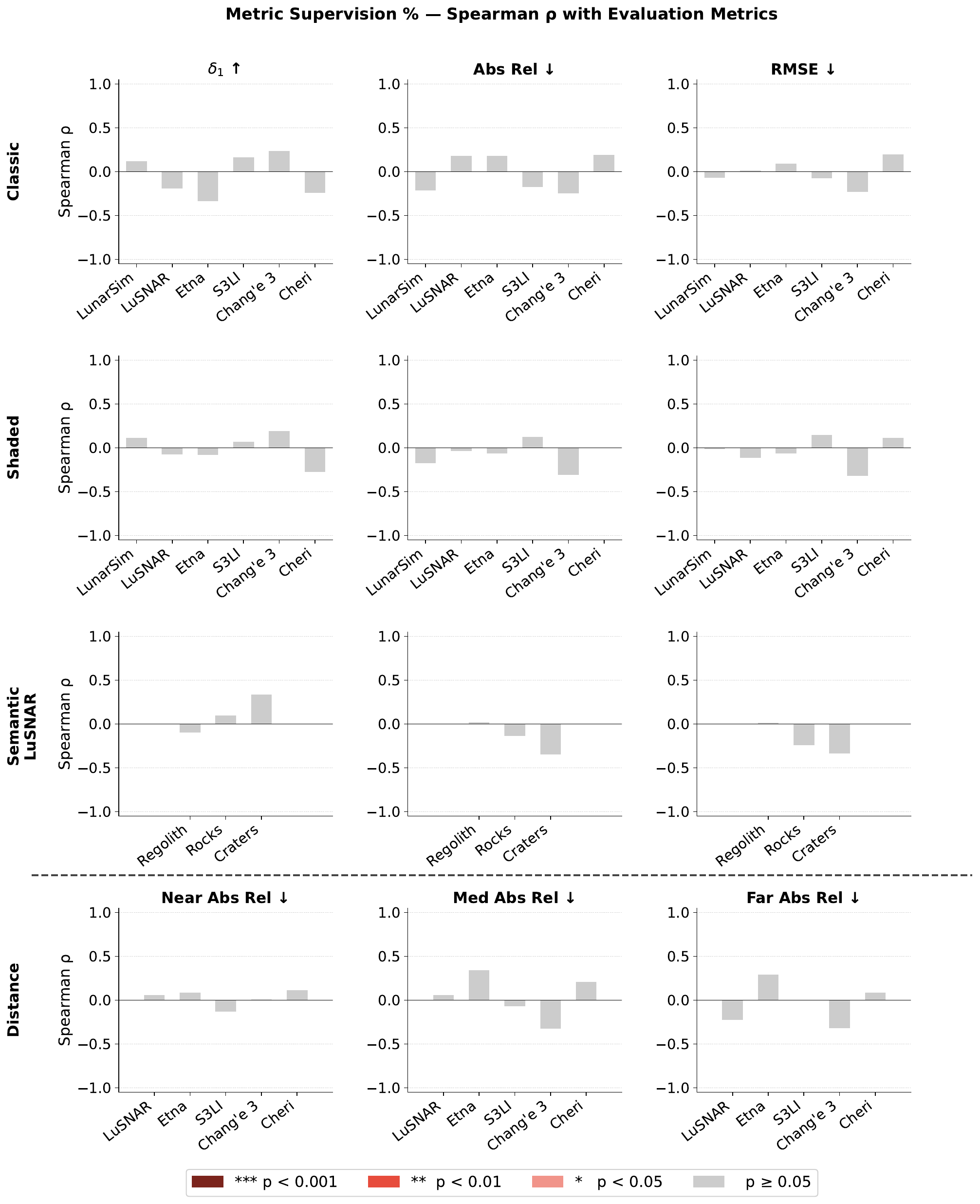}
    \caption{Spearman rank correlation (p) between the percentage of Metric Supervision and evaluation metrics. This plot illustrates the correlation of the proportion of metric supervision (percentage of the training data containing metric depth) used during training with depth estimation performance ($\delta_1$, Abs Rel, RMSE) across standard dataset evaluations (Classic), extreme lighting conditions (Shaded), specific geological features (Semantic LuSNAR: Regolith, Rocks, Craters), and varying depth intervals (Distance: Near, Medium, Far). Statistical significance is indicated by p-values.}
    \label{fig:metric_superplot}
\end{figure}

As illustrated in \cref{fig:metric_superplot}, we conclude our data analysis by evaluating the influence of metric supervision. We investigate whether training with a higher percentage of absolute metric ground truth inherently translates to better zero-shot scale recovery and structural accuracy on the Moon. 

\textbf{Overall and Shaded Performance:} Across both the classic evaluations and the extreme lighting scenarios, we observe a complete absence of statistical significance. For every dataset, including the synthetic environments, Earth analogs, and authentic Chang'e-3 imagery, the correlations between metric supervision and performance metrics ($\delta_1$, Abs Rel, RMSE) yield $p$-values well above the $0.05$ threshold. 

\textbf{Semantic and Distance-Dependent Scaling:} This lack of correlation persists even under granular inspection. Stratifying the evaluation by specific geological constructs (regolith, rocks, craters) or by distance intervals (near, medium, far) fails to reveal any localized advantages. Not a single $p$-value across these breakdowns approaches statistical significance.

\textbf{Primary Takeaway:} These universally unpromising results forcefully demonstrate that the raw volume of metric supervision does not guarantee cross-domain robustness. Models trained with near-total metric supervision perform no better in extraterrestrial environments than those relying on affine-invariant or mixed-supervision strategies. This heavily reinforces our overarching conclusion: resolving the lunar sim-to-real domain gap will require targeted architectural innovations—such as decoupling illumination or integrating physics-based priors—rather than simply scaling up metric datasets.

\section{Additional Visualizations}

\begin{figure}[h]
    \centering
    \includegraphics[width=0.9\linewidth]{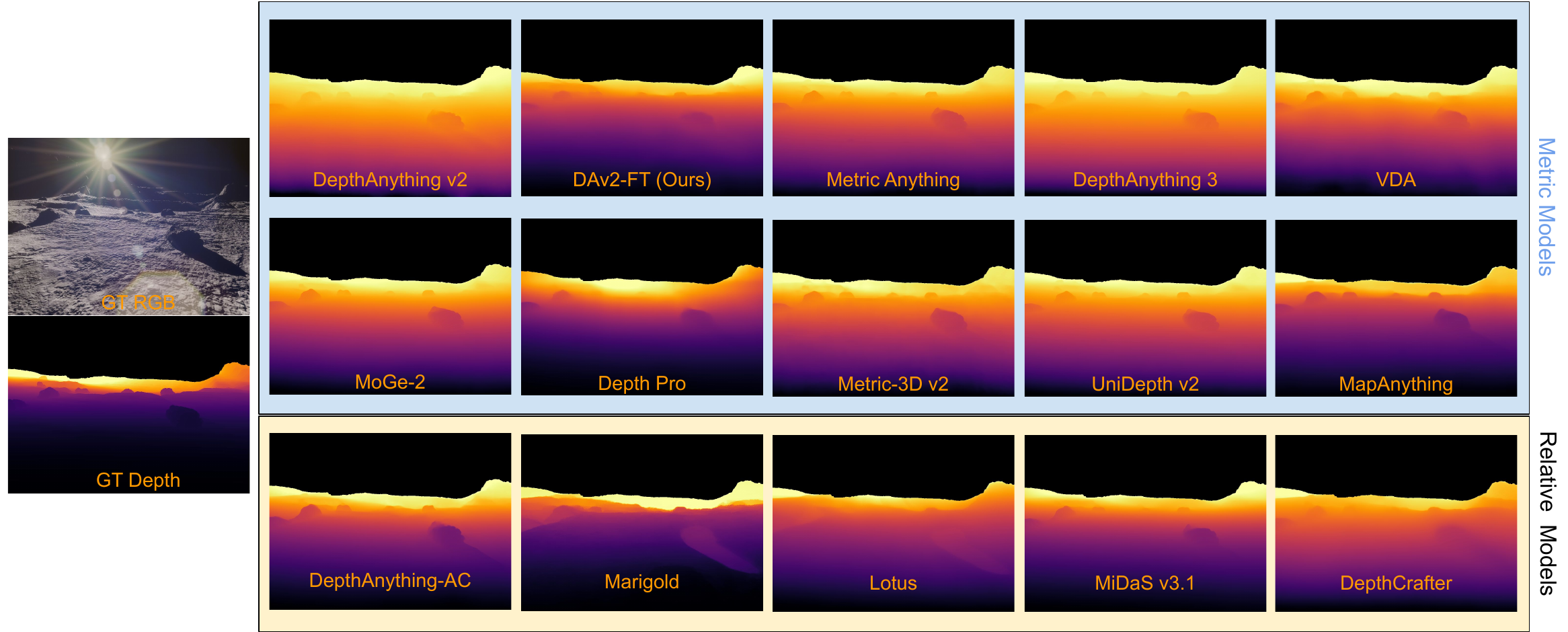}
    \caption{Qualitative depth estimation results on \lunarsim}
    \label{fig:supp_vis_s4}
\end{figure}

\begin{figure}[h]
    \centering
    \includegraphics[width=0.9\linewidth]{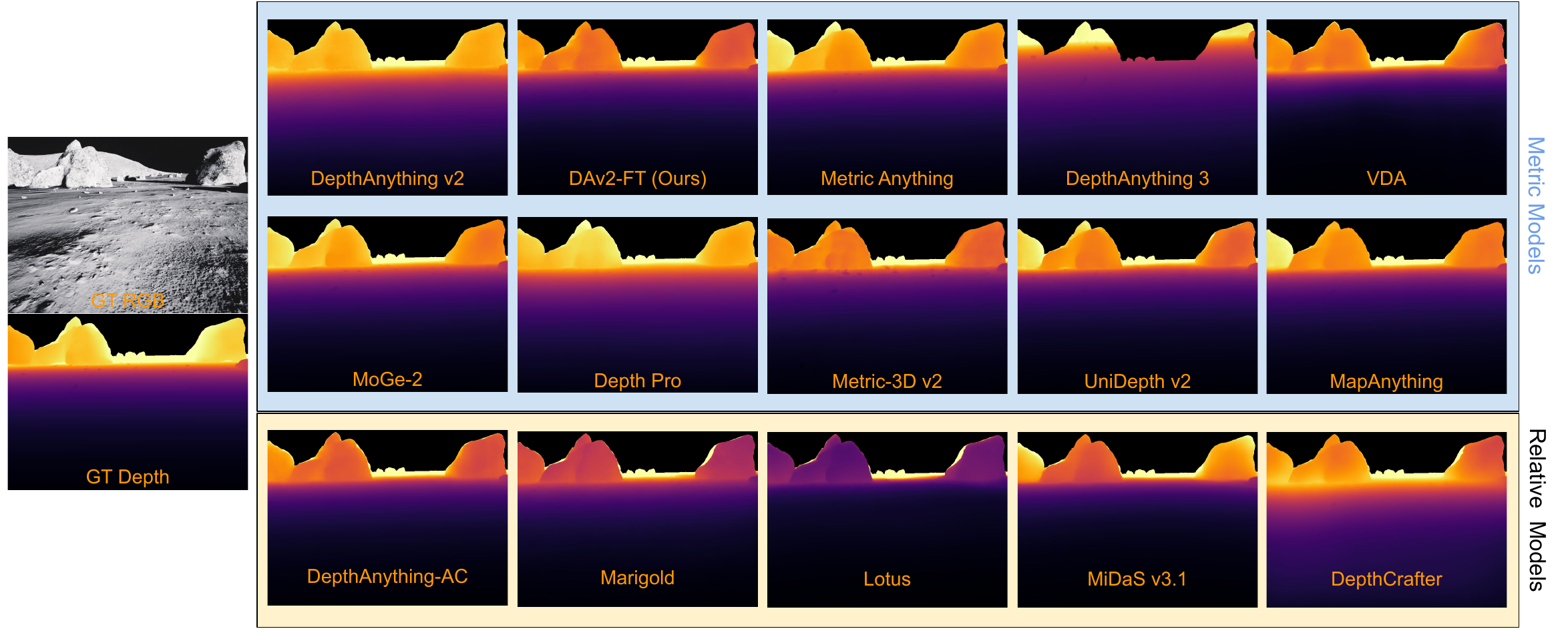}
    \caption{Qualitative depth estimation results on \lusnar}
    \label{fig:supp_vis_s5}
\end{figure}

\begin{figure}[h]
    \centering
    \includegraphics[width=0.9\linewidth]{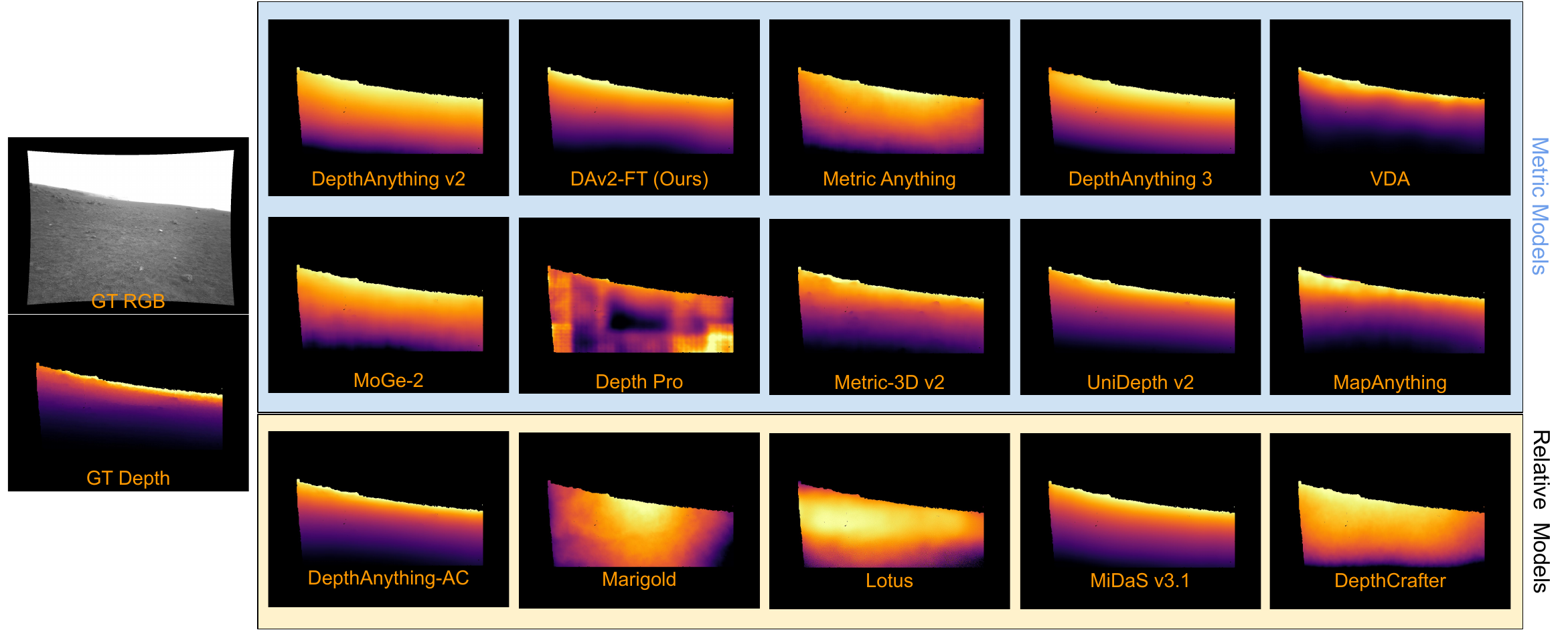}
    \caption{Qualitative depth estimation results on \etna}
    \label{fig:supp_vis_s6}
\end{figure}

\begin{figure}[h]
    \centering
    \includegraphics[width=0.9\linewidth]{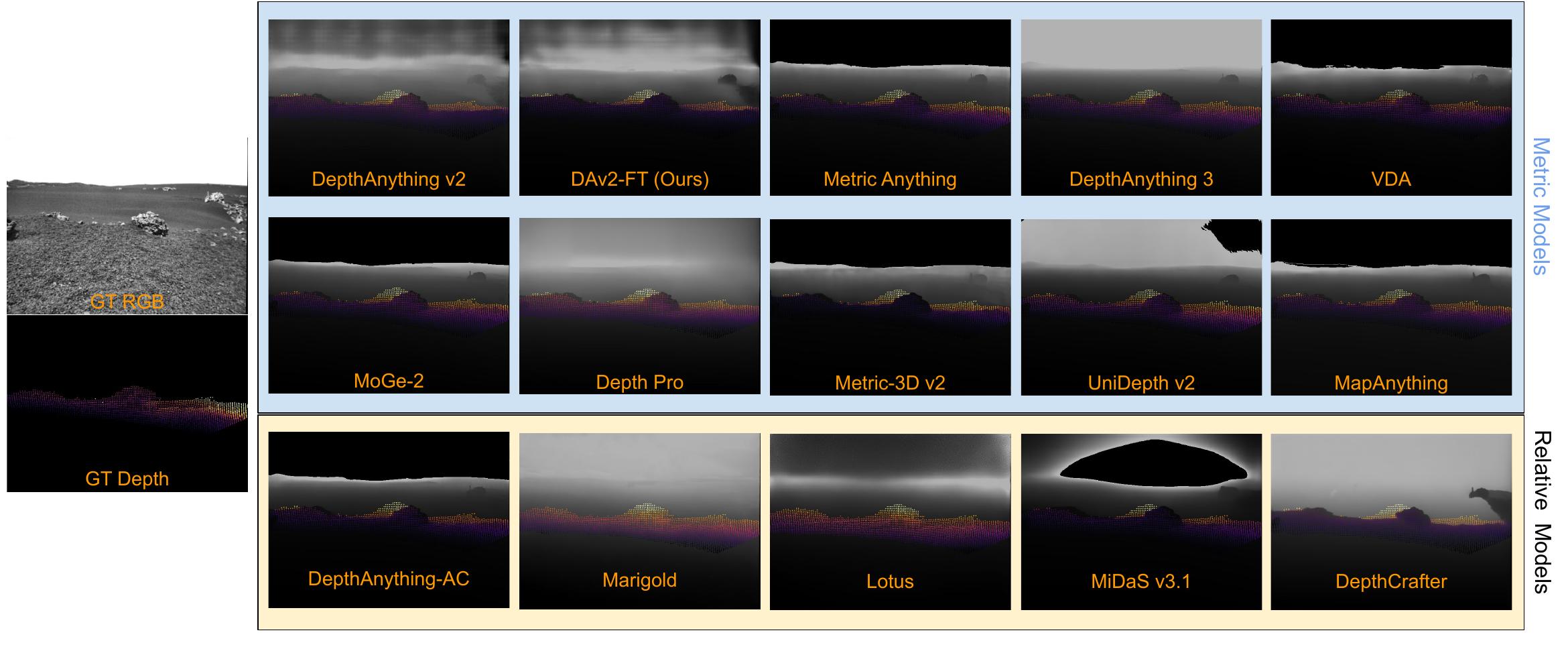}
    \caption{Qualitative depth estimation results on \seli}
    \label{fig:supp_vis_s7}
\end{figure}

\begin{figure}[h]
    \centering
    \includegraphics[width=0.9\linewidth]{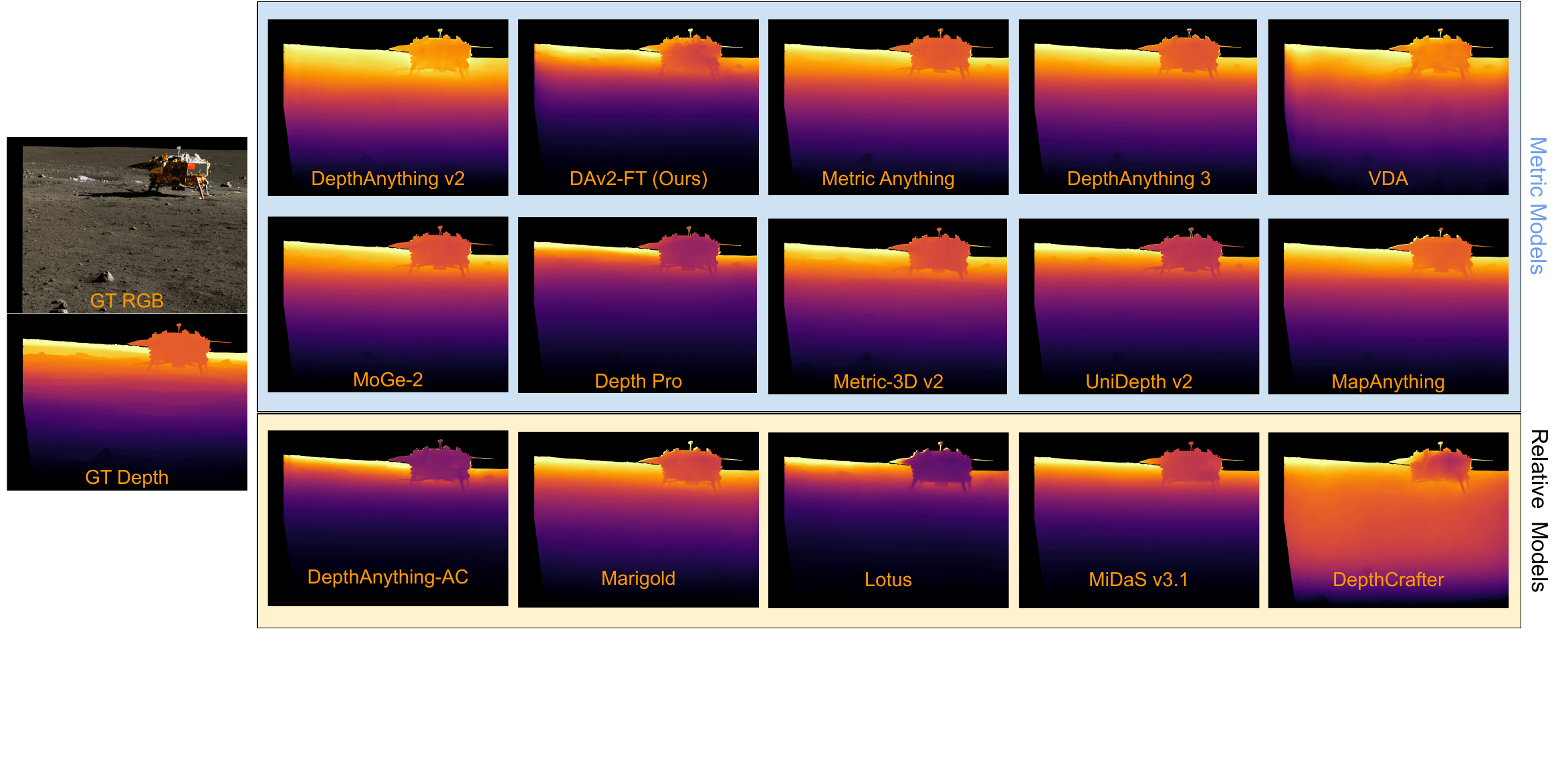}
    \caption{Qualitative depth estimation results on \change}
    \label{fig:supp_vis_s8}
\end{figure}

\begin{figure}[h]
    \centering
    \includegraphics[width=0.9\linewidth]{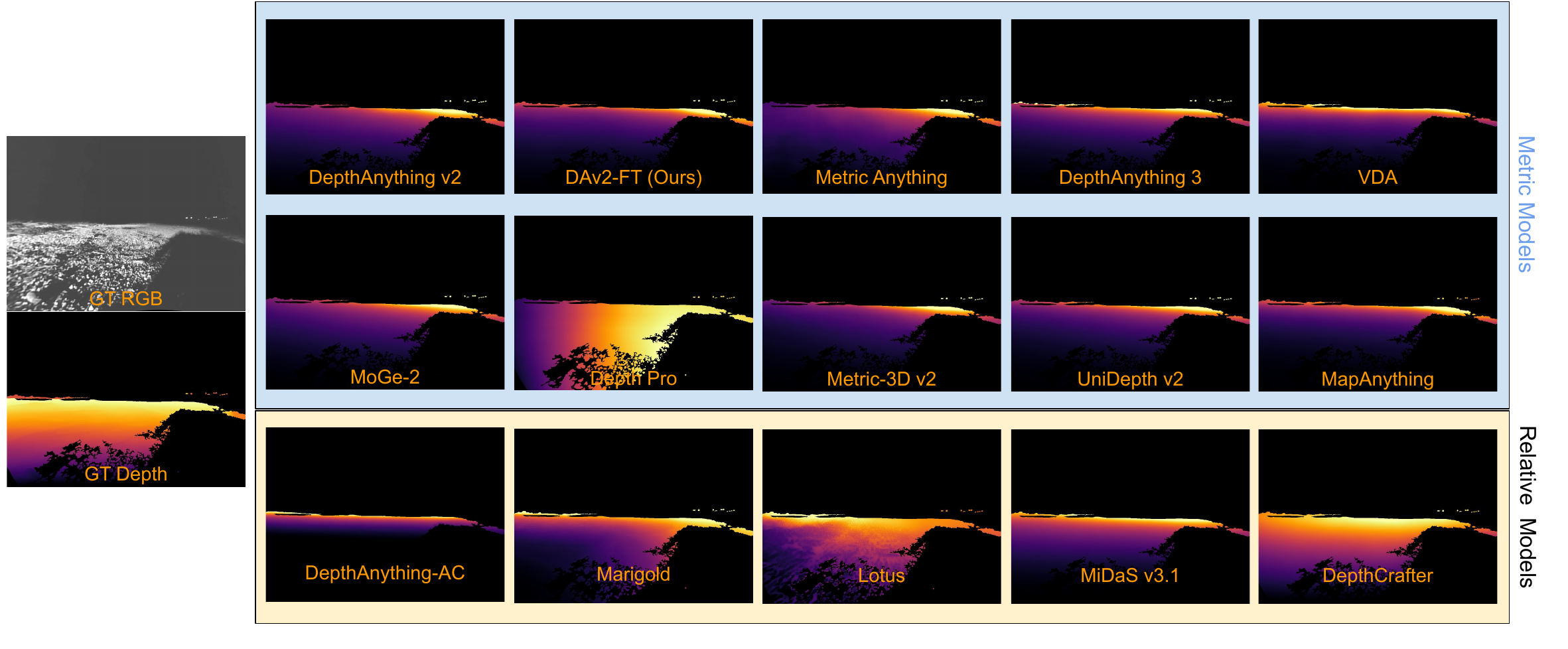}
    \caption{Qualitative depth estimation results on \cheri}
    \label{fig:supp_vis_s9}
\end{figure}

\end{document}